\documentclass{article} % For LaTeX2e
\usepackage[accepted]{icml2024}

\usepackage{microtype}
\usepackage{graphicx}
\usepackage{booktabs} 
\usepackage{hyperref}

\usepackage{mathtools}
\usepackage[utf8]{inputenc} 
\usepackage[T1]{fontenc}    
\usepackage{url}            
\usepackage{amsfonts}       
\usepackage{nicefrac}       
\usepackage{xcolor}         
\usepackage{colortbl}
\usepackage{subcaption}
\usepackage{enumitem}
\usepackage{multirow}
\usepackage{capt-of}
\usepackage{wrapfig}
\usepackage{makecell}
\usepackage{bm}
\usepackage{arydshln}
\usepackage{placeins}
\hypersetup{colorlinks=true}
\hypersetup{citecolor=mountainmeadow}
\hypersetup{linkcolor=crimson}
\hypersetup{linktoc=all}
\hypersetup{urlcolor=darkblue}
\usepackage[all]{hypcap}

\usepackage{comment}
\usepackage{algorithm}
\usepackage{algorithmic}
\usepackage{eqparbox}
\usepackage{amsmath,amsfonts,bm}

\def\eqref#1{equation~\ref{#1}}
\def\1{\bm{1}}

\DeclareMathAlphabet{\mathsfit}{\encodingdefault}{\sfdefault}{m}{sl}
\SetMathAlphabet{\mathsfit}{bold}{\encodingdefault}{\sfdefault}{bx}{n}

\usepackage{adjustbox}

\usepackage{amsmath}
\usepackage{amssymb}
\usepackage{bbm, dsfont}
\usepackage{amsthm}
\usepackage[normalem]{ulem}
\usepackage{cutwin}
\usepackage{caption}
\usepackage{threeparttable}

\renewcommand{\algorithmiccomment}[1]{\hfill\eqparbox{COMMENT}{\footnotesize $\rhd$ #1}}
\newcommand{\eat}[1]{}

\definecolor{Red}{rgb}{0.6,0,0}
\definecolor{Blue}{rgb}{0,0,0.8}
\definecolor{Green}{rgb}{0,0.4,0.7}
\definecolor{airforceblue}{rgb}{0.36, 0.54, 0.66}
\definecolor{ao(english)}{rgb}{0.0, 0.5, 0.0}
\definecolor{azure(colorwheel)}{rgb}{0.0, 0.5, 1.0}
\definecolor{crimson}{rgb}{0.86, 0.08, 0.24}
\definecolor{darkcerulean}{rgb}{0.03, 0.27, 0.49}
\definecolor{cobalt}{rgb}{0.0, 0.28, 0.67}
\definecolor{rosegold}{rgb}{0.72, 0.43, 0.47}
\definecolor{orange-red}{rgb}{1.0, 0.27, 0.0}
\definecolor{mountainmeadow}{rgb}{0.19, 0.73, 0.56}
\definecolor{malachite}{rgb}{0.04, 0.85, 0.32}
\definecolor{darkblue}{rgb}{0.0, 0.0, 0.55}
\definecolor{customblue}{rgb}{0.2, 0.35, 0.8}
\definecolor{darkgreen}{rgb}{0,0.5,0}
\definecolor{ggr}{gray}{0.92}
\newcolumntype{a}{>{\columncolor{ggr}}c}

\definecolor{gg}{HTML}{E0FEFE}
\definecolor{evgg}{HTML}{FFD6A5}
\definecolor{ggg}{gray}{0.65}

\usepackage[nameinlink]{cleveref}
\usepackage[textsize=tiny]{todonotes}
\usepackage[hang,flushmargin]{footmisc} % remove indent from the footnote

\Crefname{section}{Sec.}{Secs.}
\Crefname{algorithm}{Algo.}{Algos.}
\Crefname{table}{Tab.}{Tabs.}
\Crefname{figure}{Fig.}{Figs.}
\Crefname{appendix}{Sec.}{Secs.}

\creflabelformat{equation}{#2\textup{#1}#3}  

\usepackage{svg}

\newcommand{\highlight}[1]{{\color{crimson}{#1}}}

\newcommand{\evblue}[1]{{\color{azure(colorwheel)}{#1}}}

\usepackage{tikz}
\usepackage[most]{tcolorbox}
\usetikzlibrary{fit,calc}

\definecolor{Gray}{gray}{0.91}
\definecolor{figred}{RGB}{255, 181, 164}
\definecolor{figblue}{RGB}{156,192,231}
\definecolor{figgreen}{RGB}{253, 229, 180}
\definecolor{figgray}{RGB}{211, 211, 211}

\definecolor{commentcolor}{RGB}{110,154,155}
\definecolor{commandcolor}{RGB}{255,100,100}
\newcommand{\PyComment}[1]{\ttfamily\textcolor{commentcolor}{\# #1}}
\newcommand{\PyCode}[1]{\ttfamily\textcolor{black}{#1}}
\newcommand{\PyCommand}[1]{\ttfamily\textcolor{commandcolor}{#1}}

\icmltitlerunning{STELLA: Continual Audio-Video Pre-training with Spatio-Temporal Localized Alignment}

\begin{document}
\twocolumn[
\icmltitle{\texorpdfstring{STELLA\includegraphics[width=0.04\linewidth]{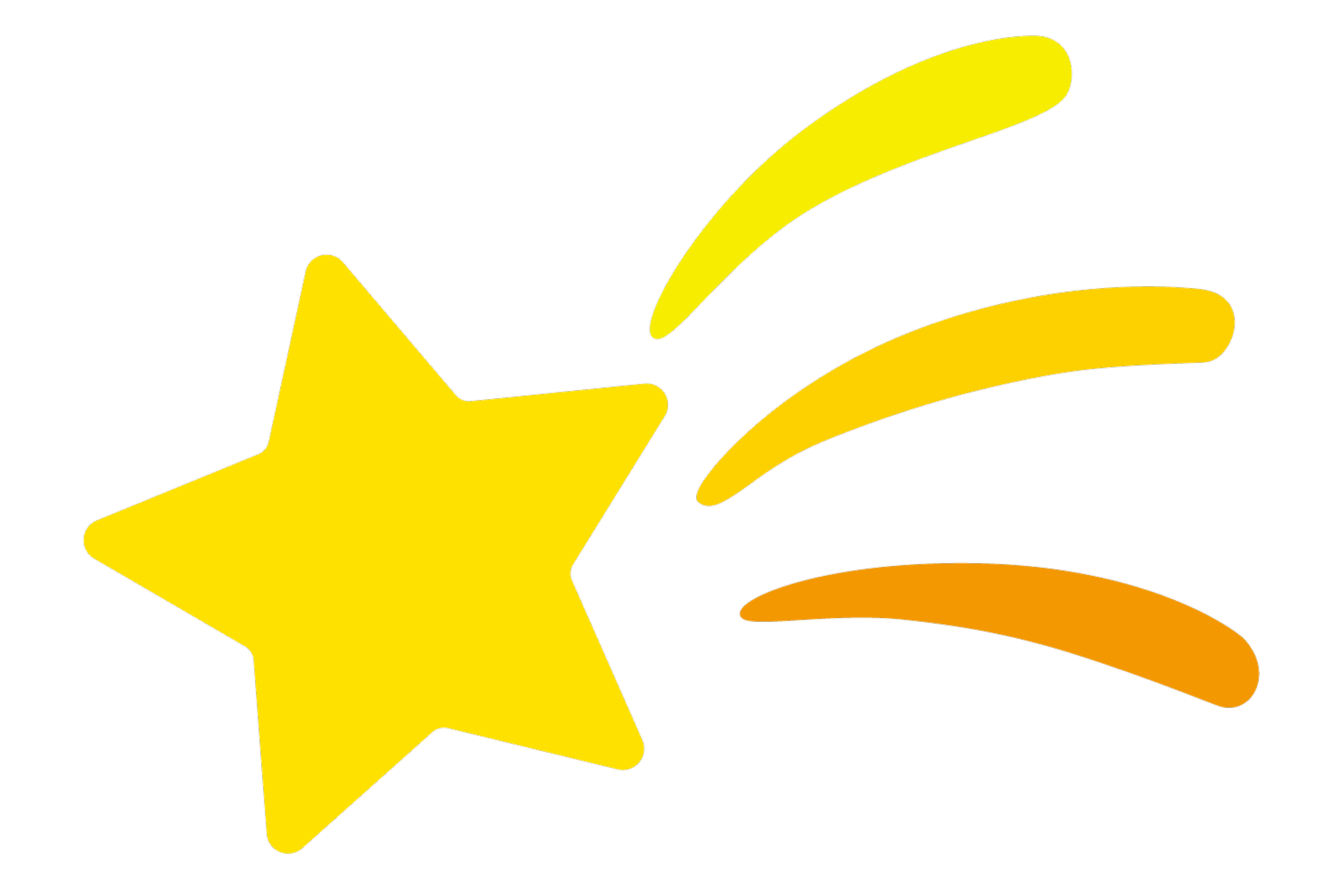}: Continual Audio-Video Pre-training with \\ Spatio-Temporal Localized Alignment}{STELLA: Continual Audio-Video Pre-training with Spatio-Temporal Localized Alignment}}

\icmlsetsymbol{equal}{*}

\begin{icmlauthorlist}
\icmlauthor{Jaewoo Lee}{kaist,equal}
\icmlauthor{Jaehong Yoon}{unc,equal}
\icmlauthor{Wonjae Kim}{naver}
\icmlauthor{Yunji Kim}{naver}
\icmlauthor{Sung Ju Hwang}{kaist,deepauto}\\
\end{icmlauthorlist}

\icmlaffiliation{kaist}{KAIST}
\icmlaffiliation{unc}{UNC Chapel Hill}
\icmlaffiliation{naver}{NAVER AI Lab}
\icmlaffiliation{deepauto}{DeepAuto}

\icmlcorrespondingauthor{Jaewoo Lee}{jwlee8877@kaist.ac.kr}
\icmlcorrespondingauthor{Jaehong Yoon}{jhyoon@cs.unc.edu}
\icmlcorrespondingauthor{Sung Ju Hwang}{sjhwang82@kaist.ac.kr}
\icmlkeywords{Machine Learning, ICML}
\vskip 0.3in
]
% \maketitle
% \printAffiliationsAndNotice{}
\printAffiliationsAndNotice{\;\;\icmlEqualContribution} % otherwise use the standard text.

\begin{abstract}
Continuously learning a variety of audio-video semantics over time is crucial for audio-related reasoning tasks in our ever-evolving world. However, this is a nontrivial problem and poses two critical challenges: \textit{sparse spatio-temporal correlation} between audio-video pairs and \textit{multimodal correlation overwriting} that forgets audio-video relations. To tackle this problem, we propose a new continual audio-video pre-training method with two novel ideas: \textit{(1) Localized Patch Importance Scoring}: we introduce a multimodal encoder to determine the importance score for each patch, emphasizing semantically intertwined audio-video patches.
\textit{(2) Replay-guided Correlation Assessment}: to reduce the corruption of previously learned audiovisual knowledge due to drift, we propose to assess the correlation of the current patches on the past steps to identify the patches exhibiting high correlations with the past steps. Based on the results from the two ideas, we perform probabilistic patch selection for effective continual audio-video pre-training. Experimental validation on multiple benchmarks shows that our method achieves a 3.69\%p of relative performance gain in zero-shot retrieval tasks compared to strong continual learning baselines, while reducing memory consumption by $\sim$45\%. Our code is available at \href{https://cl-stella.github.io/}{\textcolor{magenta}{https://cl-stella.github.io/}}.
\end{abstract}

\section{Introduction}\label{sec:intro}
Multimodal learning is an important problem for various real-world applications, as many real-world data types are multimodal, such as \textit{text-image}~\citep{liao2022text,lee2023text}, \textit{text-video}~\citep{villegas2022phenaki,hu2022make}, and \textit{audio-video}~\citep{korbar2018cooperative,xiao2020audiovisual} pairs.
While most vision-language learning~\citep{li2020cross,yan2023video,liu2023visual} assumes the availability of curated multimodal data with human-annotated descriptions, audiovisual domain~\citep{zhou2019talking,Yuan2023cav} holds a unique and practical advantage, as most videos inherently come with accompanying audios without human annotations. Thanks to this property, audiovisual multimodal learning models can leverage web-scale raw videos (e.g., YouTube, TikTok, etc.) for training with minimal human efforts in data preprocessing, and thus have achieved impressive success in audio-video compositional reasoning~\citep{Zineng2022tvlt,Po-Vao2022mavil,lin2023vision}.

\begin{figure}[t]
    \centering
    \begin{minipage}{\textwidth}
        \includegraphics[width=0.45\linewidth]{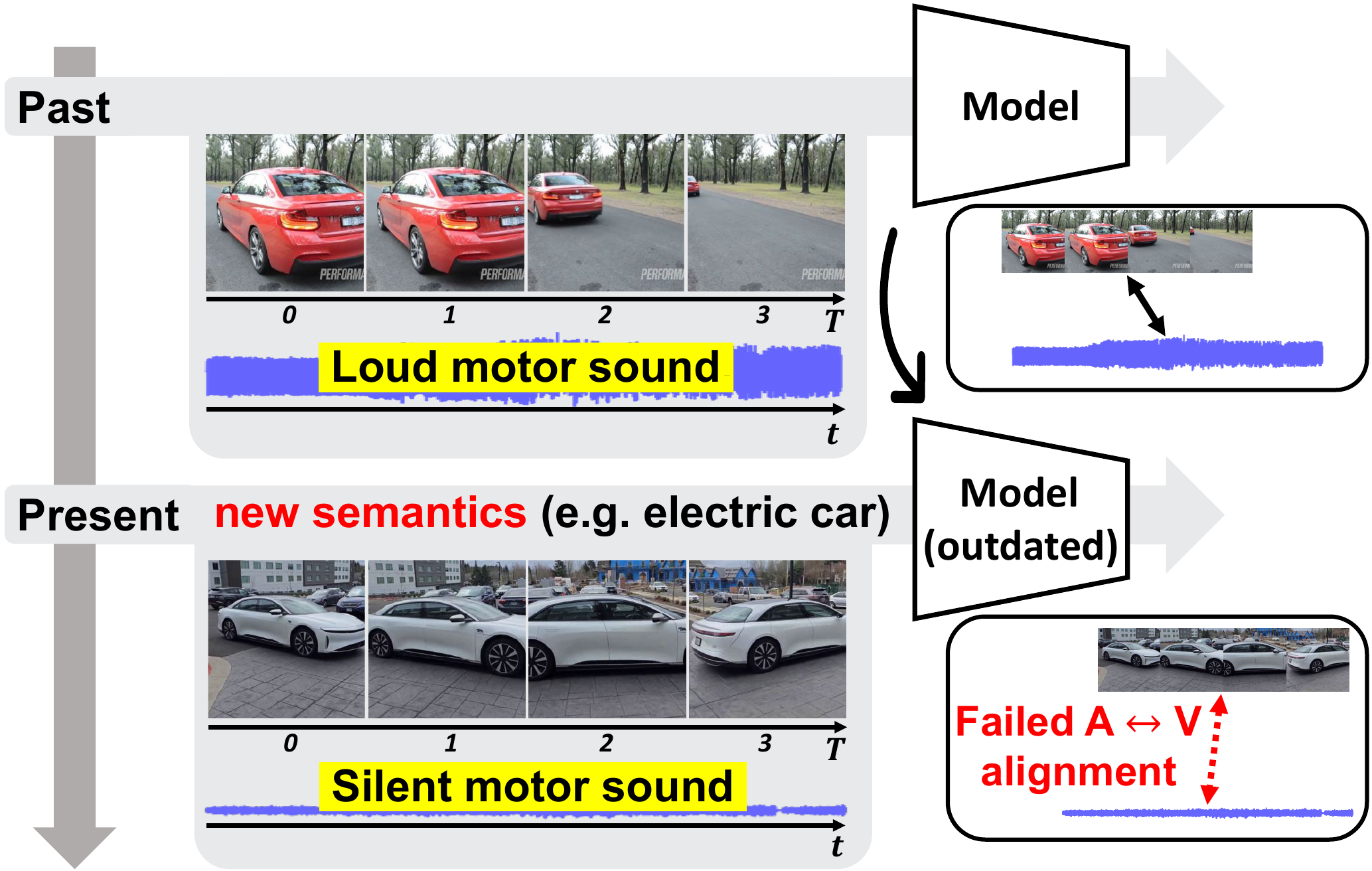}
    \end{minipage}
    % \vspace{-0.05in}
    \captionsetup{justification=justified} % Reset caption justification
    \caption{Outdated pre-trained audio-video models struggle with understanding emerging new audio-video semantics.}\label{fig:issues in offline}
    \vspace{-0.1in}
\end{figure}

However, most existing approaches~\cite{Zineng2022tvlt,Po-Vao2022mavil,Yuan2023cav} struggle when deployed to real-world scenarios, where \textbf{the distribution of training data continuously changes over time with new audio-video semantics.} For example, the audiovisual model pre-trained before electric vehicles became popular, would not be able to associate \textit{cars} with their unique acoustic cues (e.g., motor sound) (See \Cref{fig:issues in offline}). One straightforward solution to this problem is to periodically train the model from scratch using audio-video data collected from the past to the present, but this approach comes with prohibitive computation and memory costs.

\begin{figure}[t]
    \centering
    \begin{minipage}{\linewidth}
    \centering
        \includegraphics[width=0.9\linewidth]{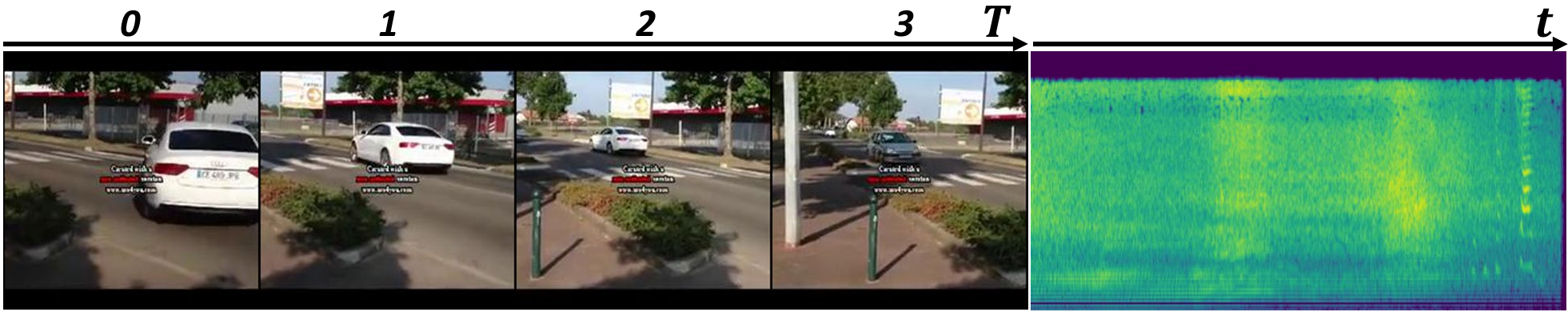}
    \end{minipage}
    \textbf{(a) Raw data}
    \vspace{0.025in}
    
    \begin{minipage}{\linewidth}
    \centering
        \includegraphics[width=0.9\linewidth]{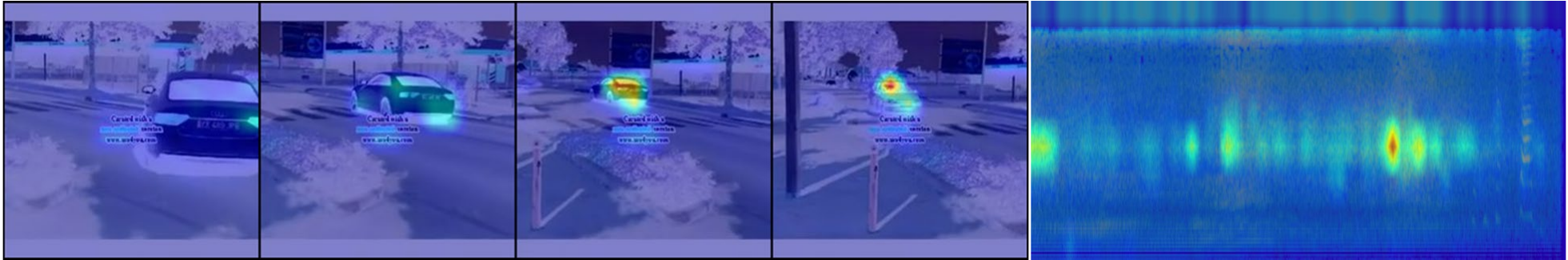}
    \end{minipage}
    \textbf{(b) Sparse audiovisual correlation}
    \vspace{0.025in}
    
    \begin{minipage}{\linewidth}
    \centering
        \includegraphics[width=0.9\linewidth]{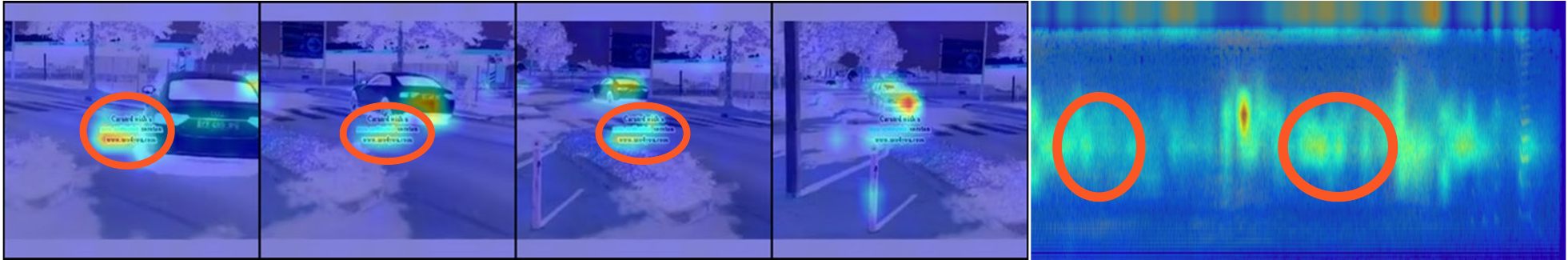}
    \end{minipage}
    \textbf{(c) Multimodal correlation forgetting (DER++)}
    \vspace{0.025in}
    
    \begin{minipage}{\linewidth}
    \centering
        \includegraphics[width=0.9\linewidth]{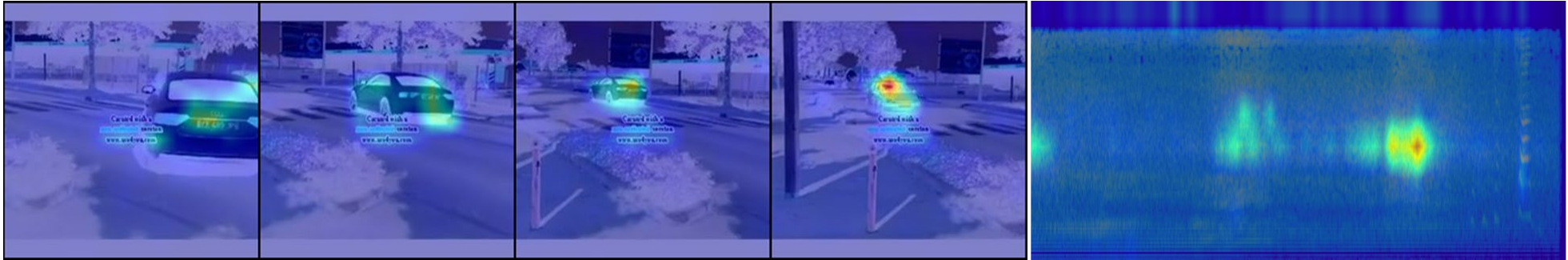}
    \end{minipage}
    \textbf{(d) Multimodal correlation forgetting (Ours)}
    % \vspace{-0.075in}
    \captionsetup{justification=justified} % Reset caption justification
    \caption{\textbf{Challenges in continual audio-video learning.} \textbf{(a):} A raw data pair describing a car and its engine sound. \textbf{(b):} Sparse correlations in cross-attention maps. \textbf{(c):} After training on a series of tasks after (b), \textit{DER++} focuses on entirely different areas (orange circle), presenting correlation forgetting. \textbf{(d):} Our \textit{STELLA} maintains consistent attention. More examples are in~\Cref{fig:supple_fading_attention}.}
    % \vspace{-0.15in}
\label{fig:fading_attention}
\end{figure}

While continual learning is a viable solution for tackling such scenarios, dealing with dynamically evolving audio-video semantics is a nontrivial problem due to two critical challenges. First, the spatio-temporal correlation between the audio-video data is highly sparse. As represented in \Cref{fig:fading_attention} \highlight{(b)}, only a few objects/regions in a video (i.e., sound sources) are strongly correlated with audio. Secondly, audio-video pre-training models encounter the issue of forgetting not only the representations of each modality but also the correlation between them. As \textit{orange circles} in \Cref{fig:fading_attention} \highlight{(c)} illustrate, the model which initially learned the accurate audio-video correlation in a car's engine video, forgets this correlation after learning on a series of audio-video tasks. It instead highlights inaccurate regions in the audio-video data, as if there were highly fine-grained multimodal alignment.

To overcome these challenges in learning multiple audio-video tasks sequentially, we propose \textit{\textbf{S}patio-\textbf{TE}mporal \textbf{L}oca\textbf{L}ized \textbf{A}lignment (\textbf{STELLA})}, a novel approach that exploits past and current information via audio-video attention maps.
Specifically, our goal is to continually pre-train the model by selecting audio and video patches that have a high correlation for its modality pair and also preserve previously learned audio-video correlation. Thereby we propose a probabilistic patch selection framework that enables the model to learn better audio-video correlations and preserve past audio-video semantics, based on two key components: first, we use the averaged cross-attention maps obtained by a lightweight multimodal encoder to compute an importance score, estimating how each audio (or video) patch is important for its modality pair. 
Further, to preserve the past correlation during continual audio-video pre-training, we leverage new cross-attention maps activated by the key and query embeddings between the current and past steps, respectively. This yields a correlation score that identifies the patches that exhibit a higher correlation with the current step than the past steps. We extensively validate our method on continual audio-video pre-training scenarios, using diverse benchmark datasets evaluated on various audiovisual downstream tasks. Our method outperforms strong baseline on various tasks with enhanced efficiency by reducing the GPU memory by $\sim$45\% during continual pre-training. We further provide extensive in-depth analysis with visualizations. 

Our paper makes the following key contributions:
\vspace{-0.1in}
\begin{itemize}[itemsep=2mm, parsep=1pt, leftmargin=*]
\item We are the first to address continual audio-video pre-training, which poses new challenges: \textit{sparse spatio-temporal correlation} between audio-video pairs and \textit{multimodal correlation overwriting} that forgets their relations.

\item We propose a novel method that leverages cross-attention maps to capture sparse audio-video relationships and mitigate forgetting of previously learned relationships.

\item We demonstrate the efficacy of our method on several audiovisual downstream tasks including retrieval, sound source localization and event localization. In particular, ours achieves 3.69\%p of performance gain in the retrieval task and reduces the GPU memory consumption by $\sim$45\% during training, compared to the strongest baseline.

\end{itemize}

\section{Related Work}\label{sec:related_work}

\paragraph{Audiovisual understanding}
Self-supervised learning on audiovisual data aims to learn transferable representations that can be applied to a variety of audio-image/video downstream tasks, including action recognition/event classification~\citep{Nagrani2021mbt, Lee2021paraeffivit}, sounding object localization~\citep{Hu2022MixandLocal, Liu2022cmerrasing}, and multimodal retrieval~\citep{Po-Vao2022mavil, Yuan2023cav}. 
Inspired by the success of Masked AutoEncoders (MAE) in visual pre-training~\citep{KHe2022mae}, recent audiovisual representation learning adopts masked modeling for comprehending audiovisual semantics~\citep{Zineng2022tvlt, Yuan2023cav}.
TVLT~\citep{Zineng2022tvlt} adopts the MAE structure and audiovisual matching to predict whether audio and visual data originated from the same video. CAV~\citep{Yuan2023cav} combines the MAE with audiovisual contrastive learning, which pulls matching audiovisual pairs closer and pushes non-matching pairs apart. Their methods assume a fixed input data distribution that does not shift throughout training. However, in the real world, a machine/agent will continuously encounter new (i.e., changing distribution) audio-video tasks/semantics. If not well managed, the methods will suffer severe performance degradation if they encounter the aforementioned shift in continual learning, a challenging and realistic scenario for multimodal learning.

\paragraph{Multimodal continual learning}
Continual learning~\citep{Kirkpatrick2016ewc,Rebuffi2017icarl,Ahn2019ucl} refers to a learning paradigm in which a model sequentially learns an unlimited number of tasks/domains. It aims to continuously adapt to new tasks while preserving previously learned knowledge/skills, which is crucial for real-world AI deployment.
A number of works have addressed supervised learning for vision tasks~\citep{Zenke2017si,Yoon2018den,Lee2020cndpm}, and very recently, a few approaches have explored continual learning with self-supervised learning~\citep{Madaan2022lump, Andrea2022clpretrain, Fini2022selfsupcl,yoon2023continual}, and multimodal learning~\citep{Yan2022incclip, Pian2023avcil, Mo2023groupav}. AV-CIL~\citep{Pian2023avcil} and CIGN~\citep{Mo2023groupav} tackle the problem of supervised continual learning for audio-video tasks. However, they require dense human annotations, such as text or audiovisual labels, and task boundary information to know when new tasks are introduced during continual learning. On the other hand, our \textit{STELLA} focuses on continual pre-training of audio-video models without any human-effort labels or task boundary information. Moreover, our work extends to investigating the impact of past data on the current audio and video attention map activation, while the AV-CIL focuses on maintaining the past visual attention map.
\section{Continual Audio-Video Pre-training} \label{sec:approach}
\subsection{Problem Statement}\label{sec:subsec:problem_statement}
In this work, we tackle the problem of continual audio-video pre-training, under the assumption that the data distribution continuously changes during pre-training, and the model does not have direct access to previously seen data and stores only a small subset in the rehearsal memory~\citep{Rolnick2019ER,Buzzega2020DER}. Furthermore, we assume a task-free scenario~\citep{Aljundi_2019_CVPR} where the model performs the pre-training and inference without the explicit knowledge of task boundaries, which is challenging yet realistic as the model does not need any human guidance on the change of data distributions. Following the setup in continual learning literature~\citep{Madaan2022lump,Sarfraz2023esmer}, we formulate pre-training of the audio-video learning model over a sequence of $\mathcal{T}$ disjoint audio-video datasets $\mathcal{D}\!=\!\{\mathcal{D}_i\}^\mathcal{T}_{i=1}$. For the $i$-th task, the model iteratively samples $B$ audio-video pairs 
$(X^{i}_{a},X^{i}_{v})\!\sim\! \mathcal{D}_{i}$\footnote{We omit the task index for brevity, unless otherwise stated.}. Here, $X_{a}\!\in\!\mathbb{R}^{B \!\times\! M \!\times\! p \!\times\! p}$ represents the audio patches, patchfied from the audio spectrogram with time ($t$) and frequency ($f$) dimensions, where $M \!=\!\left| t / p \right|\!\cdot\!\left| f / p \right|$ and $p$ is the patch size. Similarly, $X_{v}\!\in\!\mathbb{R}^{ B \!\times\! N \!\times\! p \!\times\! p}$ represents the video patches, obtained from the video clip with channel, frames ($T$), height ($h$), and width ($w$) dimensions, where $N \!=\!\left| T \right|\!\cdot\!\left| h / p \right|\!\cdot\!\left| w / p \right|$.

Following~\citep{Yuan2023cav}, the model $f(\cdot;\bm\theta)$ comprises audio-video encoders, a multimodal fusion encoder, and a decoder. For pre-training, we adopt two loss terms: \textit{reconstruction loss} ($\ell^{r}$) for masked patches to understand low-level audio-video features, and \textit{masked contrastive loss} ($\ell^{c}$) for pooled audio-video features to learn semantic relationships between the two. During each training iteration for task $i$, the model updates weights by minimizing the objective $\mathcal{L}\!=\!\ell^{r}(f_{\bm\theta}(\mathcal{D}_i))+\lambda \ell^{c}(f_{\bm\theta}(\mathcal{D}_i))$, with a balancing term $\lambda$. The detailed mathematical expressions of the loss functions are explicated in~\Cref{sec:supple:av-self-sup objectives}. Then, we evaluate the learned representations through various audiovisual downstream tasks at the end of the task.

\subsection{Challenges in Continual Audio-Video Pre-training \label{sec:subsec:false alignment}}
\begin{figure}[t]
    % \vspace{-0.05in}
    \centering
    \begin{minipage}{\textwidth}
        \includegraphics[width=0.48\linewidth]{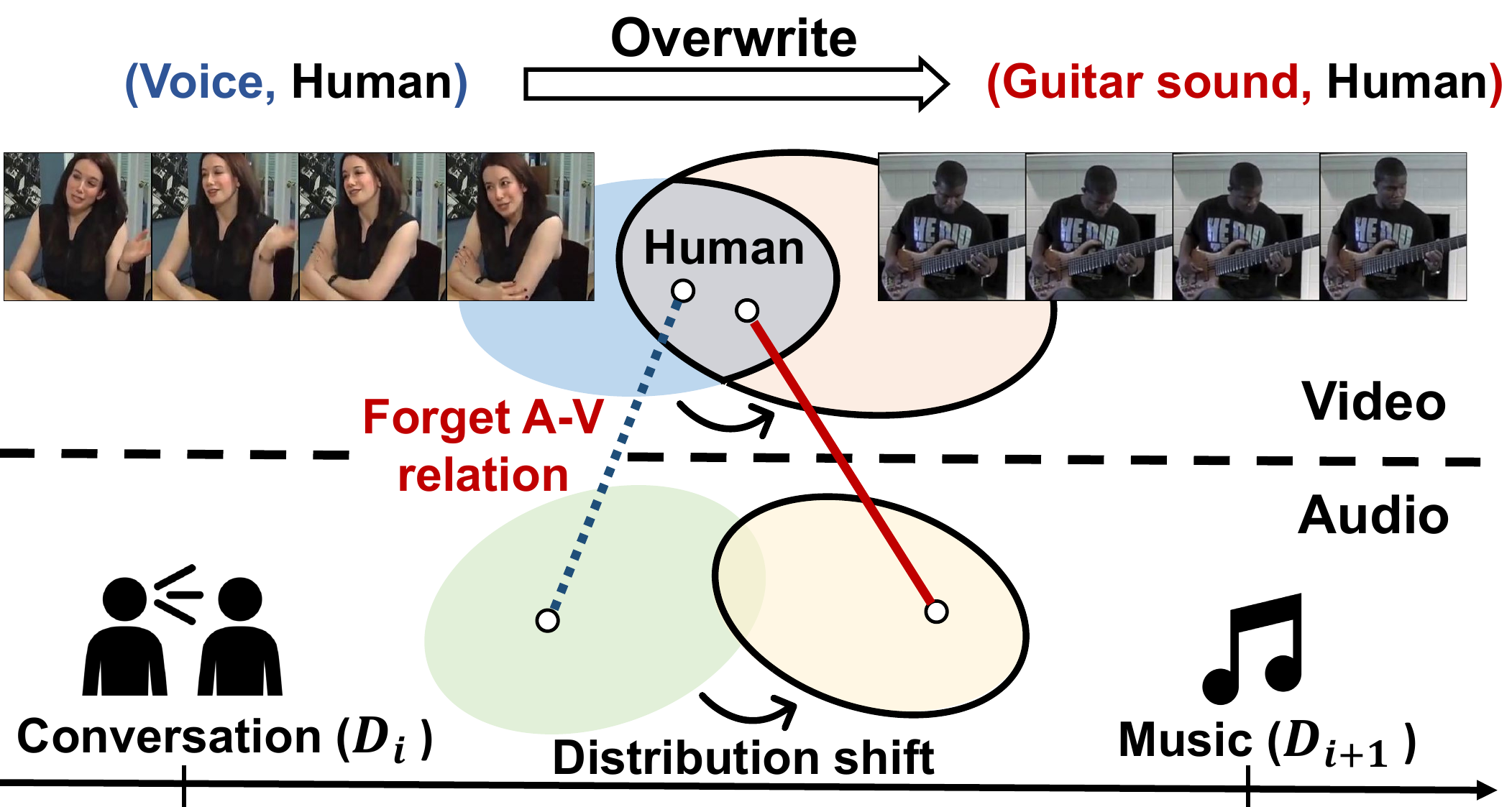}
    \end{minipage}
    % \vspace{-0.1in}
    \captionsetup{justification=justified} % Reset caption justification
    \caption{\textbf{Challenge of multimodal correlation overwriting.} Let the model be learned human voice with video frame inputs (\textcolor{blue}{blue}). During continual pre-training, the model can encounter new semantics sharing key visual objects, humans, making the model overwrite the previously learned audio information associated with humans to a new one (i.e., guitar) (\textcolor{red}{red}), resulting in forgetting.}
    \label{fig:forgetting}
    % \vspace{-0.15in}
\end{figure}

\begin{figure*}[t]
    \centering
    \vspace{0.2in}
    \begin{tabular}{c}
    \includegraphics[width=0.98\linewidth]{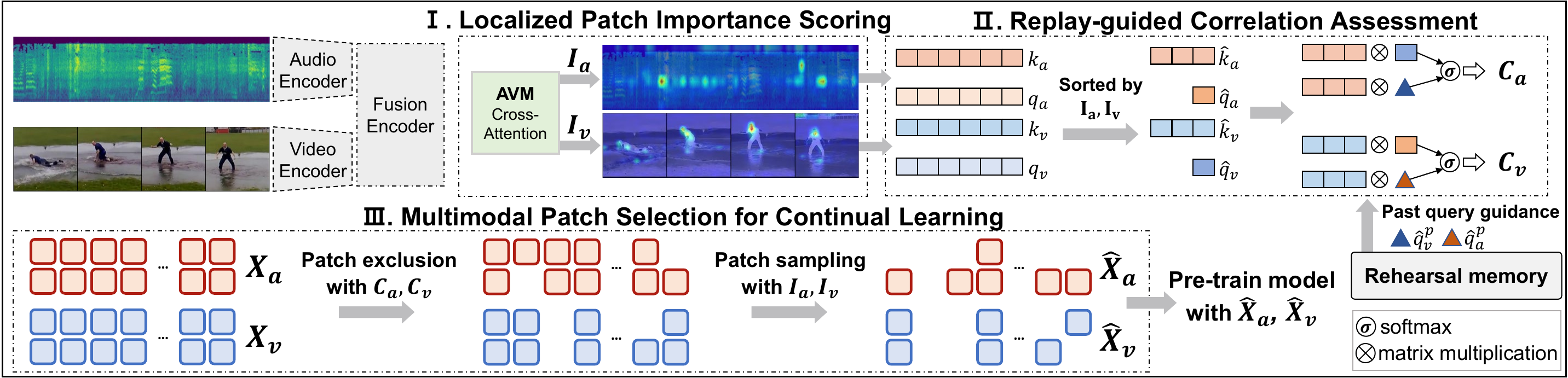}
    \end{tabular}
    % \vspace{-0.05in}
    \caption{\textbf{Overview of our approach.} Our method harnesses cross-modal attention maps from the AVM module to compute importance scores in order to identify highly correlated patches (\Cref{sec:subsec:positive region proposal}). Comparing the attention maps created by the current queries with those generated by past queries, we compute correlation scores of the current patches with the past data (\Cref{sec:subsec:forget-robust selection}). Finally, we perform a probabilistic patch selection, combining the importance scores and correlation scores to select patches for continual audio-video pre-training (\Cref{sec:subsec:patch selection}).}
    \label{fig:overview}
\end{figure*}

In this section, we delve into two key challenges in continual audio-video pre-training: 1) \textit{sparse spatio-temporal correlation} 2) \textit{multimodal correlation overwriting}. In~\Cref{fig:fading_attention} \highlight{(b)}, we visualize cross-attention heat maps and observe \textit{sparse spatio-temporal correlation} between the audio-video pair. Capturing highly correlated audio-video patches is crucial for understanding their semantics, allowing the model to focus on informative regions and learn complex multimodal relationships. It becomes more critical in continual audio-video pre-training methods in view of \textit{rehearsal memory}. They contain a small-sized rehearsal memory designed to store key information for past tasks during continual pre-training. As rehearsal memory is limited in capacity, it's important to store meaningful data/feature audio-video pairs associated with their semantics.

We also observe that the model forgets previously learned audio-video correlations after learning a sequence of tasks (\Cref{fig:fading_attention} \highlight{(c)}). In continual audio-video pre-training, the biased data distribution poses a risk of overwriting previous multimodal correlations, driven by the close correlation between current video and past audio data, and vice versa. For instance, transitioning from a past task involving human-conversational data to a current task featuring human-playing-musical-instrument data (\Cref{fig:forgetting}) weakens the audio-video correlations of human visuals and voices from the past task. Instead, the model potentially associates human visuals with musical sounds prevalent in the biased current data distribution, leading to the forgetting of the past human-voice relationships. This challenge, termed \textit{multimodal correlation overwriting}, underscores the critical need to identify data regions with high correlation to past steps.

\section{Continual Audio-Video Pre-training with Spatio-Temporal Localized Alignment}

To overcome critical challenges in earlier sections, we introduce a novel continual audio-video pre-training approach, dubbed \textit{\textbf{S}patio-\textbf{TE}mporal \textbf{L}oca\textbf{L}ized \textbf{A}lignment (\textbf{STELLA})}, illustrated in~\Cref{fig:overview}. We first propose a lightweight trainable module that determines importance scores, guiding the model to focus on spatio-temporally aligned audio-visual regions (\Cref{sec:subsec:positive region proposal}). Next, we introduce a unique process of assessing multimodal correlations between current and previous steps to compute correlation scores, identifying patches having higher correlations to the past steps (\Cref{sec:subsec:forget-robust selection}). Finally, we describe the probabilistic patch selection framework, which uses the importance and correlation scores to select audio and video patches for continual pre-training (\Cref{sec:subsec:patch selection}). Please see~\Cref{alg:stella_algo} for a detailed training process.

\subsection{Localized Patch Importance Scoring \label{sec:subsec:positive region proposal}}
Inspired by the observation that audio-video data pairs are only correlated with a sparse spatio-temporal region, we aim to capture accurate local semantics between audio and visual cues by computing importance scores for each patch to identify a few strongly associated audio-video patches. We achieve this by introducing an Audio-Video Matching (AVM) module that uses cross-attention to capture core audio-video patches. 
Given $(X_a, X_v)$, we first map audio/video patches using the modality encoders and fusion encoder to output tokens $(\bm{o}_a,\bm{o}_v)$. Then, we fed the tokens to the AVM module to map them to queries and keys ($\bm{q},\bm{k}$) to compute cross-attention maps as follows:
\begin{align}
        \bm{q}_a\!=\!\bm{o}_{a}{\mathcal{W}^{Q}_{a}},&\;\bm{k}_a\!=\!\bm{o}_{a}{\mathcal{W}^{K}_{a}},\;\bm{q}_v\!=\!\bm{o}_{v}{\mathcal{W}^{Q}_{v}},\;\bm{k}_v\!=\!\bm{o}_{v}{\mathcal{W}^{K}_{v}}, \notag \\
        \bm{A}_{a}\!&=\!\mu(\bm{q}_v,\bm{k}_a)\!=\!\bm{q}_{v}\bm{k}_{a}^{\top}/\beta*\sqrt{d}, \label{eq:cross_att_compute}\\
        \bm{A}_{v}\!&=\!\mu(\bm{q}_a,\bm{k}_v)\!=\!\bm{q}_{a}\bm{k}_{v}^{\top}/\beta*\sqrt{d}, \notag
\end{align}
where the projections ${\mathcal W}^{Q}_{a},\,{\mathcal W}^{K}_{a},\,{\mathcal W}^{Q}_{v},\,{\mathcal W}^{K}_{v}\!\in\!\mathbb{R}^{D \!\times\! H \!\times\! d}$ are trainable parameter matrices in the AVM module, $H$ is the number of heads, $D\!=\!H*d$ is the dimension size, $\beta$ denotes a temperature coefficient, $(\bm{q}_{a},\bm{k}_{a})\!\in\!\mathbb{R}^{B\!\times\!H\!\times\!M\!\times\!d}$,\, $(\bm{q}_{v},\bm{k}_{v})\!\in\!\mathbb{R}^{B\!\times\!H\!\times\!N\!\times\!d}$ are audio and video keys and queries, $\bm{A}_{a}\!\in\!\mathbb{R}^{B \!\times\! H \!\times\! N \!\times\! M}$, $\bm{A}_{v}\!\in\!\mathbb{R}^{B \!\times\! H \!\times\! M \!\times\! N}$ are computed cross-attention maps. Please see~\Cref{sec:supple:AVM training} for the detailed architecture of the AVM module.

Then, we compute the importance scores $\bm{I}_{a}\!\in\!\mathbb{R}^{B \!\times\! 
 M}$, and $\bm{I}_{v}\!\in\!\mathbb{R}^{B \!\times\! N}$ by applying Softmax normalization on the last dimension:
\begin{align}
    \begin{split}
    \bm{I}_{a}&=\texttt{MeanPool}\left(\texttt{Softmax}\left(\bm{A}_{a}\right)\right), \\
    \bm{I}_{v}&=\texttt{MeanPool}\left(\texttt{Softmax}\left(\bm{A}_{v}\right)\right).
    \label{eq:imp_score_cal}
    \end{split}
\end{align}
The importance score represents the average correlation between an audio (or a video) patch and the paired modality patches. That is, the higher value in $\bm{I}$ indicates the higher importance of the corresponding patch in view of the opposite modality (A$\leftrightarrow$V), thus helping the model to select locally aligned audio-video patches in~\Cref{sec:subsec:patch selection}.

\subsection{Replay-guided Correlation Assessment \label{sec:subsec:forget-robust selection}}

To tackle the challenge of \textit{multimodal correlation overwriting}, the model requires a careful balance between retaining previous knowledge and adapting new one. Thus, we propose to compare cross-attention maps activated by current and past queries to assess relative multimodal correlation and exclude patches exhibiting higher correlation to the past steps. Our ultimate goal is to select $\kappa_{a}$ audio and $\kappa_{v}$ video patches where $\kappa_{a}=M\cdot\rho_{a}$ and $\kappa_{v}=N\cdot\rho_{v}$, with $\rho_{a}$ and $\rho_{v}$ denoting sampling ratios for audio and video. To this end, we obtain locally aligned queries $\widehat{\bm{q}}_{a}, \widehat{\bm{q}}_{v}\!\in\!\mathbb{R}^{B\!\times\!H\!\times\!d}$ and keys $\widehat{\bm{k}}_{a}\!\in\!\mathbb{R}^{B\!\times\!H\!\times\!\kappa_{a}\!\times\!d}, \widehat{\bm{k}}_{v}\!\in\!\mathbb{R}^{B\!\times\!H\!\times\!\kappa_{v}\!\times\!d}$ using the indices sorted in ascending order based on the importance scores
$\bm{S}_{a}\!=\!\texttt{argsort}(\bm{I}_{a})$, $\bm{S}_{v}\!=\!\texttt{argsort}(\bm{I}_{v})$:
\begin{align}
    \widehat{\bm{q}}_{n}[i,:,j]&=\bm{q}_{n}[i,:,\bm{S}_{n}[i,j]],\;\bm{I}_{n}^{s}[i,j]=\bm{I}_{n}[i,\bm{S}_{n}[i,j]], \notag \\
    \widehat{\bm{q}}_{n}&\leftarrow\texttt{MeanPool}\left(\widehat{\bm{q}}_{n},\texttt{weight}\!=\!\bm{I}_{n}^{s}\right), \label{eq:discriminative_info_gather}\\
    \widehat{\bm{k}}_{n}[i,:,j]&=\bm{k}_{n}[i,:,\bm{S}_{n}[i,j]],\;i=1,\ldots,B,\;j=1,\ldots,\kappa_{n}, \notag
\end{align}
where $n\!\in\!(a,v)$ and $\texttt{MeanPool}\left(\cdot,\texttt{weight}\right)$ indicates weighted mean operation. We utilize the queries and keys to compute cross-attention maps $\widehat{\bm{A}}_{a}\!=\!\mu_(\widehat{\bm{q}}_v,\widehat{\bm{k}}_a)\!\in\!\mathbb{R}^{B \!\times\! H \!\times\!\kappa_{a}}$, $\widehat{\bm{A}}_{v}\!=\!\mu(\widehat{\bm{q}}_a,\widehat{\bm{k}}_v) \!\in\!\mathbb{R}^{B \!\times\! H \!\times\!\kappa_{v}}$. Similarly, we compute cross-attention maps $\widehat{\bm{A}}_{a}^{p}\!=\!\mu(\widehat{\bm{q}}_{v}^{p},\widehat{\bm{k}}_a)$,$\widehat{\bm{A}}_{v}^{p}\!=\!\mu(\widehat{\bm{q}}_{a}^{p},\widehat{\bm{k}}_v)$ by using the past queries $\widehat{\bm{q}}_{a}^{p},\;\widehat{\bm{q}}_{v}^{p}$, which were computed during the past steps and stored in the rehearsal memory. Each $\widehat{\bm{A}}$ shows how the given queries are correlated to the current patches. To assess the relative correlation between the past and current steps on the current patches, we stack the audio $(\widehat{\bm{A}}_{a},\widehat{\bm{A}}_{a}^{p})$ and video attention maps $(\widehat{\bm{A}}_{v},\widehat{\bm{A}}_{v}^{p})$, resulting in an extended last dimension, respectively. Subsequently, we apply Softmax normalization on the extended last dimension, resulting in correlation scores $\bm{C}_{a}$ and $\bm{C}_{v}$ as follows:
\begin{align}
    \begin{split}
    \bm{C}_{a}&=\texttt{MeanPool}\left(\texttt{Softmax}\left([\widehat{\bm{A}}_{a},\widehat{\bm{A}}_{a}^{p}]\right)\right), \\
    \bm{C}_{v}&=\texttt{MeanPool}\left(\texttt{Softmax}\left([\widehat{\bm{A}}_{v},\widehat{\bm{A}}_{v}^{p}]\right)\right).
    \label{eq:forget_induce_matrix}
    \end{split}
\end{align}
Each value in the correlation score moves closer to \textit{one} when the corresponding patch exhibits a higher multimodal correlation with the opposite modality data from the past steps compared to the correlation with its modality pair. Hence, patches with high $\bm{C}$ values should more likely be excluded to preserve previously learned multimodal correlations.

\begin{table*}[t]
\tiny
% \vspace{-0.05in}
\caption{Audiovisual zero-shot retrieval tasks on \textit{Continual-VS} and \textit{Continual-AS}. R@K means top-K recall. The results are the means of 3 independent runs. The best and the second best results are highlighted in \textbf{bold} and \underline{underline}, respectively.}
% \vspace{-0.1in}
\centering
\resizebox{0.99\textwidth}{!}{
\renewcommand{\arraystretch}{0.9}
\renewcommand{\tabcolsep}{2.5pt}
\begin{tabular}{ll cc cc cc aa c cc cc cc a a}
\toprule
& & \multicolumn{8}{c}{\textbf{Continual-VS}} && \multicolumn{8}{c}{\textbf{Continual-AS}} \\
& {\textbf{Method}}&\multicolumn{2}{c}{\textbf{R@1}} &\multicolumn{2}{c}{\textbf{R@5}}& 
\multicolumn{2}{c}{\textbf{R@10}} &\multicolumn{2}{c}{\textbf{Avg}}&\;\;\;\;&\multicolumn{2}{c}{\textbf{R@1}} &\multicolumn{2}{c}{\textbf{R@5}}& 
\multicolumn{2}{c}{\textbf{R@10}} &\multicolumn{2}{c}{\textbf{Avg}}\\
\midrule
& & $\mathcal{A} \uparrow$ & $\mathcal{F} \downarrow$ & $\mathcal{A} \uparrow$ & $\mathcal{F} \downarrow$ & $\mathcal{A} \uparrow$ & $\mathcal{F} \downarrow$ & $\mathcal{A} \uparrow$ & $\mathcal{F} \downarrow$ &\;\;\;\;& $\mathcal{A} \uparrow$ & $\mathcal{F} \downarrow$ & $\mathcal{A} \uparrow$ & $\mathcal{F} \downarrow$ & $\mathcal{A} \uparrow$ & $\mathcal{F} \downarrow$ & $\mathcal{A} \uparrow$ & $\mathcal{F} \downarrow$ \\
\midrule
\parbox[t]{2mm}{\multirow{11}{*}{\rotatebox[origin=c]{90}{Audio-to-Video}}}
& Finetune &
{\scriptsize 0.98} & {\scriptsize 4.16} &
{\scriptsize 3.75} & {\scriptsize 11.98} &
{\scriptsize 6.17} & {\scriptsize 15.35} & 
{\scriptsize 3.63} & {\scriptsize 10.50} &\;\;\;\;&
{\scriptsize 1.48} & {\scriptsize 2.90} &
{\scriptsize 3.84} & {\scriptsize 11.34} &
{\scriptsize 5.41} & {\scriptsize 17.83} & 
{\scriptsize 3.58} & {\scriptsize 10.69} \\
& ER &
{\scriptsize 4.09} & {\scriptsize 3.66} & 
{\scriptsize 11.66} & {\scriptsize 9.17} & 
{\scriptsize 17.78} & {\scriptsize 10.20} & 
{\scriptsize 11.18} & {\scriptsize 7.68} &\;\;\;\;&
{\scriptsize 4.94} & {\scriptsize 2.97} & 
{\scriptsize 12.33} & {\scriptsize 7.46} & 
{\scriptsize 17.60} & {\scriptsize 11.17} & 
{\scriptsize 11.62} & {\scriptsize 7.20} \\
& MIR &
{\scriptsize 4.59} & {\scriptsize 3.14} & 
{\scriptsize 12.26} & {\scriptsize 8.34} & 
{\scriptsize 17.51} & {\scriptsize 11.17} & 
{\scriptsize 11.45} & {\scriptsize 7.55} &\;\;\;\;&
{\scriptsize 5.21} & {\scriptsize 2.93} & 
{\scriptsize 13.16} & {\scriptsize 7.10} & 
{\scriptsize 18.04} & {\scriptsize 9.14} & 
{\scriptsize 12.14} & {\scriptsize 6.39} \\
& DER++ &
{\scriptsize 4.03} & {\scriptsize 3.62} &
{\scriptsize 13.74} & {\scriptsize 6.31} & 
{\scriptsize 19.79} & {\scriptsize 7.11} &
{\scriptsize 12.52} & {\scriptsize 5.68} &\;\;\;\;&
{\scriptsize 4.51} & {\scriptsize 3.75} & 
{\scriptsize 12.15} & {\scriptsize 8.42} & 
{\scriptsize 16.85} & {\scriptsize 11.86} & 
{\scriptsize 11.17} & {\scriptsize 8.01} \\
& GMED &
{\scriptsize 4.17} & {\scriptsize 2.73} & 
{\scriptsize 12.01} & {\scriptsize 6.84} & 
{\scriptsize 18.95} & {\scriptsize 6.33} & 
{\scriptsize 11.71} & {\scriptsize 5.30} &\;\;\;\;&
{\scriptsize 4.71} & {\scriptsize \underline{2.27}} & 
{\scriptsize 12.83} & {\scriptsize 7.45} & 
{\scriptsize 18.44} & {\scriptsize 9.18} &
{\scriptsize 11.99} & {\scriptsize 6.30} \\
& CLS-ER &
{\scriptsize 4.61} & {\scriptsize 3.20} & 
{\scriptsize 14.07} & {\scriptsize 6.77} & 
{\scriptsize 19.54} & {\scriptsize 8.92} & 
{\scriptsize 12.74} & {\scriptsize 6.30} &\;\;\;\;&
{\scriptsize 4.17} & {\scriptsize 4.50} & 
{\scriptsize 11.28} & {\scriptsize 11.06} & 
{\scriptsize 16.85} & {\scriptsize 12.55} & 
{\scriptsize 10.77} & {\scriptsize 9.37} \\
& LUMP &
{\scriptsize 3.56} & {\scriptsize 2.79} & 
{\scriptsize 11.68} & {\scriptsize 7.65} & 
{\scriptsize 17.40} & {\scriptsize 8.52} & 
{\scriptsize 10.88} & {\scriptsize 6.32} &\;\;\;\;&
{\scriptsize 3.73} & {\scriptsize 3.03} & 
{\scriptsize 13.74} & {\scriptsize \textbf{5.29}} & 
{\scriptsize \underline{19.50}} & {\scriptsize \underline{8.17}} & 
{\scriptsize 12.32} & {\scriptsize \textbf{5.50}} \\
& ESMER &
{\scriptsize 4.51} & {\scriptsize 3.68} &
{\scriptsize 14.98} & {\scriptsize 6.22} &
{\scriptsize 21.25} & {\scriptsize 7.50} &
{\scriptsize 13.58} & {\scriptsize 5.80} &\;\;\;\;&
{\scriptsize 5.18} & {\scriptsize 4.92} &
{\scriptsize \underline{14.14}} & {\scriptsize 9.19} &
{\scriptsize 18.69} & {\scriptsize 12.84} &
{\scriptsize \underline{12.67}} & {\scriptsize 8.98} \\
\cmidrule{2-19}
& \cellcolor{gg} STELLA (Ours) &
{\scriptsize \cellcolor{gg}\underline{5.34}} & {\scriptsize \cellcolor{gg}\textbf{2.04}} &
{\scriptsize \cellcolor{gg}\underline{15.04}} & {\scriptsize \cellcolor{gg}\underline{5.20}} & 
{\scriptsize \cellcolor{gg}\underline{22.10}} & {\scriptsize \cellcolor{gg}\textbf{5.90}} & 
{\scriptsize \cellcolor{gg}\underline{14.16}} & {\scriptsize \cellcolor{gg}\textbf{4.38}} &\;\;\;\;&
{\scriptsize \cellcolor{gg}\underline{5.22}} & {\scriptsize \cellcolor{gg}\textbf{2.26}} & 
{\scriptsize \cellcolor{gg}13.09} & {\scriptsize \cellcolor{gg}7.95} & 
{\scriptsize \cellcolor{gg}18.75} & {\scriptsize \cellcolor{gg}10.65} & 
{\scriptsize \cellcolor{gg}12.35} & {\scriptsize \cellcolor{gg}6.95} \\
& \cellcolor{gg} STELLA+ (Ours) &
{\scriptsize \cellcolor{gg}\textbf{5.39}} & {\scriptsize \cellcolor{gg}\underline{2.71}} &
{\scriptsize \cellcolor{gg}\textbf{16.76}} & {\scriptsize \cellcolor{gg}\textbf{5.15}} & 
{\scriptsize \cellcolor{gg}\textbf{24.18}} & {\scriptsize \cellcolor{gg}\underline{5.99}} & 
{\scriptsize \cellcolor{gg}\textbf{15.44}} & {\scriptsize \cellcolor{gg}\underline{4.62}} &\;\;\;\;&
{\scriptsize \cellcolor{gg}\textbf{5.36}} & {\scriptsize \cellcolor{gg}4.24} & 
{\scriptsize \cellcolor{gg}\textbf{16.76}} & {\scriptsize \cellcolor{gg}\underline{5.54}} & 
{\scriptsize \cellcolor{gg}\textbf{23.65}} & {\scriptsize \cellcolor{gg}\textbf{7.44}} & 
{\scriptsize \cellcolor{gg}\textbf{15.26}} & {\scriptsize \cellcolor{gg}\underline{5.74}} \\
\cmidrule{2-19}

& Multitask &
{\scriptsize 6.45} & $-$ & 
{\scriptsize 20.19} & $-$ & 
{\scriptsize 29.01} & $-$ & 
{\scriptsize 18.55} & $-$ &\;\;\;\;&
{\scriptsize 8.28} & $-$ & 
{\scriptsize 24.14} & $-$ & 
{\scriptsize 33.74} & $-$ & 
{\scriptsize 22.05} & $-$ \\
\midrule

\parbox[t]{2mm}{\multirow{11}{*}{\rotatebox[origin=c]{90}{Video-to-Audio}}}
& Finetune & 
{\scriptsize 1.22} & {\scriptsize 4.47} & 
{\scriptsize 4.17} & {\scriptsize 11.23} & 
{\scriptsize 6.95} & {\scriptsize 14.67} & 
{\scriptsize 4.11} & {\scriptsize 10.12} &\;\;\;\;&
{\scriptsize 1.50} & {\scriptsize \textbf{3.23}} & 
{\scriptsize 4.08} & {\scriptsize 10.04} & 
{\scriptsize 6.33} & {\scriptsize 14.43} & 
{\scriptsize 3.97} & {\scriptsize 9.23} \\
& ER~ &
{\scriptsize 3.28} & {\scriptsize 3.94} & 
{\scriptsize 11.30} & {\scriptsize 8.86} & 
{\scriptsize 16.40} & {\scriptsize 11.37} & 
{\scriptsize 10.33} & {\scriptsize 8.06} &\;\;\;\;&
{\scriptsize 3.70} & {\scriptsize 4.36} & 
{\scriptsize 10.76} & {\scriptsize 10.34} & 
{\scriptsize 15.68} & {\scriptsize 15.06} & 
{\scriptsize 10.05} & {\scriptsize 9.92} \\
& MIR~ &
{\scriptsize 3.54} & {\scriptsize 3.47} & 
{\scriptsize 11.82} & {\scriptsize 9.11} & 
{\scriptsize 16.69} & {\scriptsize 12.90} & 
{\scriptsize 10.68} & {\scriptsize 8.49} &\;\;\;\;&
{\scriptsize 4.26} & {\scriptsize 4.59} & 
{\scriptsize 11.29} & {\scriptsize 9.87} & 
{\scriptsize 15.97} & {\scriptsize 13.73} & 
{\scriptsize 10.51} & {\scriptsize 9.40} \\
& DER++ &
{\scriptsize 3.49} & {\scriptsize 3.86} & 
{\scriptsize 13.22} & {\scriptsize 7.09} & 
{\scriptsize 19.03} & {\scriptsize 9.04} &
{\scriptsize 11.91} & {\scriptsize 6.66} &\;\;\;\;&
{\scriptsize 4.23} & {\scriptsize 4.50} & 
{\scriptsize 11.66} & {\scriptsize 10.10} & 
{\scriptsize 16.24} & {\scriptsize 13.97} & 
{\scriptsize 10.71} & {\scriptsize 9.52} \\
& GMED &
{\scriptsize 3.71} & {\scriptsize 2.61} & 
{\scriptsize 11.87} & {\scriptsize 6.46} & 
{\scriptsize 17.20} & {\scriptsize 9.57} & 
{\scriptsize 10.93} & {\scriptsize 6.21} &\;\;\;\;&
{\scriptsize 3.99} & {\scriptsize 4.42} & 
{\scriptsize 10.65} & {\scriptsize 10.39} & 
{\scriptsize 15.41} & {\scriptsize 14.78} & 
{\scriptsize 10.02} & {\scriptsize 9.86} \\
& CLS-ER &
{\scriptsize 4.09} & {\scriptsize 3.11} & 
{\scriptsize 13.30} & {\scriptsize 6.96} & 
{\scriptsize 19.43} & {\scriptsize 9.68} & 
{\scriptsize 12.27} & {\scriptsize 6.58} &\;\;\;\;&
{\scriptsize 4.25} & {\scriptsize 4.58} & 
{\scriptsize 9.78} & {\scriptsize 11.65} & 
{\scriptsize 13.45} & {\scriptsize 17.65} & 
{\scriptsize 9.16} & {\scriptsize 11.29} \\
& LUMP &
{\scriptsize 3.24} & {\scriptsize 3.30} &
{\scriptsize 11.02} & {\scriptsize 7.55} & 
{\scriptsize 16.91} & {\scriptsize 9.13} & 
{\scriptsize 10.39} & {\scriptsize 6.66} &\;\;\;\;&
{\scriptsize 3.13} & {\scriptsize 3.91} & 
{\scriptsize 10.60} & {\scriptsize \underline{8.63}} & 
{\scriptsize 16.02} & {\scriptsize \underline{12.26}} & 
{\scriptsize 9.92} & {\scriptsize \underline{8.27}} \\
& ESMER &
{\scriptsize 4.65} & {\scriptsize 2.74} &
{\scriptsize 14.54} & {\scriptsize 6.27} &
{\scriptsize 20.80} & {\scriptsize 8.36} &
{\scriptsize 13.33} & {\scriptsize 5.79} &\;\;\;\;&
{\scriptsize 4.39} & {\scriptsize 4.92} &
{\scriptsize 11.55} & {\scriptsize 12.16} &
{\scriptsize 16.41} & {\scriptsize 16.41} &
{\scriptsize 10.78} & {\scriptsize 11.16} \\
\cmidrule{2-19}
& \cellcolor{gg}STELLA (Ours)&
\cellcolor{gg}{\scriptsize \underline{5.30}} & \cellcolor{gg}{\scriptsize \underline{2.40}} &
\cellcolor{gg}{\scriptsize \underline{15.43}} & \cellcolor{gg}{\scriptsize \underline{4.84}} & 
\cellcolor{gg}{\scriptsize \underline{21.47}} & \cellcolor{gg}{\scriptsize \underline{6.70}} & 
\cellcolor{gg}{\scriptsize \underline{14.07}} & \cellcolor{gg}{\scriptsize \underline{4.65}} &\;\;\;\;&
\cellcolor{gg}{\scriptsize \underline{4.49}} & \cellcolor{gg}{\scriptsize \underline{3.39}} & 
\cellcolor{gg}{\scriptsize \underline{12.08}} & \cellcolor{gg}{\scriptsize 9.00} & 
\cellcolor{gg}{\scriptsize \underline{17.31}} & \cellcolor{gg}{\scriptsize 12.75} & 
\cellcolor{gg}{\scriptsize \underline{11.29}} & \cellcolor{gg}{\scriptsize 8.38} \\
& \cellcolor{gg}STELLA+ (Ours)&
\cellcolor{gg}{\scriptsize \textbf{5.86}} & \cellcolor{gg}{\scriptsize \textbf{1.56}} &
\cellcolor{gg}{\scriptsize \textbf{17.21}} & \cellcolor{gg}{\scriptsize \textbf{4.09}} & 
\cellcolor{gg}{\scriptsize \textbf{23.53}} & \cellcolor{gg}{\scriptsize \textbf{6.02}} & 
\cellcolor{gg}{\scriptsize \textbf{15.53}} & \cellcolor{gg}{\scriptsize \textbf{3.89}} &\;\;\;\;&
\cellcolor{gg}{\scriptsize \textbf{5.48}} & \cellcolor{gg}{\scriptsize 4.06} & 
\cellcolor{gg}{\scriptsize \textbf{15.65}} & \cellcolor{gg}{\scriptsize \textbf{7.13}} & 
\cellcolor{gg}{\scriptsize \textbf{22.29}} & \cellcolor{gg}{\scriptsize \textbf{8.92}} & 
\cellcolor{gg}{\scriptsize \textbf{14.47}} & \cellcolor{gg}{\scriptsize \textbf{6.70}} \\
\cmidrule{2-19}

& Multitask &
{\scriptsize 6.85} & $-$ & 
{\scriptsize 21.93} & $-$ & 
{\scriptsize 30.63} & $-$ & 
{\scriptsize 19.80} & $-$ &\;\;\;\;&
{\scriptsize 8.05} & $-$ & 
{\scriptsize 25.81} & $-$ & 
{\scriptsize 35.60} & $-$ & 
{\scriptsize 23.15} & $-$ \\

\bottomrule
\end{tabular}}
\label{tab:retrieval_table}
% \vspace{-0.1in}
\end{table*}

\subsection{Multimodal Patch Selection for Continual Learning \label{sec:subsec:patch selection}}

Leveraging the importance score $\bm{I}_{v}$ and correlation score $\bm{C}_{v}$, we enhance multimodal alignment and stability of the continual pre-training by sorting video patch indices. Initially, a Bernoulli distribution on $\bm{C}_{v}$ produces $\bm{F}_{v}$. True values in $\bm{F}_{v}$ indicate that the corresponding patches are chosen to be excluded. Hence, we zero out elements in $\bm{I}_{v}$ aligned with the True values in $\bm{F}_{v}$ to create $\tilde{\bm{I}}_{v}$. Subsequently, applying a multinomial probability distribution to $\tilde{\bm{I}}_{v}$ yields the informative video patch indices $\tilde{\bm{S}}_{v}\in\mathbb{R}^{B \!\times\! N}$:
\vspace{-0.025in}
\begin{align}
    \begin{split}
    \tilde{\bm{I}}_{v}[i,j]&=
    \begin{cases}
        0 & \text{if } \bm{F}_{v}[i,j]\;\;\;i\!=\!1,\ldots,B \\
        \bm{I}_{v}[i,j] & \text{otherwise}\;\;\;j\!=1,\ldots,N,
    \end{cases} \\
    \tilde{\bm{S}}_{v}&=\texttt{Multinomial}\left(\tilde{\bm{I}}_{v}, \right).
    \label{eq:video_patch_select}
    \end{split}
\end{align}
\vspace{-0.025in}
Similarly, we utilize the importance score $\bm{I}_{a}$ and correlation score $\bm{C}_{a}$ to generate the informative audio patch indices. To preserve the local correlation among audio patches by temporal continuity, we segment $\bm{I}_{a}$ into time chunks. To this end, we reshape the importance score $\bm{I}_{a}$ into a time-frequency dimension, average along the frequency dimension, and split the time dimension with time chunk size $\bm{L}_c$. This operation yields $\bm{I}^{c}_{a}\!\in\!\mathbb{R}^{B \!\times\!\left| t / p \right| / \left| \bm{L}_c \right|}$, which indicates the importance of audio time chunks. For $\bm{C}_{a}$, we apply Bernoulli probability distribution to generate $\bm{F}_{a}$.

We select informative time chunks with high $\bm{I}^{c}_{a}$ values while excluding the indices aligned with True values in $\bm{F}_{a}$ to generate the informative audio patch indices $\tilde{\bm{S}}_{a}\in\mathbb{R}^{B \!\times\! M}$. The detailed steps of audio patch selection are in ~\Cref{alg:audio_time_chunk_select_pytorch}.

Finally, based on $\tilde{\bm{S}}_{a},\tilde{\bm{S}}_{v}$, we select $\kappa_a,\kappa_v$ of audio, video patches to form new input $(\widehat{X}_{a},\widehat{X}_{v})$. Substituting $(X_{a},X_{v})$ into $(\widehat{X}_{a},\widehat{X}_{v})$ enables the model to effectively learn new audio-video relationships while preserving previously learned ones with enhanced efficiency. The final patch selection is performed as follows:
\begin{align}
\hspace*{-3mm}
    \begin{split}
    \widehat{X}_{n}[i,j]&\!=\!X_{n}[i,\tilde{\bm{S}}_{n}[i,j]],\;
    i\!=\!1,\ldots,B,\;j\!=\!1,\ldots,\kappa_n,
    \label{eq:final_patch_select}
    \end{split}
\hspace*{-3mm}
\end{align}
where $n\!\in\!(a,v)$. With the selected patches, we perform continual pre-training based on the \textit{DER++} framework with the penalty loss ($\ell^{p}$), which encourages the model to maintain the features of the rehearsal memory to mitigate their drifts. Hence, our final pre-training objective is $\mathcal{L}=\ell^{r}+\lambda \ell^{c}+\alpha \ell^{p}$, where $\alpha$ is a hyperparameter for the penalty loss.

Efficient rehearsal memory usage is crucial especially in continual audio-video learning scenarios due to the large video sizes. The effective storage of past data can notably augment the diversity of data within the memory. To address this, we propose \textit{STELLA+}, an extension of \textit{STELLA}, where memory stores the selected patches instead of raw data (\Cref{alg:stella_p_algo}). The introduction of \textit{STELLA+} represents a distinct and complementary direction to \textit{STELLA}, demonstrating the efficacy of efficient memory utilization.

\section{Experiments} \label{sec:experiments}

In this section, we experimentally validate the effectiveness of our method in task-free continual audio-video pre-training. We start by outlining our experimental setup in~\Cref{sec:exp:experiment setup}, covering datasets, evaluation methods, evaluation metrics, and baseline methods employed for our experiments. Subsequently, we present the experimental results and conduct a comprehensive analysis in~\Cref{sec:exp:experimental result}.

\subsection{Experimental Setup \label{sec:exp:experiment setup}}
\paragraph{Evaluation Protocol} We validate our method on continual audio-video pre-training over VGGSound~\citep{vggsound} and AudioSet~\citep{audioset} datasets, consisting of 10s videos. We split each dataset into multiple tasks based on its high-level category information. We name them as \textit{Continual-VS} and \textit{Continual-AS}, respectively. For evaluation, we conduct various audiovisual downstream tasks: retrieval, sound source localization, and event localization. Further details, including data split, data statistics, and downstream tasks, are provided in~\Cref{sec:supple:CL evaluation protocol}.

\paragraph{Baselines} To quantitatively assess our method, we compare its performance with several task-free continual learning methods: ER~\citep{Rolnick2019ER}, MIR~\citep{Aljundi2019MIR}, DER++~\citep{Buzzega2020DER}, GMED~\citep{Jin2021GMED}, CLS-ER~\citep{Arani2022clser}, LUMP~\citep{Madaan2022lump}, and ESMER~\citep{Sarfraz2023esmer}. The details of the baseline methods are explicated in~\Cref{sec:supple:Implementation Details}. All methods employ reservoir sampling~\citep{Vitter1985reservoir} to sample past instances from the rehearsal memory for $2K$ (\textit{Continual-VS}) and $5K$ (\textit{Continual-AS}) instances during continual pre-training, except for \textit{STELLA+}, which adjusts instance count based on sampling ratios ($\rho_{a}, \rho_{v}$) to match the memory size of other methods. We additionally report the result of \textit{Finetune}, the model continually pre-trained without additional methods, and \textit{Multitask}, the model pre-trained with the entire datasets. They serve as lower and upper bounds, respectively, in assessing learned representation.

\paragraph{Evaluation Metrics} After each end of pre-training on $\mathcal{D}_{t}$, we estimate task-specific performances $\{\textit{acc}_{t,i}\}^{t}_{i=1}$, where $\textit{acc}_{t,i}$ denotes the performance of the downstream task associated with $\mathcal{D}_{i}$ when evaluated with $f_{\theta,t}$, the model pre-trained up to the $t$-th task. Here, no task boundary information is employed in performance estimation. For the evaluation, we adopt two conventional metrics in continual learning: \textbf{(1) Average accuracy}($\mathcal{A}$) is the mean accuracy across all tasks after the completion of pre-training on $\mathcal{D}_{\mathcal{T}}$, and it is formulated as $\mathcal{A}\!=\!\frac{1}{\mathcal{T}}\sum^{\mathcal{T}}_{i=1}\textit{acc}_{\mathcal{T},i}$. \textbf{(2) Average Forgetting}($\mathcal{F}$) measures the average amount of catastrophic forgetting for each task, quantified as the difference between its maximum accuracy and accuracy at the completion of pre-training on $\mathcal{D}_{\mathcal{T}}$, calculated as, $\mathcal{F}\!=\!\frac{1}{\mathcal{T}-1}\sum^{\mathcal{T}-1}_{i=1}\!\underset{t\in\{1,\ldots, \mathcal{T}-1\}}{\text{max}}\!\left(\textit{acc}_{t,i}-\textit{acc}_{\mathcal{T},i}\right)$.

\subsection{Analysis for Continual Audio-Video Pre-training \label{sec:exp:experimental result}}
\begin{figure*}[t]
\centering
\begin{minipage}{0.51\linewidth}
\centering
\resizebox{\linewidth}{!}{%
    \begin{tabular}{cc}
    % \hspace{-0.1in}
        \includegraphics[width=0.5\linewidth]{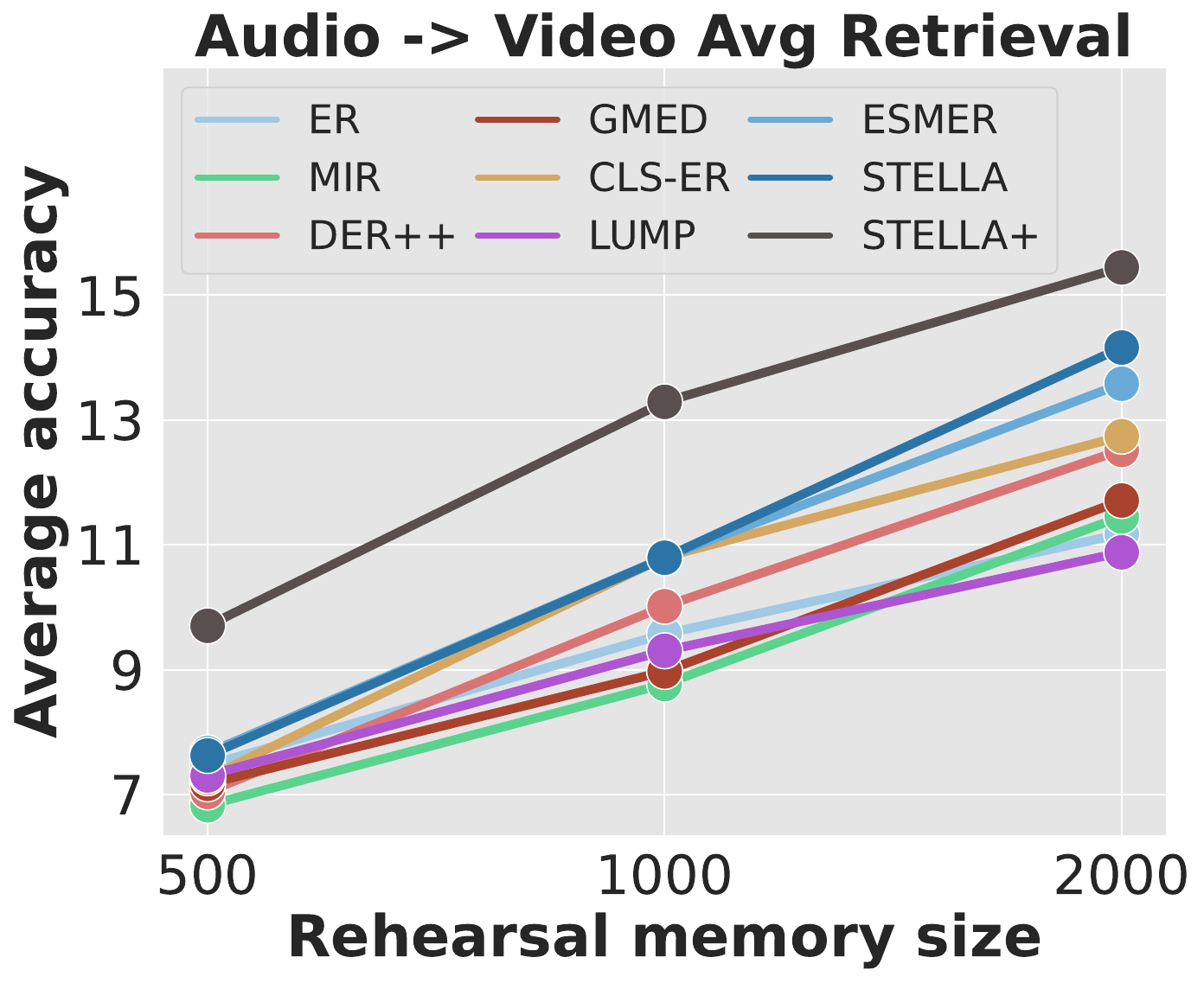}&
        \hspace{-0.2in}
        \includegraphics[width=0.5\linewidth]{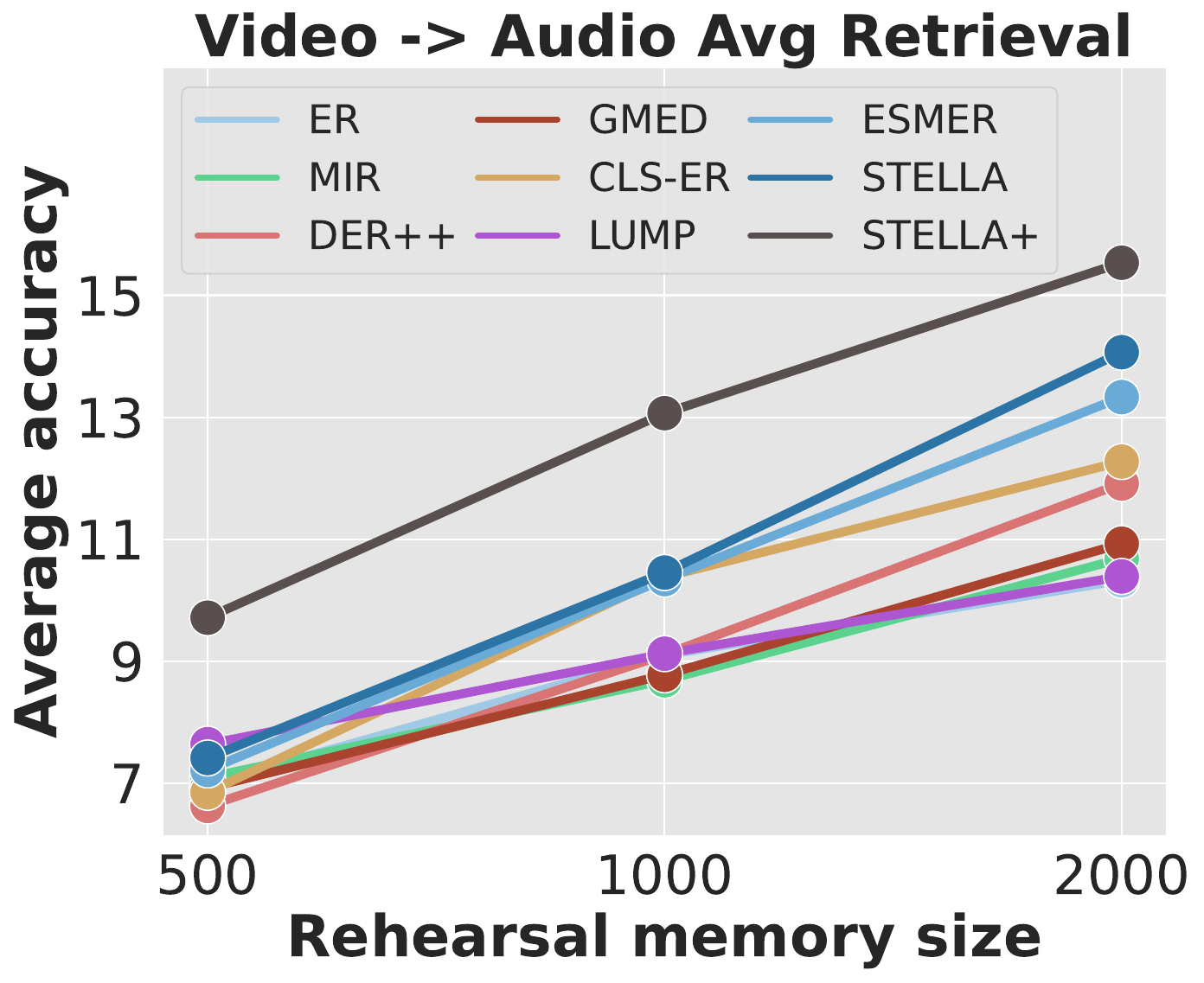}
        \end{tabular}
    \hspace{-0.1in}
}
% \vspace{-0.15in}
\caption{\textbf{Downstream performance on various rehearsal memory sizes.} We evaluate downstream task performances on the pre-trained models with various rehearsal memory sizes on the \textit{Continual-VS}.}
\label{fig:memory_size}

\centering
\hspace{-0.2in}
\begin{minipage}{0.62\linewidth}
    \centering
    \begin{tabular}{c}
        \includegraphics[width=\linewidth]{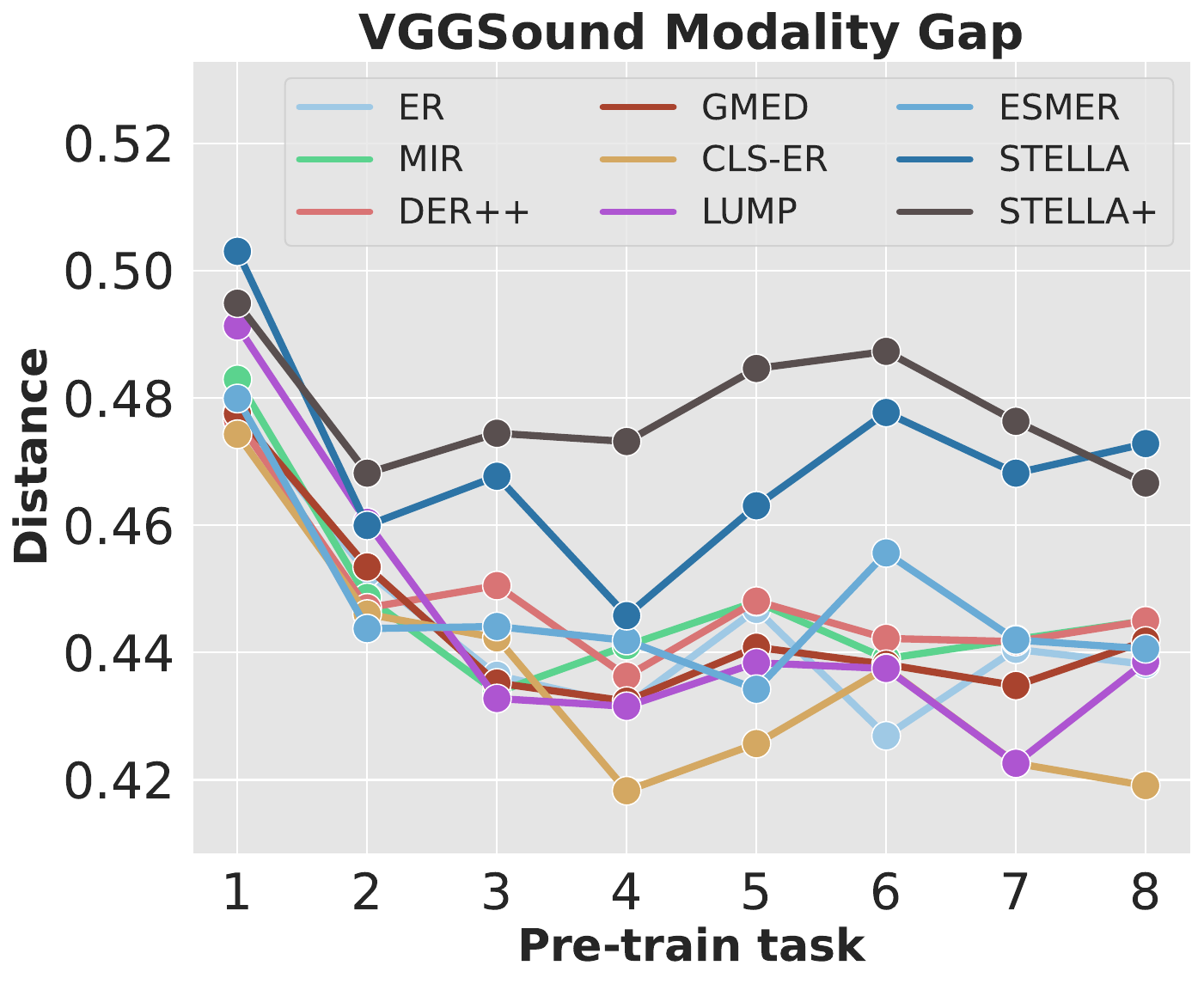}
    \end{tabular}
\end{minipage}
\begin{minipage}{0.38\linewidth}
    \centering
    \begin{tabular}{c}
        \includegraphics[width=\linewidth]{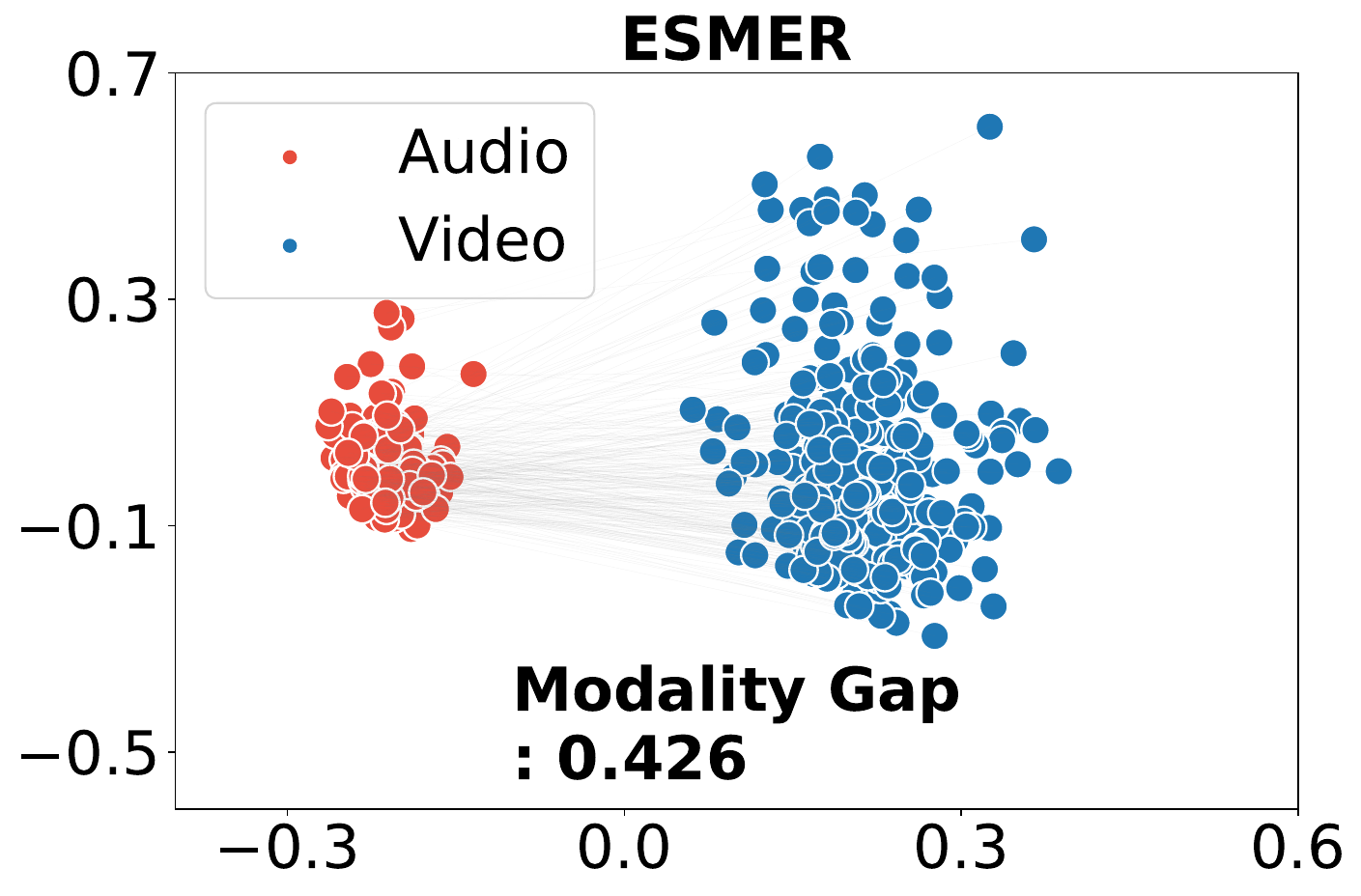}
    \end{tabular}
    \vspace{0.025in}
        \centering
    \begin{tabular}{c}
        \includegraphics[width=\linewidth]{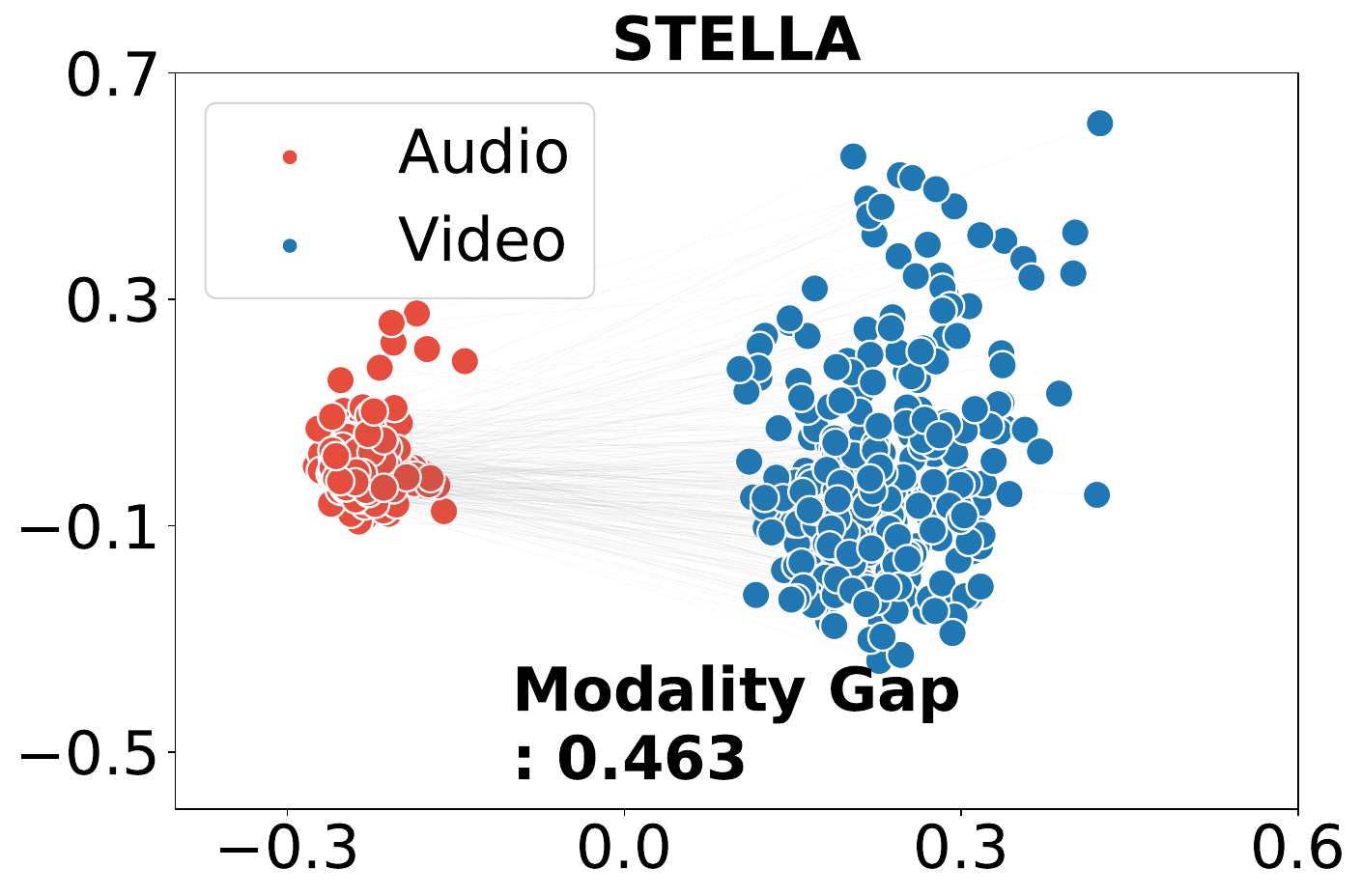}
    \end{tabular}
\end{minipage}
\hspace{-0.1in}
\vspace{-0.1in}
\caption{\textbf{Modality gap estimation.} \textbf{(Left)}: Estimation of modality gap after the completion of each task. (\textit{Continual-VS}) \textbf{(Right)}: Visualizations of modality gap corresponding to the music task with the model pre-trained up to the last task in the \textit{Continual-VS} dataset with \textit{ESMER} (top) and our method (bottom).}
\label{fig:modality_gap}

\end{minipage}
\hspace{0.1in}
\begin{minipage}{0.46\linewidth}
\input{materials/figures/5_sound_source_localization}
\end{minipage}
% \vspace{-0.15in}
\end{figure*}

\textbf{STELLA achieves \evblue{superior Zero-shot Audiovisual Retrieval performance}} compared to strong baselines. We perform audio-to-video and video-to-audio zero-shot retrieval tasks in \textit{Continual-VS} and \textit{Continual-AS} to quantitatively assess the learned audio-video correlation from the continual pre-training (\Cref{tab:retrieval_table}). For the \textit{Continual-VS}, both \textit{STELLA} and \textit{STELLA+} outperform other baselines, exhibiting substantial enhancements of 0.58\%p, 1.86\%p and 0.74\%p, 2.20\%p in average audio-to-video and video-to-audio retrieval scores, respectively. In the \textit{Continual-AS}, \textit{STELLA+} exhibits prominent performance advantages, with  2.59\%p and 3.69\%p improvements in average audio-to-video and video-to-audio retrieval scores. Notably, our methods consistently achieve high R@1 scores across all tasks. These results imply that our approach of continually pre-training on the selected patches enhances the model's ability to comprehend the audio-video relationship by accurately capturing sparse spatio-temporal correlations. For a thorough investigation, we conduct further experiments with shuffled task orders in~\Cref{sec:supple:Additional Experimental Results}. We also explore the influence of rehearsal memory size on zero-shot task performances, presenting the results in~\Cref{fig:memory_size}. Our methods consistently surpass other baselines, underscoring their effectiveness in adapting to diverse memory constraints.
\begin{table}[t]
\centering
% \vspace{-0.05in}
\begin{minipage}{0.48\textwidth}
\centering
\tiny
\caption{\textbf{Efficiency analysis.} GPU memory occupancy (GPU M.) is measured in GB. Throughput (T.P.) is measured in sample/sec. Both are estimated in single V100 with a batch size of 15 for STELLA++ and 9 for other methods.}
% \vspace{-0.05in}
\resizebox{\linewidth}{!}{%
\renewcommand{\arraystretch}{1.05}
\renewcommand{\tabcolsep}{3pt}
\begin{tabular}{l cc cc c c}
    \toprule
    \textbf{Method} & \multicolumn{2}{c}{\textbf{A$\rightarrow$V}} & \multicolumn{2}{c}{\textbf{V$\rightarrow$A}} & \textbf{GPU M.$\downarrow$} & \textbf{T.P.$\uparrow$}\\
    & $\mathcal{A} \uparrow$ & $\mathcal{F} \downarrow$ & $\mathcal{A} \uparrow$ & $\mathcal{F} \downarrow$ & & \\
    \midrule
    Finetune & 3.63 & 10.50 & 4.11 & 10.12 & 18.34 & \textbf{29.46} \\
    ER & 11.18 & 7.68 & 10.33 & 8.06 & 30.95 & 17.70 \\
    MIR & 11.45 & 7.55 & 10.68 & 8.49 & 31.17 & 5.73 \\
    DER++ & 12.52 & 5.68 & 11.91 & 6.66 & 30.95 & 17.79 \\
    GMED & 11.71 & 5.30 & 10.93 & 6.21 & 32.03 & 5.63 \\
    CLS-ER & 12.74 & 6.30 & 12.27 & 6.58 & 32.50 & 15.24 \\
    LUMP & 10.88 & 6.32 & 10.39 & 6.66 & 18.36 & \underline{26.67} \\
    ESMER & 13.58 & 5.80 & 13.33 & 5.79 & 31.45 & 14.88 \\
    \midrule
    \cellcolor{gg}STELLA (Ours) & \cellcolor{gg}14.16 & \cellcolor{gg}\underline{4.38} & \cellcolor{gg}14.07 & \cellcolor{gg}4.65 & \cellcolor{gg}\underline{17.45} & \cellcolor{gg}17.29 \\
    \cellcolor{gg}STELLA+ (Ours) & \cellcolor{gg}\underline{15.44} & \cellcolor{gg}4.62 & \cellcolor{gg}\underline{15.53} & \cellcolor{gg}\underline{3.89} & \cellcolor{gg}\textbf{17.15} & \cellcolor{gg}18.11 \\
    \cellcolor{gg}STELLA++ (Ours) & \cellcolor{gg}\textbf{17.01} & \cellcolor{gg}\textbf{3.20} & \cellcolor{gg}\textbf{16.62} & \cellcolor{gg}\textbf{3.27} & \cellcolor{gg}24.69 & \cellcolor{gg} - \\
    \bottomrule
\end{tabular}}
\label{tab:sub:efficiency}
\end{minipage}%
\par
\vspace{0.1in}
\begin{minipage}{0.48\textwidth}
\centering
\tiny
\caption{\textbf{Sampling methods.} Experiments with various sampling methods. LPIS: Localized Patch Importance Scoring in~\Cref{sec:subsec:positive region proposal}, RCA: Replay-guided Correlation Assessment in~\Cref{sec:subsec:forget-robust selection}.}
% \vspace{-0.05in}
\resizebox{\linewidth}{!}{%
\renewcommand{\arraystretch}{1.15}
\renewcommand{\tabcolsep}{2pt}
\begin{tabular}{lcc cc cc c}
    \toprule
    \textbf{Method} & \textbf{LPIS} & \textbf{RCA} & \multicolumn{2}{c}{\textbf{A$\rightarrow$V}} & \multicolumn{2}{c}{\textbf{V$\rightarrow$A}} & \textbf{GPU M.$\downarrow$} \\
    & & & $\mathcal{A} \uparrow$ & $\mathcal{F} \downarrow$ & $\mathcal{A} \uparrow$ & $\mathcal{F} \downarrow$ & \\
    \midrule
        Random & $-$ & $-$ & 12.64 & 6.46 & 12.55 & 6.58 & \textbf{16.63} \\ 
    MATS & $-$ & $-$ & 12.91 & 6.55 & 12.70 & 6.80 & 21.30 \\
    \midrule
    \multirow{4}{*}{{STELLA (Ours)}} & $-$ & $-$ & 12.52 & 5.68 & 11.91 & 6.66 & 30.95 \\
    & $\checkmark$ & $-$ & 13.44 & 5.50 & 13.27 & 5.94 & 17.48 \\
    & $-$ & $\checkmark$ & 13.40 & 5.30 & 12.94 & 5.44 & 17.48 \\
    & $\checkmark$ & $\checkmark$ & \textbf{14.16} & \textbf{4.38} & \textbf{14.07} & \textbf{4.65} & 17.45 \\
    \bottomrule
\end{tabular}}
\label{tab:sub:sampling method}
\end{minipage}
\par
% \vspace{-0.1in}
\end{table}

\begin{table*}[t]
    \centering    
    \caption{\textbf{Audiovisual downstream tasks.} We finetune models continually pre-trained on \textit{Continual-VS} tasks. \textbf{(a):} Finetuning with the MSR-VTT~\citep{Xu2016msrvtt} train dataset, we measure audiovisual retrieval performance. \textbf{(b):} We randomly initialize and finetune a MLP classifier, attached on the top of the models, using the entire \textit{Continual-VS} dataset. \textbf{(c):} We finetune a randomly initialized decoder with the AVSBench~\citep{zhou2022avseg} training dataset. MIoU (Mean Intersection over Union) measures the average overlap between predicted segments and ground truth segments. The best and the second best results are highlighted in \textbf{bold} and \underline{underline}, respectively.}
    \label{tab:downstream_tasks}
    \begin{minipage}{0.48\linewidth}
    \centering
    \small{\textbf{(a) MSR-VTT audiovisual retrieval}}\\
        \vspace{0.05in}
        \centering
\tiny
\resizebox{\linewidth}{!}{%
\renewcommand{\arraystretch}{1.0}
\renewcommand{\tabcolsep}{3.8pt}
\begin{tabular}{l c c c c c c c c}
    \toprule
    \textbf{Method} & \multicolumn{3}{c}{\textbf{A$\rightarrow$V}} & \multicolumn{3}{c}{\textbf{V$\rightarrow$A}} \\
    & \textbf{R@1} & \textbf{R@5} & \textbf{R@10} & \textbf{R@1} & \textbf{R@5} & \textbf{R@10} \\
    \midrule
    Finetune & 1.00 & 4.15 & 6.44 & 1.33 & 3.19 & 6.15 \\
    ER & 2.26 & 7.89 & 13.38 & 2.26 & 8.78 & 13.42 \\
    MIR & 2.48 & 7.59 & 11.89 & 1.85 & 7.37 & 11.81 \\
    DER++ & 1.93 & 8.23 & 13.75 & 2.52 & 8.30 & 13.42 \\
    GMED & 1.67 & 6.81 & 11.81 & 1.44 & 6.04 & 11.59 \\
    CLS-ER & 2.15 & 8.45 & 12.93 & 2.15 & 7.63 & 12.82 \\
    LUMP & 1.78 & 7.70 & 12.07 & 1.59 & 7.04 & 11.81 \\
    ESMER & 2.33 & 8.37 & 13.78 & 2.30 & 8.48 & 13.93 \\
    \midrule
    \cellcolor{gg}STELLA (Ours) & \cellcolor{gg} \textbf{2.70} & \cellcolor{gg} \underline{8.70} & \cellcolor{gg} \underline{13.96} & \cellcolor{gg} \textbf{2.67} & \cellcolor{gg} \underline{8.81} & \cellcolor{gg} \underline{14.30} \\
    \cellcolor{gg}STELLA+ (Ours) & \cellcolor{gg} \underline{2.37} & \cellcolor{gg} \textbf{9.11} & \cellcolor{gg} \textbf{15.07} & \cellcolor{gg} \underline{2.44} & \cellcolor{gg} \underline{10.14} & \cellcolor{gg} \textbf{15.62} \\
    \bottomrule
\end{tabular}}
        \par
        % \small{\textbf{(a) MSR-VTT audiovisual retrieval}}
    \end{minipage}
    \begin{minipage}{0.24\linewidth}
    \centering
    \small{\textbf{(b) Audiovisual classification}}\\
        \vspace{0.05in}
        \centering
\tiny
\resizebox{1.0\textwidth}{!}{
    \renewcommand{\arraystretch}{0.9}
    \renewcommand{\tabcolsep}{3.8pt}
    \begin{tabular}{l c}
        \toprule
        {\textbf{Method}} & Accuracy \\
        \midrule
        Finetune & {57.04} \\
        ER & {57.09} \\
        MIR & {56.82} \\
        DER++ & {57.23} \\
        GMED & {57.34} \\
        CLS-ER & {57.23} \\
        LUMP & {57.70} \\
        ESMER & {57.72} \\
        \cmidrule{1-2}
        \cellcolor{gg}STELLA (Ours)&
        {\cellcolor{gg}\underline{58.20}} \\
        \cellcolor{gg}STELLA|+ (Ours)&
        {\cellcolor{gg}\textbf{58.54}} \\
        \cmidrule{1-2}
        Multitask & {59.94} \\
        \bottomrule
    \end{tabular}
}
        \par
        % \small{\textbf{(b) Audiovisual classification}}
    \end{minipage}
    \begin{minipage}{0.24\linewidth}
    \centering
        \small{\textbf{(c) Audiovisual segmentation}}\\
        \vspace{0.05in}
        \centering
\tiny
\resizebox{1.0\textwidth}{!}{
    \renewcommand{\arraystretch}{0.9}
    \renewcommand{\tabcolsep}{5.8pt}
    \begin{tabular}{l c}
        \toprule
        {\textbf{Method}} & MIoU \\
        \midrule
        Finetune & {54.77} \\
        ER & {54.64} \\
        MIR & {54.69} \\
        DER++ & {55.42} \\
        GMED & {55.92} \\
        CLS-ER & {55.89} \\
        LUMP & {55.34} \\
        ESMER & {55.84} \\
        \cmidrule{1-2}
        \cellcolor{gg}STELLA (Ours)&
        {\cellcolor{gg}\underline{56.59}} \\
        \cellcolor{gg}STELLA|+ (Ours)&
        {\cellcolor{gg}\textbf{57.26}} \\
        \cmidrule{1-2}
        Multitask & {58.51} \\
        \bottomrule
    \end{tabular}
}
        \par
        % \small{\textbf{(c) Audiovisual segmentation}}
    \end{minipage}
\end{table*}

\textbf{STELLA is \evblue{significantly efficient in terms of GPU Memory Consumption and Throughput}}. Pre-training on the spatio-temporally aligned subset of audio-video patches also enhances efficiency. In~\Cref{tab:sub:efficiency}, we compare GPU memory occupancy and throughput across different methods. \textit{STELLA} consumes significantly less GPU memory than baselines, even surpassing \textit{Finetune} in efficiency. Compared to \textit{DER++}, \textit{STELLA+} achieves a 44.59\% gain in efficiency, further enhancing throughput. In order to explore the benefits of reduced GPU memory usage, we conduct experiments with \textit{STELLA+} with an increased batch size. Specifically, we increase the batch size by 1.66 times and denote this version of \textit{STELLA+} as \textit{STELLA++}. As shown in~\Cref{tab:sub:efficiency}, \textit{STELLA++} outperforms all baselines, including \textit{STELLA+}. We expect that increasing batch size for contrastive learning-based models enhances the model's ability to accurately distinguish between various inputs and increases stability during continual pre-training. In the case of rehearsal memory burden, the extra cost required in \textit{STELLA} for storing the queries, importance scores, and correlation scores in the memory is negligible (+~$0.16$ GB), based upon the fact that the size of the memory itself is $5.47$ GB and that \textit{CLS-ER} and \textit{ESMER} maintain additional models, which require +$~1.42$ GB and +$~0.71$ GB additional memory, respectively.

\textbf{\evblue{Core components in STELLA contribute} to improving evaluation performance}.
To validate our patch selection method, we compare our two core components with \textit{MATS}~\citep{Hwang2023mats}, an adaptive patch selection method aiming to discard redundant patches during video pre-training, and with a simple random patch selection method, denoted as \textit{Random}. We decompose \textit{STELLA} into Localized Patch Importance Scoring (\textit{LPIS}) and Replay-guided Correlation Assessment (\textit{RCA}). All the above methods follow the default sampling ratio and were built upon \textit{DER++}. In \textit{Continual-VS} zero-shot retrieval tasks, \textit{LPIS} and \textit{RCA} show competitive results against baselines including \textit{MATS} and \textit{Random} (\Cref{tab:sub:sampling method}). \textit{LPIS} enhances the model's audio-video semantics comprehension. Conversely, \textit{RCA} demonstrates more robustness in forgetting but with a lower average retrieval score, indicating a need for improved guidance in understanding audio-video semantics. Combining both components, \textit{STELLA} achieves improved performances, emphasizing the importance of considering both the sparse correlation and forgetting in continual audio-video pre-training.

\textbf{STELLA \evblue{excels in various audiovisual downstream tasks}}.
To evaluate the acquired transferable knowledge through continual audio-video pre-training, we perform diverse audiovisual downstream tasks. Compared to the earlier zero-shot retrieval tasks, we use the models that have been continually pre-trained up to the final task of \textit{Continaul-VS}, and then evaluate them on different audiovisual datasets. First, we conduct audiovisual retrieval experiments on the MSR-VTT~\citep{Xu2016msrvtt} dataset. We train the pre-trained models on the MSR-VTT training dataset according to the training objective in~\Cref{sec:subsec:problem_statement} and evaluate them on the MSR-VTT test dataset. As shown in~\Cref{tab:downstream_tasks} \highlight{(a)}, our methods consistently outperform the baselines, demonstrating that our methods excel at understanding relationships in audio-video pairs. Second, we perform audiovisual classification experiments on the entire \textit{Continual-VS} datasets with class labels. Specifically, we finetune a randomly initialized MLP classifier, which is attached to the top of the continually pre-trained models, using the datasets. Then, we test the models' classification performance using the evaluation datasets of \textit{Continual-VS}. This setup ensures that the classification results reflect the quality of audio-video representations learned throughout the continual audio-video pre-training process. Experimental results in~\Cref{tab:downstream_tasks} \highlight{(b)} demonstrate that our methods yield superior audio-video representations, leading to enhanced classification performance over baseline methods. This improvement is due to our approach's ability to identify patches with high audio-video correlation, thereby enhancing the model's comprehension of audio-video data during continual pre-training. Furthermore, we conduct audiovisual segmentation experiments. Following the experiments in~\citep{lin2023vision}, we finetune a randomly initialized decoder, attached on top of the continually pre-trained models, for the audiovisual segmentation task with the training dataset of the AVSBench~\citep{zhou2022avseg}, and test the performance on the AVSBench test dataset. The results, shown in~\Cref{tab:downstream_tasks} \highlight{(c)}, indicate that our methods surpass the baselines. This suggests that our pre-trained models have a superior multimodal ability to spatially localize sound sources given corresponding audio, demonstrating the efficacy of our continual pre-training approach. Finally, we perform a sound source localization task on the AVE~\citep{Tian2020avparsing} dataset to assess the model's ability to detect sound sources within visual scenes. As shown in~\Cref{fig:sound_source_localization}, given audio containing a barking dog, all methods struggle to precisely locate the sound source, concentrating on the uncorrelated object (green bottle) in the visual scene. In contrast, the AVM module in \textit{STELLA} stands out by precisely identifying the correct sound source, proving its efficacy in aligning multimodal data even in continual pre-training scenarios. This qualitative result further strengthens our quantitative evaluation of the audiovisual segmentation task in~\Cref{tab:downstream_tasks} \highlight{(c)}. More examples of the sound source localization task are illustrated in~\Cref{fig:supple_sound_source_localization}. Additional results for other audiovisual downstream tasks, including event localization and retrieval tasks, are available in~\Cref{sec:supple:Additional Experimental Results}.

\textbf{STELLA can \evblue{preserve the modality gap} between audio and video embeddings even after continual learning}.
Recent research in multimodal learning~\citep{Liang2022MindtheGap} reveals that embeddings cluster by modality in representation space. Such modality-dependent clustering behavior introduces the concept of modality gap, which refers to the distance between these clusters (\Cref{fig:modality_gap} \highlight{(Right)}). A larger modality gap is generally considered favorable under well-separated modality clusters since it indicates that the model can distinguish between different modalities effectively. Hence, in the context of continual audio-video pre-training, maintaining a large modality gap between the two modalities throughout tasks is desirable, as deviating from it suggests a departure from the optimal state. Hence, during continual pre-training, we estimate the modality gap at the end of each task, utilizing evaluation data of each task. The estimated modality gaps of baselines are presented in~\Cref{fig:modality_gap} \highlight{(Left)}. Our methods consistently maintain the highest modality gap compared to other approaches. Moreover, our methods exhibit small modality gap declines, indicating that the models suffer less from the forgetting of previous multimodal correlations, which supports the validity of our approach in preventing \textit{modality correlation overwriting} in~\Cref{sec:subsec:forget-robust selection} to address the issue of audio-video relation forgetting. \Cref{sec:supple:Modality Gap Analysis} provides more analysis using the modality gap including \textit{Continual-AS} and about two key components of our approach. Besides, some previous works~\citep{udandarao2022understanding} observe that reducing modality gaps also has benefits. Based on the modality gap analysis \citep{udandarao2022understanding}, there exists a modality gap that yields the best downstream task performances. However, we would like to emphasize that we use the modality to estimate the change in the modality gap throughout continual pre-training, not to find the best modality gap of the backbone model.

\section{Conclusion}
In this paper, we investigate the critical challenges in continual audio-video pre-training under the task-free scenario, where the model continuously learns a course of audio-video multimodal tasks sequentially and cannot access previous tasks and task oracle both on pre-training and fine-tuning. We empirically observe that the audio-video models suffer from the issue of sparse spatiotemporal correlation and representational forgetting of audio-video relationships. To overcome these limitations, we propose a novel continual audio-video multimodal pre-training method for the first time that adaptively captures sparse audio-video attention to learn accurate audio-video relationships while mitigating forgetting from previously learned relationships without requiring task identification.

\newpage

\section*{Impact Statement}
In this work, we suggest \textit{STELLA} and compare it with other recent baselines in continual audio-video pre-training scenarios. Both methods use rehearsal memories to store the subset of pre-train data from the sequence of tasks. Since the sampling process is random, all methods cannot effectively alleviate the problem of privacy issues when storing videos in the rehearsal memory. One potential way to alleviate the problem is to save the subset of audio and video patches as in \textit{STELLA+}. We sincerely hope that more effective ways to solve privacy issues in rehearsal memory will be investigated while maintaining the benefits of rehearsal-based continual learning methods.

\section*{Acknowledgements}
This work was supported by Institute for Information \& communications Technology Promotion(IITP) grant funded by the Korea government(MSIP) (No.2019-0-00075 Artificial Intelligence Graduate School Program(KAIST)), the National Research Foundation of Korea(NRF) grant funded by the Korea government(MSIT) (No. RS-2023-00256259), Google Research Grant and Google Cloud Research Credits program with the award (e8f24127-e549-4d6b-b5a7-1885b4d29d20) and KAIST-NAVER Hypercreative AI Center.

\bibliography{references}
\bibliographystyle{icml2024}

\newpage
\clearpage
\appendix
\onecolumn

\paragraph{\LARGE{Appendix}}

\paragraph{Organization} The supplementary file is organized as follows: First, we explain the implementation details for our experiments in~\Cref{sec:supple:Implementation Details}. Then, we outline the evaluation protocol of our experiments in~\Cref{sec:supple:CL evaluation protocol}. In~\Cref{sec:supple:av-self-sup objectives}, we elaborate on the audio-video self-supervised objectives used for pre-training the model. Additionally,~\Cref{sec:supple:AVM training} presents a detailed account of the training procedure for the AVM module. We provide additional experimental results in~\Cref{sec:supple:Additional Experimental Results}. \Cref{sec:supple:Hyperparameter Tuning Results} showcases the outcomes of our hyperparameter tuning process. Furthermore, in~\Cref{sec:supple:Modality Gap Analysis}, we conduct more analysis on our experimental results using the modality gap. We present PyTorch-like pseudo code for audio patch selection in~\Cref{sec:supple:audio_constraint_pytorch}. We provide STELLA and STELLA+ algorithms in~\Cref{sec:supple:stella_p_algo}. In~\Cref{sec:supple:fading audio-visual attention} we provide more examples of visualization that show challenges in audio-video lifelong pre-training. Finally, \Cref{sec:supple:limitations} outlines the limitations of our study.

\section{Implementation Details \label{sec:supple:Implementation Details}}
\paragraph{Hyperparameter configurations.} We referred to the original papers for initial settings of hyperparameters of continual learning methods. Based on the initial settings, we tune the hyperparameters for our continual audio-video representation learning. Searched hyperparameters are listed in~\Cref{tab:baseline_hyp}. In our method,  $\alpha$ denotes a multiplier for the penalty loss to minimize the distance between obtained logits from the buffer instances and their logits stored at the past timestep. We also listed our pre-training and fine-tuning hyperparameters in~\Cref{tab:training_hyp}.
\begin{table}[ht]
\centering
\caption{\textbf{Continual learning method hyperparameters.}}
\vspace{0.1in}
\resizebox{0.9\textwidth}{!}{\begin{tabular}{lll}
\toprule
METHOD  & Continual-VS    & Continual-AS \\ 
\midrule
ER & - & - \\
MIR & $C\;:\;5$ & $C\;:\;5$ \\
DER++ & $\alpha\;:\;0.5$ & $\alpha\;:\;1.0$ \\
GMED & $\alpha\;:\;0.1\;\beta\;:\;0.05\;\gamma\;:\;1.0$ & $\alpha\;:\;0.1\;\beta\;:\;0.01\;\gamma\;:\;1.0$ \\
CLS-ER & $\lambda\;:\;0.1\;\alpha_{S}\;:\;0.999\;\alpha_{P}\;:\;0.999\;r_{S}\;:\;0.6\;r_{P}\;:\;0.8$ & $\lambda\;:\;0.1\;\alpha_{S}\;:\;0.999\;\alpha_{P}\;:\;0.999\;r_{S}\;:\;0.6\;r_{P}\;:\;0.8$  \\
LUMP & $\lambda\;:\;0.1$ & $\lambda\;:\;0.05$  \\
ESMER & $\alpha_l\;:\;0.99\;\beta\;:\;1.0\;\gamma\;:\;0.15\;\alpha\;:\;0.999\;r\;:\;0.2$ & $\alpha_l\;:\;0.99\;\beta\;:\;1.0\;\gamma\;:\;0.2\;\alpha\;:\;0.999\;r\;:\;0.2$ \\
STELLA (Ours) & $\alpha\;:\;0.5\;\beta\;:\;0.4\;\rho_{a}\;:0.5\;\rho_{v}\;:0.5$ & $\alpha\;:\;0.5\;\beta\;:\;0.1\;\rho_{a}\;:0.5\;\rho_{v}\;:0.5$ \\
\bottomrule
\end{tabular}}
\label{tab:baseline_hyp}
\end{table}
\begin{table}[t]
\centering
\caption{\textbf{Audio-Video pre-training and fine-tuning hyperparameters.}}
\vspace{0.01in}
\resizebox{0.8\textwidth}{!}{
\renewcommand{\arraystretch}{1.2}
\renewcommand{\tabcolsep}{9pt}
\begin{tabular}{lccccccc}
\toprule
 & \multicolumn{2}{c}{Pretrain} & \multicolumn{4}{c}{Finetune} \\
\midrule
{Dataset} & {Continual-VS} & {Continual-AS} & {MSR-VTT} & {AVC} & {AVS} & {AVE} \\
\cmidrule{2-8}
Optimizer & \multicolumn{2}{c}{Adam} & \multicolumn{4}{c}{AdamW}\\
Optimizer momentum & \multicolumn{6}{c}{$\beta_1, \beta_2=0.95, 0.999$}\\
Learning rate & \multicolumn{3}{c}{1e-4} & 1e-4 & 5e-4 & 1e-3\\
Weight decay & \multicolumn{4}{c}{5e-7} & \multicolumn{2}{c}{5e-6}\\
Learning rate schedule & \multicolumn{2}{c}{-} & \multicolumn{4}{c}{CosineScheduler}\\
Warmup epochs & \multicolumn{4}{c}{-} & 3 & 2 \\
Epoch & 10 & 15 & 15 & 10 & 20 & 15 \\
Batch size & 48 & 36 & \multicolumn{3}{c}{48} & 12 \\
GPUs & \multicolumn{3}{c}{4 A100 or 4 V100} & \multicolumn{3}{c}{4 Titan X Pascal} \\
Audio Random Time Shifting & \multicolumn{4}{c}{yes} & \multicolumn{2}{c}{no} \\
Audio Random Noise & \multicolumn{4}{c}{yes} & \multicolumn{2}{c}{no} \\
Audio Norm Mean & \multicolumn{6}{c}{-5.081} \\
Audio Norm STD & \multicolumn{6}{c}{4.485} \\
Video MultiScaleCrop & \multicolumn{6}{c}{yes} \\
Video Norm Mean & \multicolumn{6}{c}{[0.485, 0.456, 0.406]} \\
Video Norm STD & \multicolumn{6}{c}{[0.229, 0.224, 0.225]} \\
\bottomrule
\end{tabular}}
\label{tab:training_hyp}
\end{table}

\paragraph{Baselines.} ER~\citep{Rolnick2019ER} employs rehearsal memory and learns the past data in the memory during training on the current task to mitigate forgetting. All the baselines below employ the rehearsal memory to store the subset of past data. MIR~\citep{Aljundi2019MIR} introduces a strategy that retrieves data the model is likely to forget during the current task and trains the model with the retrieved data. To retrieve the data, it pseudo-updates the model with the data in the current step and finds the mini-batch of past data that gives the highest training loss. DER++~\citep{Buzzega2020DER} matches stored logits in the rehearsal memory from past tasks with the current ones, ensuring a smoother transition and preventing abrupt changes in the logits during training. In our setting, we store both audio and video logits in the rehearsal memory and apply the method independently. GMED~\citep{Jin2021GMED} tackles forgetting by using gradient information to update past data in the rehearsal memory. The data is updated to maximize interference of the current task to help the model retain past knowledge. Hence, it virtually updates the model with data from the current step and calculates the relative gradient by the past data to update the past data. CLS-ER~\citep{Arani2022clser} draws inspiration from the complementary learning system theory and maintains two models to retain short-term memories and long-term memories; one quickly adapts to new tasks and the other is slowly updated to retrain past knowledge. The slowly updated model transfers retained knowledge to the adaptable one, ensuring the retention of past information. LUMP~\citep{Madaan2022lump} integrates past and current data by mixing the two data, rather than replaying the past data together with data from the current task to handle the forgetting issue. In our setting, we integrate the past and current video and audio respectively with the same ratio. Lastly, ESMER~\citep{Sarfraz2023esmer} employs a semantic memory model that has the same structure as the pre-trained model to slowly integrate the knowledge encoded in the weights. It refers to the memory model to alleviate the effect of the data from the current batch that induces abrupt drift in the learned representations in order to reduce forgetting. The suggested method effectively handles the abrupt representation changes when the data distribution shifts. 

\section{Continual pre-training evaluation protocol \label{sec:supple:CL evaluation protocol}}

\paragraph{Audiovisual Dataset Configuration}
In this section, we specify how we design our continual audio-video pre-training experiments using two benchmark datasets: VGGSound and AudioSet. To mimic the data distribution shift due to the new audio-video semantics described in~\Cref{sec:intro}, we split the dataset according to the high-level categories. For the VGGSound dataset, we split the dataset into eight tasks based on the category labels~\citep{vggsound}. Each task dataset consists of 6k-8k video clips from 20 different classes. We name it as \textit{\textit{Continual-VS}}. Then, we construct another pre-training dataset by combining the unused training dataset in VGGSound with the AudioSet-20k~\citep{audioset}, resulting in a total of 104k video clips. We took care to exclude the unused VGGSound video samples whose class labels are present in the \textit{Continual-VS}. Using the merged dataset, we pre-train the backbone weights before continual pre-training. This ensures that the model does not underperform at the initial continual pre-training stages while the model does not acquire any task-specific knowledge at the beginning. For the \textit{Continual-VS} continual pre-training, we follow the task sequence: sports$\rightarrow$music$\rightarrow$vehicle$\rightarrow$people$\rightarrow$animals$\rightarrow$home\&nature$\rightarrow$others part1(tools\&others)$\rightarrow$others part2(remaining others).

Similarly, we divided the AudioSet dataset into seven tasks, following class hierarchy information~\citep{audioset}. We name it as \textit{\textit{Continual-AS}}. Compared to \textit{Continual-VS}, it exhibits imbalanced dataset size among tasks and contains much larger clips. To ensure proper pre-training for the \textit{Continual-AS} experiments, we pre-train the model with the entire VGGSound dataset to avoid any potential performance issues during the initial stages of continual pre-training. We randomly shuffle the pre-train order and follow the task sequence: human$\rightarrow$vehicle$\rightarrow$nature$\rightarrow$animal$\rightarrow$others$\rightarrow$home$\rightarrow$music.

For downstream tasks, we use two audiovisual datasets: MSR-VTT~\citep{Xu2016msrvtt} and AVE~\citep{Tian2020avparsing}. MRS-VTT consists of 10,000 video clips from 20 different categories. We collect video clips that contain audio modality on both the training dataset and the test dataset. This yields $\sim6k$ and $\sim0.9k$ video clips, respectively. We finetune the continually pre-trained models on the MSR-VTT training dataset and evaluate on the test dataset to perform audiovisual bi-directional retrieval tasks. In the case of the AVE dataset, it contains $\sim4k$ videos with 28 different event categories. Since the dataset is a subset of AudioSet, we conduct experiments on the pre-trained models on \textit{Continaul-VS} only. With this dataset, we perform two downstream tasks: sound source localization, which requires the models to locate the sounding objects in the visual scene, and audiovisual event localization, which asks the model to classify audiovisual events for each time step. Given that all the downstream task datasets represent unseen data for the pre-trained models, they allow us to gauge the extent to which the model has acquired general knowledge of audio-video correlations during continual audio-video pre-training.

\paragraph{Audiovisual downstream task configuration}
When constructing audiovisual zero-shot retrieval tasks for model performance evaluation, we refer to the CAV~\citep{Yuan2023cav} for both the \textit{Continual-VS} and \textit{Continual-AS} experiments. We employ the zero-shot retrieval task in CAV, but exclude evaluation samples that belong to the classes that are not included in any of the tasks. In the audiovisual event localization task, we follow experimental setups in~\citep{lin2023vision}. In the fine-tuning stage of the retrieval and event localization task, we freeze the backbone model, connect it to a randomly initialized trainable linear classifier, and train the classifier with the training dataset to evaluate the acquired representation.

\section{Audio-Video Self-supervised Objectives \label{sec:supple:av-self-sup objectives}}
\begin{figure*}[t]
    
    \centering
    \begin{tabular}{c}
    \includegraphics[width=1.0\linewidth]{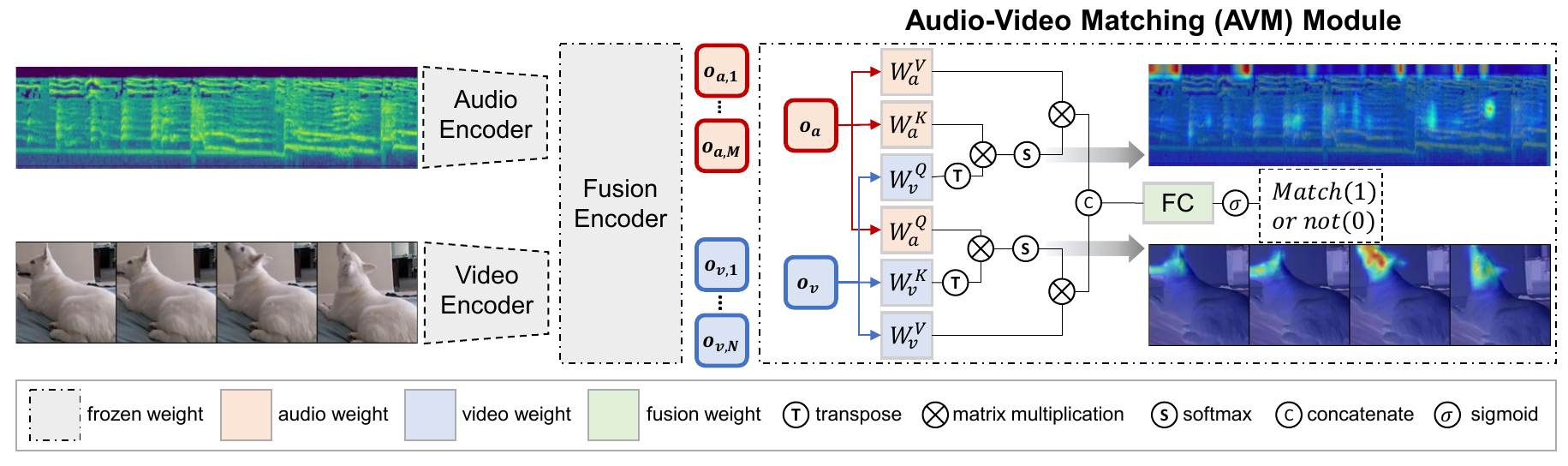}
    \end{tabular}
    % \vspace{-0.15in}
    \caption{\textbf{Overview of AVM module}: The AVM (Audio-Visual Matching) module is self-supervised with the audio-video matching objective. It classifies if the given audio-video pair is positive(audio and video are from the same video) or negative(audio and video are from different videos).}
    \label{fig:avm_overview}
\end{figure*}
Given audio-video data $(X_a,X_v)$, we obtain $D$-dimensional embedding patches $\bm{a}$ and $\bm{v}$ as follows:
\begin{align}
\begin{split}
\bm{a}&=\texttt{Conv2d}\left(X_{a},\bm{w}_{a}\right), ~~\bm{v}=\texttt{Conv2d}\left(X_{v},\bm{w}_{v}\right),
\label{eq:conv_map}
\end{split}
\end{align}
where $\bm{w}_{a},\bm{w}_{v}$ denote the weights of convolutional layers, $\bm{a}\in\mathbb{R}^{B \!\times\! M \!\times\! D}$, and $\bm{v}\in\mathbb{R}^{B \!\times\! N \!\times\! D}$.

The backbone Transformer consists of an audio encoder ($E_a(\cdot)$), a video encoder ($E_v(\cdot)$), a multimodal fusion encoder ($E_f(\cdot)$), and a decoder ($D(\cdot)$). Then we pre-train the model by minimizing the mask reconstruction loss $\ell^{r}$:
\begin{align}
\begin{split}
&\tilde{\bm{a}},~\tilde{\bm{v}}=E_f\left(E_a\left(\bm{m}_a\otimes \bm{a}\right),~E_v\left(\bm{m}_v\otimes \bm{v}\right)\right), \\
&\ell^{r}=\ell^{r}_{a}+\ell^{r}_{v}=
\frac{1}{B}\sum^{B}_{i=1}\left[
\frac{\left(D\left(\tilde{\bm{a}}_{i}\right)-\bm{m}_{a,i}\otimes X_{a,i}\right)^2}{|\bm{m}_{a,i}|}+
\frac{\left(D\left(\tilde{\bm{v}}_{i}\right)-\bm{m}_{v,i}\otimes X_{v,i}\right)^2}{|\bm{m}_{v,i}|}
\right].
\label{eq:recon-loss}
\end{split}
\end{align}
where $\otimes$ denotes vector-matrix multiplication while preserving the input's dimensionality. Random audio $\bm{m}_{a}$ and video mask $\bm{m}_{v}$ are drawn by a binary distribution. In this paper, we set a probability of $0.8$ for masking, consistent with~\cite{Po-Vao2022mavil}. Using the unmasked patches, we aim to learn the model to reconstruct the masked audio and video patches.

In addition, we also minimize masked contrastive loss to learn the semantic relationship between audio and video representation pairs by pulling those that share the same semantics while pushing the others. Following by~\cite{Yuan2023cav}, we pass the masked input patches to audio and video encoders, and subsequently map obtained features (i.e., outputs) to the fusion encoder with modality-specific layer normalization for the masked contrastive learning:
\begin{align}
\begin{split}
&\bm{c}_{a}=\texttt{MeanPool}\left(E_f\left(E_a\left(\bm{m}_a\otimes \bm{a}\right),LN_a\right)\right),\;\;
\bm{c}_{v}=\texttt{MeanPool}\left(E_f\left(E_v\left(\bm{m}_v\otimes \bm{v}\right),LN_v\right)\right),\\
&\ell^c=-\frac{1}{B}\sum^{B}_{i=1}\left[\texttt{log}\left(\frac{\texttt{exp}(\bm{c}_{a,i}^{\top}\bm{c}_{v,i}/\tau)}{\sum^{B}_{j=1}\texttt{exp}(\bm{c}_{a,i}^{\top}\bm{c}_{v,j}/\tau)}\right) + \texttt{log}\left(\frac{\texttt{exp}(\bm{c}_{v,i}^{\top}\bm{c}_{a,i}/\tau)}{\sum^{B}_{j=1}\texttt{exp}(\bm{c}_{v,i}^{\top}\bm{c}_{a,j}/\tau)}\right)\right],
\label{eq:contr-loss}
\end{split}
\end{align}
where $\tau$ is temperature hyperparameter, and $LN_a$ and $LN_v$ indicate modality-specific layer normalization for audio and video each.

\section{Training of Audio-Video Matching Module \label{sec:supple:AVM training}}
\paragraph{AVM training procedure.} In the following section, we describe the training process of the AVM module, as illustrated in \Cref{fig:avm_overview}. 
Given audio-video patch pairs $(\bm{a},\bm{v})$ with the batch size of $B$, we propagate patch inputs to the frozen encoder for each modality and obtain audio-video representation pairs. In order to update the module to capture the multimodal correlation between audio and its video pair, we randomly split them into positive and negative pairs, where we construct negative pairs by randomly shuffling the audio patches to pair with unmatched video patches. Next, we project the obtained positive and negative pairs into fusion space $(\bm{o}_a,\;\bm{o}_v)\!=\!E_f(E_a(\bm{a}),E_v(\bm{v}))$ through the fusion encoder. Subsequently, the input pairs are fed into the AVM module. They are projected to keys, queries, and values for the cross-attention operation, by passing through trainable projection layers. The above process can be summarized as follows:
\begin{align}
\begin{split}
\bm{q}_{a}\!=\!\bm{o}_{a}{\mathcal W}^{Q}_{a},\;\bm{k}_{a}\!=\!\bm{o}_{a}{\mathcal W}^{K}_{a},\;\bm{v}_{a}\!=\!\bm{o}_{a}{\mathcal W}^{V}_{a}&,\;\;~
\bm{q}_{v}\!=\!\bm{o}_{v}{\mathcal W}^{Q}_{v},\;\bm{k}_{v}\!=\!\bm{o}_{v}{\mathcal W}^{K}_{v},\;\bm{v}_{v}\!=\!\bm{o}_{v}{\mathcal W}^{V}_{v},\\
\bm{V}_{a}\!=\!\texttt{Softmax}\left(\mu(\bm{q}_v,\bm{k}_a,\beta\!=\!1)\right)\cdot\bm{v}_{a}&,\;\;~
\bm{V}_{v}\!=\!\texttt{Softmax}\left(\mu(\bm{q}_a,\bm{k}_v,\beta\!=\!1)\right)\cdot\bm{v}_{v},
\label{eq:avm_process}
\end{split}
\end{align}
where the projections ${\mathcal W}^{Q}_{a},\,{\mathcal W}^{K}_{a},\,{\mathcal W}^{V}_{a},\,{\mathcal W}^{Q}_{v},\,{\mathcal W}^{K}_{v},\,{\mathcal W}^{V}_{v}\!\in\!\mathbb{R}^{D \!\times\! H \!\times\! d}$ are trainable parameter matrices; $D\!=\!H*d$. $\bm{V}_{a}\!\in\!\mathbb{R}^{B \!\times\! H \!\times\! N \!\times\! d},\;\bm{V}_{v}\!\in\!\mathbb{R}^{B \!\times\! H \!\times\! M \!\times\! d}$ are values highlighted by the cross-attention maps.

Next, we average the values head-wise and patch-wise, and concatenate the resulting two values into $\bm{va}\!\in\!\mathbb{R}^{B \!\times\! 2D}$ in order to merge the multimodal information. Then it is passed to fully connected (\texttt{FC}) layers, which serve as the classification head. These \texttt{FC} layers take $\bm{va}$ as input, generating a vector $\hat{\bm{y}}\!\in\!\mathbb{R}^{B}$ that predicts whether each input pair corresponds to a negative of positive pair. For training the AVM module, we employ the binary cross-entropy loss to classify audio-video pairs, i.e.,
\begin{align}
\begin{split}
\widehat{\bm{V}}_{av}&=\texttt{Concat}\left(\texttt{MeanPool}\left(\bm{V}_{a}\right),\texttt{MeanPool}\left(\bm{V}_{v}\right)\right),\\
\widehat{\bm{y}}&=\texttt{Sigmoid}\left(\texttt{FC}(\widehat{\bm{V}}_{av})\right),~\mathcal{L}^{avm}=-\bm{y}\left(\texttt{log}(\widehat{\bm{y}})\right),
\label{eq:avm_loss}
\end{split}
\end{align}
Here, $\bm{y}\!=\!\{0,1\}^{B}$ represents ground truth labels, with $\bm{y}_{i}$ taking the value 0 when the $i\-th$ input audio-video pair is a negative pair and 1 otherwise. We pre-train the AVM module along with the backbone model. During the weight update process in the AVM module, the gradient computed from the audio-video matching objective does not propagate through the backbone encoder. This design choice ensures exploiting the AVM at a low cost. Moreover, the AVM only increases 3.18\% of the total backbone model size (707.8 MB), which is efficient compared to methods like \textit{CLS-ER} or \textit{ESMER} which require additional backbones during training.

\section{Additional Experimental Results \label{sec:supple:Additional Experimental Results}}
\begin{figure*}[t]
    \centering
    
    \begin{minipage}{0.63\textwidth}
        \centering
        \includegraphics[width=0.46\linewidth]{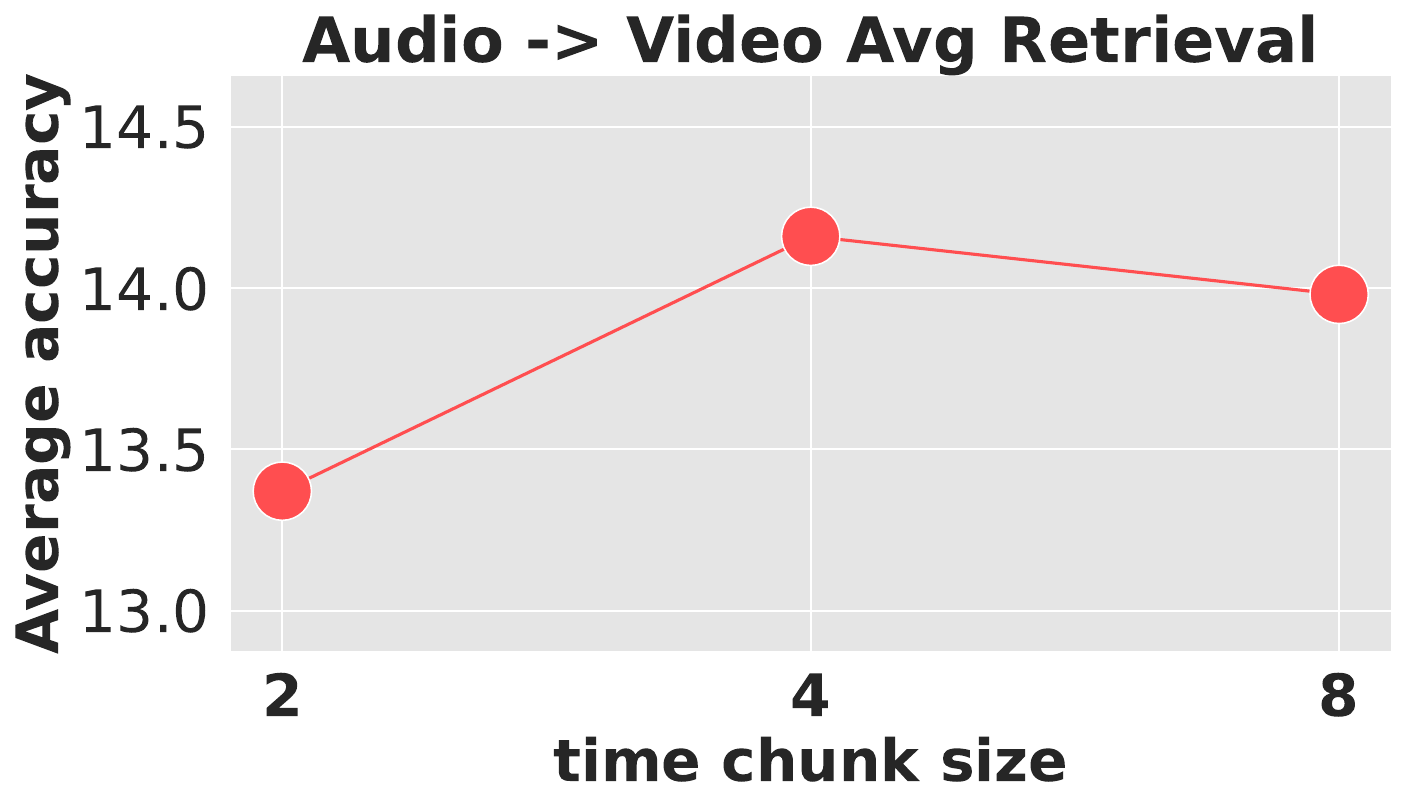}
        \includegraphics[width=0.46\linewidth]{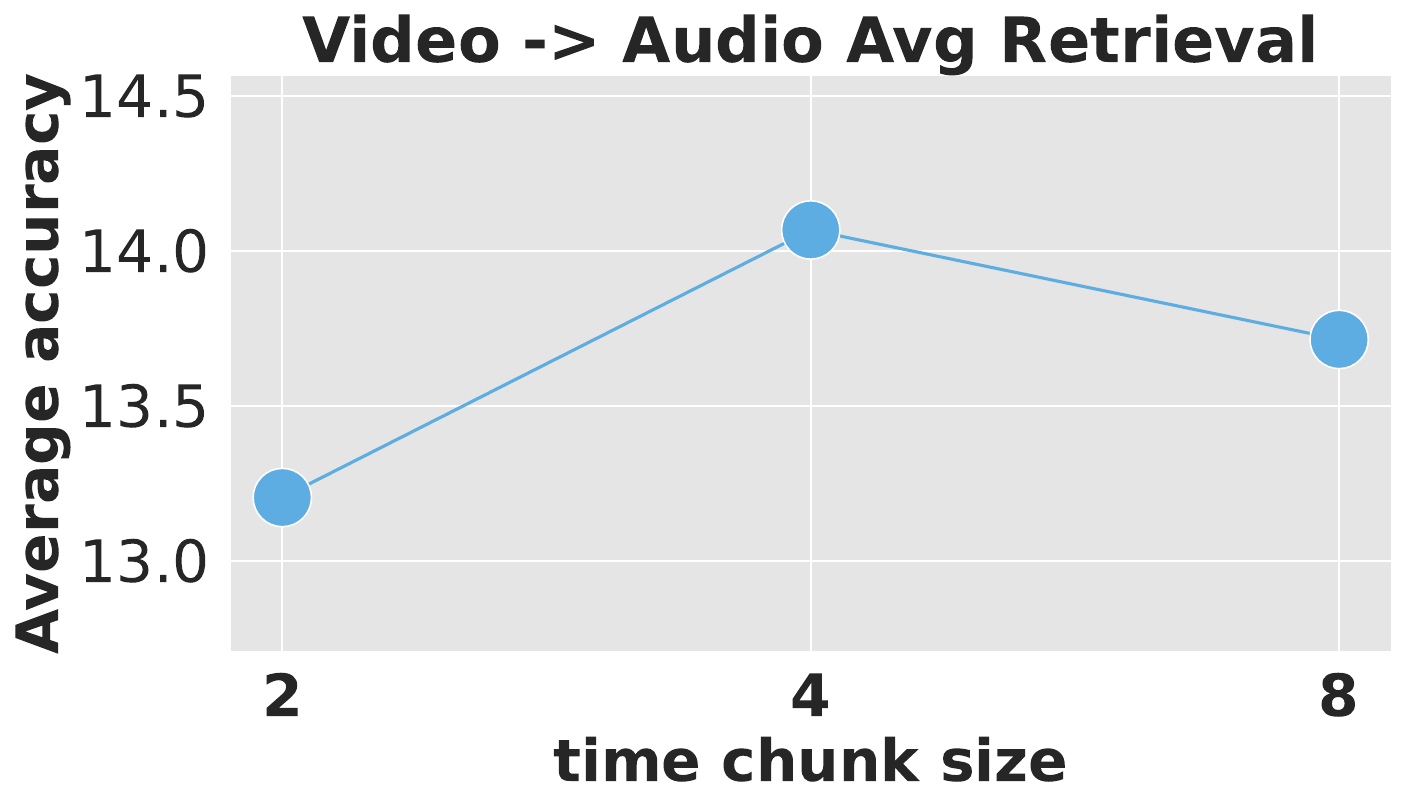}
        \caption*{\textbf{(a) Time chunk sizes}}
    \end{minipage}
    \hspace{0.5cm}
    \begin{minipage}{0.32\textwidth}
        \centering
        \resizebox{1.0\textwidth}{!}{%
\renewcommand{\arraystretch}{1.2}
\renewcommand{\tabcolsep}{5.0pt}
\begin{tabular}{l cc cc}
    \toprule
    Method & \multicolumn{2}{c}{A$\rightarrow$V} & \multicolumn{2}{c}{V$\rightarrow$A} \\
    & $\mathcal{A} \uparrow$ & $\mathcal{F} \downarrow$ & $\mathcal{A} \uparrow$ & $\mathcal{F} \downarrow$ \\
    \midrule
    Frequency  & 13.42 & 5.51 & 12.76 & 6.40 \\
    No constraint  & 12.67 & 6.55 & 12.78 & 6.61 \\
    \textbf{Time} & \textbf{14.16} & \textbf{4.38} & \textbf{14.07} & \textbf{4.65} \\
    \bottomrule
\end{tabular}}
\label{tab:audio_select_method}
        \vspace{0.2in}
        \caption*{\textbf{(b) Audio selection methods}}
    \end{minipage}
    \captionsetup{justification=justified} % Reset caption justification
    \caption{\textbf{Variation of audio patch selection.} \textbf{(a)}: Average retrieval task performance on various time chunk sizes. \textbf{(b)}: Average retrieval task performance on various audio selection methods.} 
    \label{fig:audio_select_hyp_result}
    % \vspace{-0.075in}
\end{figure*}
\begin{table*}[t]
\tiny
% \vspace{-0.2in}
\caption{Results of audiovisual zero-shot retrieval task on \textit{Continual-VS} and \textit{Continual-AS}. We randomly shuffle the task sequences for continual pre-training. For the \textit{Continual-VS}, we follow the task order: music $\rightarrow$ others part1 $\rightarrow$ home\&nature $\rightarrow$ sports $\rightarrow$ others part2 $\rightarrow$ vehicle $\rightarrow$ animals $\rightarrow$ people. For the \textit{Continual-AS}, we follow the task order: nature $\rightarrow$ human $\rightarrow$ home $\rightarrow$ vehicle $\rightarrow$ music $\rightarrow$ animal $\rightarrow$ others. R@K means top-K recall. The results are the means of 3 independent runs. The best and the second best results are highlighted in \textbf{bold} and \underline{underline}, respectively.}
% \vspace{-0.1in}
\centering
\resizebox{1.0\textwidth}{!}{
\renewcommand{\arraystretch}{0.9}
\renewcommand{\tabcolsep}{2.5pt}
\begin{tabular}{ll cc cc cc aa cc cc cc a a}
\toprule
& & \multicolumn{8}{c}{\textbf{Continual-VS}} & \multicolumn{8}{c}{\textbf{Continual-AS}} \\
& {\textbf{Method}}&\multicolumn{2}{c}{\textbf{R@1}} &\multicolumn{2}{c}{\textbf{R@5}}& 
\multicolumn{2}{c}{\textbf{R@10}} &\multicolumn{2}{c}{\textbf{Avg}}&\multicolumn{2}{c}{\textbf{R@1}} &\multicolumn{2}{c}{\textbf{R@5}}& 
\multicolumn{2}{c}{\textbf{R@10}} &\multicolumn{2}{c}{\textbf{Avg}}\\
\midrule
& & $\mathcal{A} \uparrow$ & $\mathcal{F} \downarrow$ & $\mathcal{A} \uparrow$ & $\mathcal{F} \downarrow$ & $\mathcal{A} \uparrow$ & $\mathcal{F} \downarrow$ & $\mathcal{A} \uparrow$ & $\mathcal{F} \downarrow$ & $\mathcal{A} \uparrow$ & $\mathcal{F} \downarrow$ & $\mathcal{A} \uparrow$ & $\mathcal{F} \downarrow$ & $\mathcal{A} \uparrow$ & $\mathcal{F} \downarrow$ & $\mathcal{A} \uparrow$ & $\mathcal{F} \downarrow$ \\
\midrule
\parbox[t]{2mm}{\multirow{11}{*}{\rotatebox[origin=c]{90}{Audio-to-Video}}}
& Finetune &
{\scriptsize 0.80} & {\scriptsize 4.15} &
{\scriptsize 2.96} & {\scriptsize 12.23} &
{\scriptsize 5.05} & {\scriptsize 16.91} & 
{\scriptsize 2.94} & {\scriptsize 11.10} & 
{\scriptsize 1.50} & {\scriptsize 4.72} & 
{\scriptsize 5.49} & {\scriptsize 10.41} & 
{\scriptsize 9.80} & {\scriptsize 11.91} & 
{\scriptsize 5.60} & {\scriptsize 9.01} \\
& ER &
{\scriptsize 3.89} & {\scriptsize 3.06} & 
{\scriptsize 12.10} & {\scriptsize 6.55} & 
{\scriptsize 18.30} & {\scriptsize 7.74} & 
{\scriptsize 11.43} & {\scriptsize 5.78} & 
{\scriptsize 4.52} & {\scriptsize 3.16} & 
{\scriptsize 12.72} & {\scriptsize 6.93} & 
{\scriptsize 18.83} & {\scriptsize 8.00} & 
{\scriptsize 12.02} & {\scriptsize 6.03} \\
& MIR &
{\scriptsize 4.02} & {\scriptsize 2.97} & 
{\scriptsize 12.54} & {\scriptsize 6.16} & 
{\scriptsize 17.99} & {\scriptsize 8.09} & 
{\scriptsize 11.52} & {\scriptsize 5.74} & 
{\scriptsize 4.69} & {\scriptsize \underline{2.95}} & 
{\scriptsize 13.22} & {\scriptsize 6.50} & 
{\scriptsize 18.98} & {\scriptsize 8.81} & 
{\scriptsize 12.30} & {\scriptsize 6.09} \\
& DER++ &
{\scriptsize 4.23} & {\scriptsize 3.35} & 
{\scriptsize 12.92} & {\scriptsize 7.31} & 
{\scriptsize 18.62} & {\scriptsize 9.45} & 
{\scriptsize 11.92} & {\scriptsize 6.70} & 
{\scriptsize 4.32} & {\scriptsize 4.27} & 
{\scriptsize 12.29} & {\scriptsize 8.46} & 
{\scriptsize 18.74} & {\scriptsize 10.18} & 
{\scriptsize 11.78} & {\scriptsize 7.64} \\
& GMED &
{\scriptsize 3.90} & {\scriptsize 2.94} & 
{\scriptsize 11.51} & {\scriptsize 7.41} & 
{\scriptsize 17.65} & {\scriptsize 8.87} & 
{\scriptsize 11.02} & {\scriptsize 6.41} & 
{\scriptsize 4.70} & {\scriptsize \textbf{2.48}} & 
{\scriptsize 12.56} & {\scriptsize \underline{4.55}} & 
{\scriptsize 18.62} & {\scriptsize \textbf{5.05}} & 
{\scriptsize 11.96} & {\scriptsize \textbf{4.03}} \\
& CLS-ER &
{\scriptsize 3.94} & {\scriptsize 3.35} & 
{\scriptsize 12.96} & {\scriptsize 7.19} & 
{\scriptsize 18.09} & {\scriptsize 10.66} & 
{\scriptsize 11.66} & {\scriptsize 7.07} & 
{\scriptsize 5.16} & {\scriptsize 2.97} & 
{\scriptsize 14.33} & {\scriptsize 6.88} & 
{\scriptsize 20.24} & {\scriptsize 8.74} & 
{\scriptsize 13.24} & {\scriptsize 6.20} \\
& LUMP &
{\scriptsize 4.06} & {\scriptsize \textbf{2.18}} & 
{\scriptsize 13.21} & {\scriptsize \textbf{4.66}} & 
{\scriptsize 19.34} & {\scriptsize \textbf{5.58}} & 
{\scriptsize 12.20} & {\scriptsize \textbf{4.14}} & 
{\scriptsize 4.45} & {\scriptsize 3.40} & 
{\scriptsize 13.05} & {\scriptsize 6.25} & 
{\scriptsize 19.45} & {\scriptsize 7.28} & 
{\scriptsize 12.32} & {\scriptsize 5.64} \\
& ESMER &
{\scriptsize 4.38} & {\scriptsize 3.36} &
{\scriptsize 13.31} & {\scriptsize 8.28} &
{\scriptsize 19.39} & {\scriptsize 9.20} &
{\scriptsize 12.36} & {\scriptsize 6.95} &
{\scriptsize \underline{5.43}} & {\scriptsize 3.85} &
{\scriptsize \underline{15.81}} & {\scriptsize 6.20} &
{\scriptsize \underline{21.40}} & {\scriptsize 8.81} &
{\scriptsize \underline{14.21}} & {\scriptsize 6.29} \\
\cmidrule{2-18}

& \cellcolor{gg}STELLA (Ours)&
{\scriptsize \cellcolor{gg} \underline{4.72}} & {\scriptsize \cellcolor{gg} \underline{2.89}} &
{\scriptsize \cellcolor{gg} \underline{14.17}} & {\scriptsize \cellcolor{gg} 5.74} &
{\scriptsize \cellcolor{gg} \underline{19.94}} & {\scriptsize \cellcolor{gg} \underline{5.74}} &
{\scriptsize \cellcolor{gg} \underline{12.94}} & {\scriptsize \cellcolor{gg} 4.79} &
{\scriptsize \cellcolor{gg} 4.97} & {\scriptsize \cellcolor{gg} 3.47} &
{\scriptsize \cellcolor{gg} 13.91} & {\scriptsize \cellcolor{gg} 5.59} &
{\scriptsize \cellcolor{gg} 20.30} & {\scriptsize \cellcolor{gg} \underline{6.70}} &
{\scriptsize \cellcolor{gg} 13.06} & {\scriptsize \cellcolor{gg} 5.25} \\
& \cellcolor{gg}STELLA+ (Ours)&
{\scriptsize \cellcolor{gg} \textbf{4.90}} & {\scriptsize \cellcolor{gg} 3.19} &
{\scriptsize \cellcolor{gg} \textbf{16.42}} & {\scriptsize \cellcolor{gg} \underline{4.72}} &
{\scriptsize \cellcolor{gg} \textbf{23.49}} & {\scriptsize \cellcolor{gg} 5.89} &
{\scriptsize \cellcolor{gg} \textbf{14.94}} & {\scriptsize \cellcolor{gg} \underline{4.60}} &
{\scriptsize \cellcolor{gg} \textbf{5.77}} & {\scriptsize \cellcolor{gg} 3.90} &
{\scriptsize \cellcolor{gg} \textbf{17.51}} & {\scriptsize \cellcolor{gg} \textbf{4.49}} &
{\scriptsize \cellcolor{gg} \textbf{23.72}} & {\scriptsize \cellcolor{gg} 7.07} &
{\scriptsize \cellcolor{gg} \textbf{15.67}} & {\scriptsize \cellcolor{gg} \underline{5.15}} \\
\cmidrule{2-18}

& Multitask &
{\scriptsize 6.45} & $-$ & 
{\scriptsize 20.19} & $-$ & 
{\scriptsize 29.01} & $-$ & 
{\scriptsize 18.55} & $-$ &
{\scriptsize 8.28} & $-$ & 
{\scriptsize 24.14} & $-$ & 
{\scriptsize 33.74} & $-$ & 
{\scriptsize 22.05} & $-$ \\
\midrule

\parbox[t]{2mm}{\multirow{11}{*}{\rotatebox[origin=c]{90}{Video-to-Audio}}}
& Finetune & 
{\scriptsize 0.78} & {\scriptsize 3.77} & 
{\scriptsize 3.00} & {\scriptsize 11.68} & 
{\scriptsize 5.21} & {\scriptsize 15.86} & 
{\scriptsize 3.00} & {\scriptsize 10.44} & 
{\scriptsize 1.42} & {\scriptsize 5.11} & 
{\scriptsize 6.54} & {\scriptsize 10.30} & 
{\scriptsize 10.43} & {\scriptsize 13.48} & 
{\scriptsize 6.13} & {\scriptsize 9.63} \\
& ER~ &
{\scriptsize 3.57} & {\scriptsize 2.76} & 
{\scriptsize 11.66} & {\scriptsize 7.67} & 
{\scriptsize 16.75} & {\scriptsize 10.76} & 
{\scriptsize 10.66} & {\scriptsize 7.06} & 
{\scriptsize 4.01} & {\scriptsize 4.31} & 
{\scriptsize 12.47} & {\scriptsize 7.27} & 
{\scriptsize 19.32} & {\scriptsize \underline{9.26}} & 
{\scriptsize 11.93} & {\scriptsize 6.95} \\
& MIR~ &
{\scriptsize 3.35} & {\scriptsize 3.15} & 
{\scriptsize 11.37} & {\scriptsize 7.74} & 
{\scriptsize 16.62} & {\scriptsize 10.11} & 
{\scriptsize 10.45} & {\scriptsize 7.00} & 
{\scriptsize 4.25} & {\scriptsize 3.43} & 
{\scriptsize 12.92} & {\scriptsize 6.93} & 
{\scriptsize 19.43} & {\scriptsize 9.78} & 
{\scriptsize 12.20} & {\scriptsize 6.71} \\
& DER++ &
{\scriptsize 4.08} & {\scriptsize 3.10} & 
{\scriptsize 12.78} & {\scriptsize 9.02} & 
{\scriptsize 18.77} & {\scriptsize 11.30} & 
{\scriptsize 11.88} & {\scriptsize 7.81} & 
{\scriptsize 4.31} & {\scriptsize 4.35} & 
{\scriptsize 12.60} & {\scriptsize 9.59} & 
{\scriptsize 18.93} & {\scriptsize 12.27} & 
{\scriptsize 11.95} & {\scriptsize 8.74} \\
& GMED &
{\scriptsize 3.42} & {\scriptsize 3.80} & 
{\scriptsize 11.45} & {\scriptsize 7.76} & 
{\scriptsize 17.06} & {\scriptsize 9.94} & 
{\scriptsize 10.64} & {\scriptsize 7.17} & 
{\scriptsize 4.20} & {\scriptsize \textbf{1.87}} & 
{\scriptsize 12.97} & {\scriptsize \textbf{6.04}} & 
{\scriptsize 19.98} & {\scriptsize \textbf{8.11}} & 
{\scriptsize 12.38} & {\scriptsize \textbf{5.34}} \\
& CLS-ER &
{\scriptsize 3.49} & {\scriptsize 3.85} & 
{\scriptsize 12.28} & {\scriptsize 8.05} & 
{\scriptsize 17.75} & {\scriptsize 11.31} & 
{\scriptsize 11.17} & {\scriptsize 7.74} & 
{\scriptsize 4.85} & {\scriptsize 5.48} & 
{\scriptsize 13.37} & {\scriptsize 9.17} & 
{\scriptsize 19.69} & {\scriptsize 11.36} & 
{\scriptsize 12.64} & {\scriptsize 8.67} \\
& LUMP &
{\scriptsize 3.98} & {\scriptsize \textbf{1.67}} & 
{\scriptsize 12.44} & {\scriptsize \textbf{5.17}} & 
{\scriptsize 18.11} & {\scriptsize \textbf{7.27}} & 
{\scriptsize 11.51} & {\scriptsize \textbf{4.70}} & 
{\scriptsize 4.23} & {\scriptsize 4.06} & 
{\scriptsize 13.53} & {\scriptsize \underline{6.09}} & 
{\scriptsize 19.27} & {\scriptsize 9.53} & 
{\scriptsize 12.34} & {\scriptsize 6.56} \\
& ESMER &
{\scriptsize \underline{4.44}} & {\scriptsize 3.35} &
{\scriptsize 13.32} & {\scriptsize 8.69} &
{\scriptsize 19.47} & {\scriptsize 10.27} &
{\scriptsize 12.41} & {\scriptsize 7.44} &
{\scriptsize \underline{5.12}} & {\scriptsize 5.48} &
{\scriptsize \underline{14.73}} & {\scriptsize 8.79} &
{\scriptsize \underline{20.35}} & {\scriptsize 12.41} &
{\scriptsize \underline{13.40}} & {\scriptsize 8.89} \\
\cmidrule{2-18}
& \cellcolor{gg}STELLA (Ours)&
{\scriptsize \cellcolor{gg} 4.18} & {\scriptsize \cellcolor{gg} 2.54} &
{\scriptsize \cellcolor{gg} \underline{13.81}} & {\scriptsize \cellcolor{gg} 6.56} &
{\scriptsize \cellcolor{gg} \underline{19.90}} & {\scriptsize \cellcolor{gg} 8.88} &
{\scriptsize \cellcolor{gg} \underline{12.63}} & {\scriptsize \cellcolor{gg} 5.99} &
{\scriptsize \cellcolor{gg} 4.86} & {\scriptsize \cellcolor{gg} \underline{2.92}} &
{\scriptsize \cellcolor{gg} 14.20} & {\scriptsize \cellcolor{gg} 6.41} &
{\scriptsize \cellcolor{gg} 20.00} & {\scriptsize \cellcolor{gg} 9.82} &
{\scriptsize \cellcolor{gg} 13.02} & {\scriptsize \cellcolor{gg} \underline{6.38}} \\
& \cellcolor{gg}STELLA+ (Ours)&
{\scriptsize \cellcolor{gg} \textbf{5.28}} & {\scriptsize \cellcolor{gg} \underline{1.81}} &
{\scriptsize \cellcolor{gg} \textbf{15.35}} & {\scriptsize \cellcolor{gg} \underline{6.33}} &
{\scriptsize \cellcolor{gg} \textbf{21.97}} & {\scriptsize \cellcolor{gg} \underline{8.01}} &
{\scriptsize \cellcolor{gg} \textbf{14.20}} & {\scriptsize \cellcolor{gg} \underline{5.38}} &
{\scriptsize \cellcolor{gg} \textbf{5.57}} & {\scriptsize \cellcolor{gg} 3.80} &
{\scriptsize \cellcolor{gg} \textbf{16.67}} & {\scriptsize \cellcolor{gg} 6.96} &
{\scriptsize \cellcolor{gg} \textbf{23.91}} & {\scriptsize \cellcolor{gg} 9.28} &
{\scriptsize \cellcolor{gg} \textbf{15.38}} & {\scriptsize \cellcolor{gg} 6.68} \\
\cmidrule{2-18}
\cmidrule{2-18}

& Multitask &
{\scriptsize 6.85} & $-$ & 
{\scriptsize 21.93} & $-$ & 
{\scriptsize 30.63} & $-$ & 
{\scriptsize 19.80} & $-$ &
{\scriptsize 8.05} & $-$ & 
{\scriptsize 25.81} & $-$ & 
{\scriptsize 35.60} & $-$ & 
{\scriptsize 23.15} & $-$ \\

\bottomrule
\end{tabular}}
\label{tab:shuffle_retrieval_table}
% \vspace{-0.15in}
\end{table*}

\paragraph{Audio patch selection strategy.}
When executing the selection of audio patches guided by the audio importance score $\bm{I}_{a}$, our approach involves selecting patches in time-wise segments, following the procedure detailed in~\Cref{alg:audio_time_chunk_select_pytorch}. As spectrogram patches exhibit local correlation driven by their temporal continuity~\citep{Po-Yao2022audiomae}, the strategy for audio patch selection becomes pivotal in maintaining these intrinsic properties. The challenge lies in striking a balance between retaining time continuity and eliminating redundant information within the spectrogram.

In pursuit of this balance, we conduct various experiments on the audio patch selection approach. The width of the time chunk assumes significance; a chunk that is too narrow could disrupt time continuity, while one that is excessively wide might not concisely capture core information. To validate our approach and assess the efficacy of time-wise chunk selection, we conduct two distinct sets of experiments.

The first experiment involves evaluating the model's performance across varying time chunk widths. A noteworthy observation from~\Cref{fig:audio_select_hyp_result} \highlight{(a)}: adopting a size of 2 results in a noticeable performance decline. This potentially signifies the criticality of upholding the local correlation inherent in audio patches. Moving on to the second experiment, we explore various selection methods, inspired by the spectrogram masking techniques detailed in~\citep{Po-Yao2022audiomae}. We test two variants of audio patch selection: Frequency indicates an approach of choosing audio patches frequency-wise, while No-constraint indicates selecting audio patches without any constraints; applying the same patch selection procedure as in the video patch selection. As shown in~\Cref{fig:audio_select_hyp_result} \highlight{(b)}, time-wise selection exhibits superior performance compared to alternative audio selection methodologies, meaning that preserving audio information in time-chunk minimizes loss of audio properties.

\paragraph{Shuffle task orders.} In addition to the main experiment results presented in~\Cref{tab:retrieval_table}, we conduct supplementary investigations with the intention of enhancing the reliability of our findings. Specifically, we carry out experiments on shuffled task sequences. For the \textit{Continual-VS}, we randomize the original pre-train task sequence, leading to modified order: music$\rightarrow$others part1$\rightarrow$home\&nature$\rightarrow$sports$\rightarrow$others part2$\rightarrow$vehicle$\rightarrow$animals$\rightarrow$people. Likewise, in the case of the \textit{Continual-AS} experiment, we apply a similar task sequence shuffling, resulting in the following order: nature$\rightarrow$human$\rightarrow$home$\rightarrow$vehicle$\rightarrow$music$\rightarrow$animal$\rightarrow$others. Note that the \textit{Continual-VS} experiment is conducted on 36 batch size, unlike the main \textit{Continual-VS} experiment which is conducted on 48 batch size. We present the corresponding audiovisual zero-shot retrieval task results in~\Cref{tab:shuffle_retrieval_table}. Our method shows competitive or better performance compared to other baselines, which coincides with the results in~\Cref{tab:retrieval_table}. This indicates that our method is robust under varying conditions, thereby enhancing the credibility of our analysis.

\begin{table*}[t]
    \caption{\textbf{Downstream tasks} \textbf{(a):} MSR-VTT audiovisual retrieval. MSR-VTT audiovisual retrieval task performances. We use the models continually pre-trained until completion of the last task of \textit{Continual-AS}. \textbf{(b):} We randomly initialize and finetune a MLP classifier with AVE dataset~\citep{ave}. The best and the second best results are highlighted in \textbf{bold} and \underline{underline}, respectively.}
    \label{fig:supple_downstream_tasks}
    \centering
    \begin{minipage}{0.5\textwidth}
    \centering
        {\textbf{(a) MSR-VTT audiovisual retrieval}}
        \vspace{0.05in}\\
        \centering
\tiny
\resizebox{\linewidth}{!}{%
\renewcommand{\arraystretch}{1.1}
\renewcommand{\tabcolsep}{3.4pt}
\begin{tabular}{l c c c c c c c c}
    \toprule
    \textbf{Method} & \multicolumn{3}{c}{\textbf{A$\rightarrow$V}} & \multicolumn{3}{c}{\textbf{V$\rightarrow$A}} \\
    & \textbf{R@1} & \textbf{R@5} & \textbf{R@10} & \textbf{R@1} & \textbf{R@5} & \textbf{R@10} \\
    \midrule
    Finetune & 0.52 & 2.81 & 4.82 & 0.67 & 2.82 & 5.08 \\
    ER & 1.48 & 6.70 & 11.48 & 1.74 & 7.19 & 12.07 \\
    MIR & 1.56 & 5.97 & 10.23 & 1.85 & 6.93 & 11.89 \\
    DER++ & 2.74 & 9.08 & 14.49 & 2.45 & 9.49 & 14.60 \\
    GMED & 2.07 & 8.04 & 13.11 & 2.70 & 8.44 & 12.89 \\
    CLS-ER & 2.78 & \underline{9.40} & 14.43 & \underline{2.89} & 8.73 & 14.54 \\
    LUMP & 2.33 & 8.15 & 12.75 & 2.04 & 7.93 & 12.45 \\
    ESMER & \underline{2.89} & 9.70 & \underline{15.56} & 2.70 & \textbf{10.22} & \underline{16.04} \\
    \midrule
    \cellcolor{gg}STELLA (Ours) & \cellcolor{gg} 2.74 & \cellcolor{gg} 9.26 & \cellcolor{gg} 15.37 & \cellcolor{gg} 2.85 & \cellcolor{gg} 9.48 & \cellcolor{gg} 15.56 \\
    \cellcolor{gg}STELLA+ (Ours) & \cellcolor{gg} \textbf{2.93} & \cellcolor{gg} \textbf{10.22} & \cellcolor{gg} \textbf{16.33} & \cellcolor{gg} \textbf{3.67} & \cellcolor{gg} \textbf{10.22} & \cellcolor{gg} \textbf{16.26} \\
    \bottomrule
\end{tabular}}
\label{tab:sub:msrvtt_retrieval_audioset}
        
    \end{minipage}
    \hspace{0.5in}
    \begin{minipage}{0.28\textwidth}
    \centering
        {\textbf{(b) Audiovisual event localization}}
        \vspace{0.05in}\\
        \centering
\tiny
\resizebox{1.0\textwidth}{!}{
\renewcommand{\tabcolsep}{5.5pt}
    \begin{tabular}{ll c}
        \toprule
        & {\textbf{Method}} & Acc \\
        \midrule
        \parbox[t]{2mm}{\multirow{9}{*}{\rotatebox[origin=c]{90}{AVE}}}
        & Finetune & {52.56} \\
        & ER & {54.98} \\
        & MIR & {56.13} \\
        & DER++ & {55.81} \\
        & GMED & {55.98} \\
        & CLS-ER & {\underline{56.39}} \\
        & LUMP & {55.06} \\
        & ESMER & {55.60} \\
        \cmidrule{2-3}
        & \cellcolor{gg}STELLA (Ours)&
        {\cellcolor{gg}\textbf{56.68}} \\
        & \cellcolor{gg}STELLA+ (Ours)&
        {\cellcolor{gg}\textbf{56.68}} \\
        \cmidrule{2-3}
        & Multitask & {57.73} \\
        \bottomrule
    \end{tabular}
}
\label{tab:ave_table}
        
    \end{minipage}    
\end{table*}
\paragraph{MSR-VTT retrieval task.} We provide additional experiment results on the MSR-VTT retrieval task in~\Cref{fig:supple_downstream_tasks} \highlight{(a)}. In this experiment, we use the models continually pre-trained up to the last task of \textit{Continual-AS}. We follow the training configurations in~\Cref{tab:training_hyp}. The experiment results show that our methods consistently show competitive results, which supports that our methods obtain general audio-video correlations that are transferable to retrieval tasks.

\paragraph{Audiovisual event localization.} We conduct an audiovisual event localization (AVE) task to showcase the effectiveness of our method in precisely aligning audio and video streams. Following the experimental setup outlined in~\cite{lin2023vision}, we utilize the AVE dataset~\citep{ave} for the experiment. To assess whether continually pre-trained models can adapt to the downstream task involving the unseen dataset, we use the model pre-trained on all tasks in the sequence within the \textit{Continual-VS} experiment. The training process adheres to the hyperparameters described in~\Cref{tab:training_hyp}, wherein the backbone model remains frozen while training the linear classifier. We present the summarized result in~\Cref{fig:supple_downstream_tasks} \highlight{(b)}. This result demonstrates that our method surpasses other baseline methods. This underscores the strength of our method in adapting the downstream task that necessitates a sophisticated grasp of audio-video alignment at a high level.

\begin{figure*}[h!]
    \centering
    % \small
    \begin{minipage}{\textwidth}
        \centering
        \includegraphics[width=0.9\linewidth]{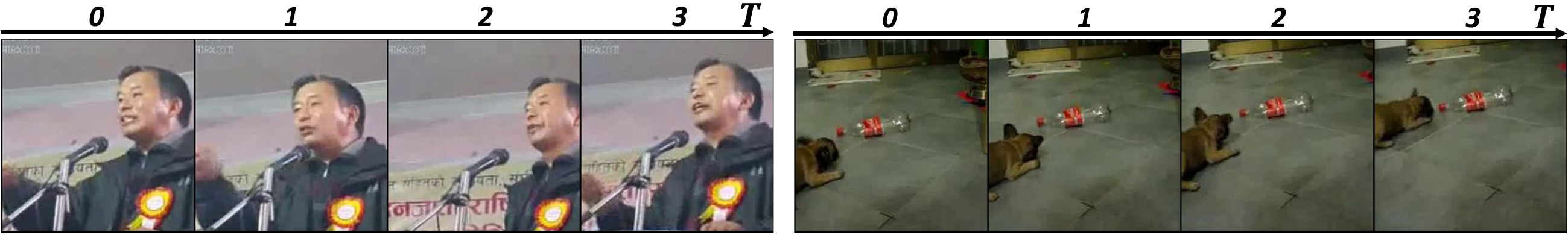}
    \end{minipage}
    \textbf{(a) Raw data}
    \vspace{0.05in}
    
    \begin{minipage}{\textwidth}
        \centering
        \includegraphics[width=0.9\linewidth]{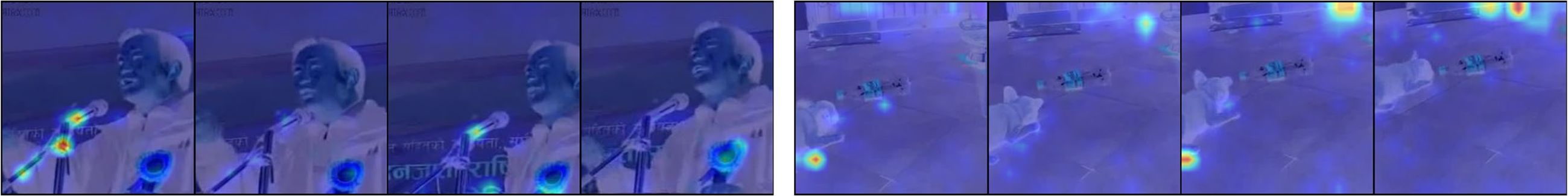}
    \end{minipage}
    \textbf{(b) Audiovisual attention (CLS-ER)}
    \vspace{0.05in}

    \begin{minipage}{\textwidth}
        \centering
        \includegraphics[width=0.9\linewidth]{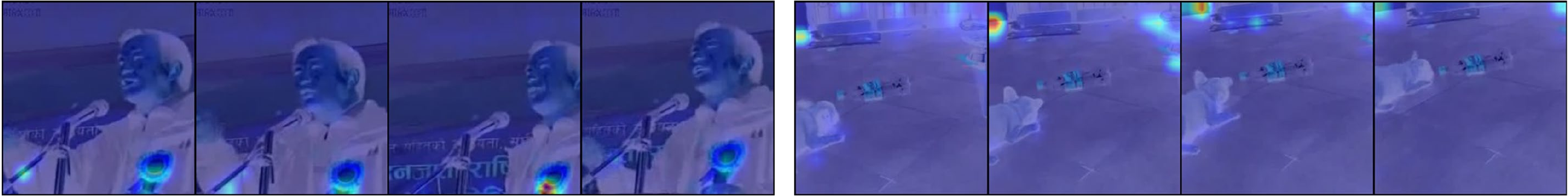}
    \end{minipage}
    \textbf{(c) Audiovisual attention (ESMER)}
    \vspace{0.05in}

    \begin{minipage}{\textwidth}
        \centering
        \includegraphics[width=0.9\linewidth]{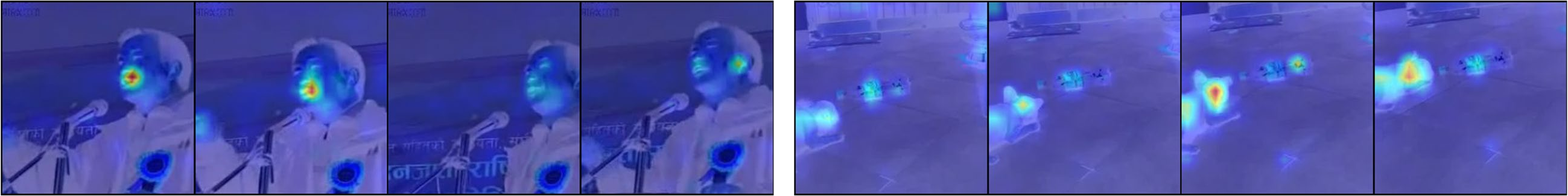}
    \end{minipage}
    \textbf{(d) Audiovisual attention (STELLA)}
    \vspace{0.15in}

        \begin{minipage}{\textwidth}
        \centering
        \includegraphics[width=0.9\linewidth]{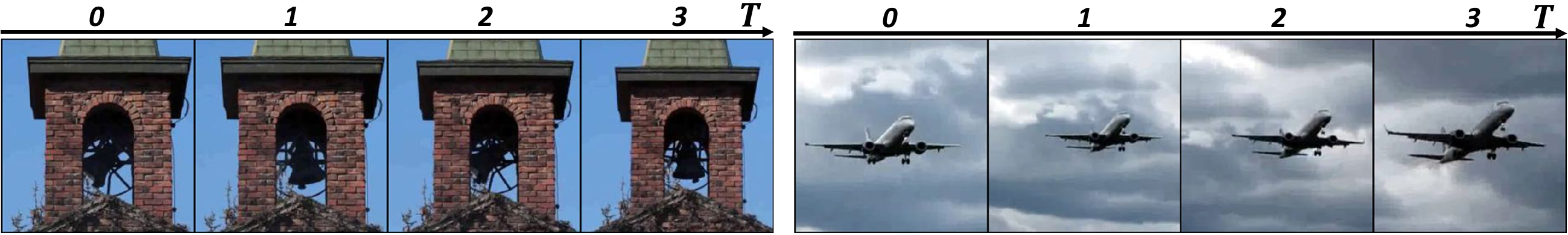}
    \end{minipage}
    \textbf{(a) Raw data}
    \vspace{0.05in}
    
    \begin{minipage}{\textwidth}
        \centering
        \includegraphics[width=0.9\linewidth]{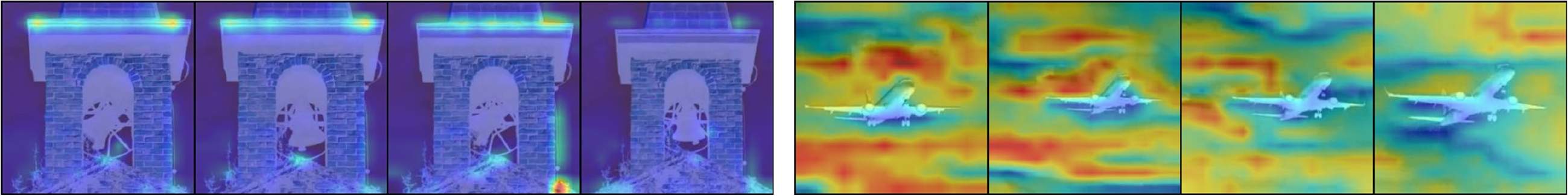}
    \end{minipage}
    \textbf{(b) Audiovisual attention (CLS-ER)}
    \vspace{0.05in}

    \begin{minipage}{\textwidth}
        \centering
        \includegraphics[width=0.9\linewidth]{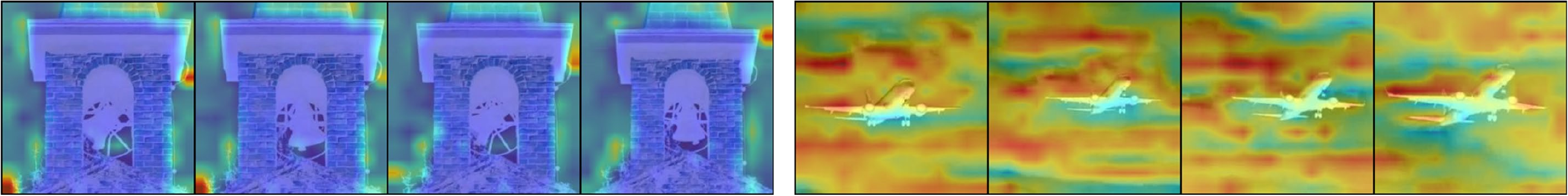}
    \end{minipage}
    \textbf{(c) Audiovisual attention (ESMER)}
    \vspace{0.05in}

    \begin{minipage}{\textwidth}
        \centering
        \includegraphics[width=0.9\linewidth]{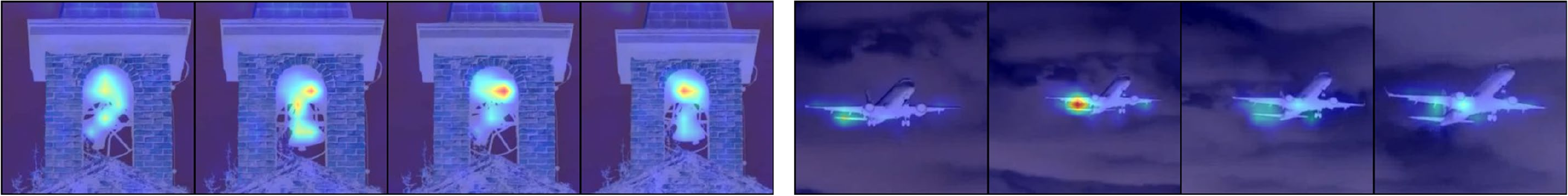}
    \end{minipage}
    \textbf{(d) Audiovisual attention (STELLA)}
    \vspace{0.15in}

    \captionsetup{justification=justified} % Reset caption justification
    \caption{\textbf{Sound source localization} \textbf{(a)} Examples of raw video frames. \textbf{(b)\textasciitilde(c)}: We visualize cross-attention maps using cosine similarity between each video patch and averaged audio embedding. \textbf{(d)}: We use the AVM module in \textit{STELLA}, continually pre-trained with the backbone mode, to visualize cross-attention maps. Our method is much more effective in capturing potential sound sources compared to the ability of the backbone to capture the sources.}
    \label{fig:supple_sound_source_localization}
    % \vspace{-0.075in}
\end{figure*}

\paragraph{Sound source localization.} We provide more visualization results of the sound source localization in~\Cref{fig:supple_sound_source_localization}. Our method consistently shows superior ability in locating potential sound sources in the visual scenes.

\section{Hyperparamter Tuning Results \label{sec:supple:Hyperparameter Tuning Results}}
\paragraph{Patch sampling ratio.}
\begin{wraptable}{t}{0.35\textwidth}
    \centering
    % \vspace{-0.15in}
    \caption{Retrieval result by sampling ratios.}
    % \vspace{-0.1in}
    \begin{adjustbox}{width=\linewidth}
    \begin{tabular}{lc cc cc}
        \toprule
        & Ratio(\%) & \multicolumn{2}{c}{A$\rightarrow$V} & \multicolumn{2}{c}{V$\rightarrow$A} \\
        & & $\mathcal{A} \uparrow$ & $\mathcal{F} \downarrow$ & $\mathcal{A} \uparrow$ & $\mathcal{F} \downarrow$ \\
        \midrule
        \parbox[t]{2mm}{\multirow{3}{*}{\parbox[t]{2cm}{$\rho_a$}}}
        & 37.5 & 13.76 & 4.77 & 13.52 & 5.53 \\ 
        & \textbf{50} & \textbf{14.16} & \textbf{4.38} & \textbf{14.07} & \textbf{4.65}  \\
        & 62.5 & 13.77 & 5.04 & 13.46 & 5.06 \\
        \midrule
        \parbox[t]{2mm}{\multirow{3}{*}{\parbox[t]{2cm}{$\rho_v$}}}
        & 37.5 & 13.35 & 5.57 & 13.39 & 5.93 \\
        & \textbf{50} & \textbf{14.16} & \textbf{4.38} & \textbf{14.07} & \textbf{4.65}  \\
        & 62.5 & 13.82 & 4.50 & 13.53 & 5.27 \\
        \bottomrule
    \end{tabular}
    \end{adjustbox}
    \label{tab:sample_ratio_hyp}
\end{wraptable}
Central to our approach is the identification of patches that exhibit a high localized alignment with their corresponding modality pairs while being robust to catastrophic forgetting of learned representation, enabling the retention of meaningful information. Achieving the right balance in the sampling ratio is critical: an excessively low sampling ratio hinders the model from accessing essential data, while an overly high ratio hampers the model's ability to disregard redundant or forget-inducing information.

For the audio sampling ratio, we systematically assess three options —37.5\%, 50\%, and 62.5\%— while maintaining the video sampling ratio $\rho_v$ at 50\%.~\Cref{tab:sample_ratio_hyp} shows that sampling 50\% of audio patches ensures high performance compared to the other sampling ratios. It is noteworthy that the other sampling ratios still yield competitive performance compared to the baselines. As we transition to optimizing the sampling ratio for video patches, we conduct experiments using three sampling ratios -37.5\%, 50\%, and 62.5\%- alongside the audio sampling ratio $\rho_a$ at 50\%. As demonstrated in~\Cref{tab:sample_ratio_hyp}, employing a 50\% video sampling ratio ensures high performance.

\paragraph{Inference temperature in AVM module.}
\begin{wraptable}{t}{0.35\textwidth}
    \centering
    % \vspace{-0.15in}
    \renewcommand{\tabcolsep}{10pt}
    \caption{Retrieval result by temperature values.}
    % \vspace{-0.1in}
    \begin{adjustbox}{width=\linewidth}
    \begin{tabular}{c cc cc}
        \toprule
        $\beta$ & \multicolumn{2}{c}{A$\rightarrow$V} & \multicolumn{2}{c}{V$\rightarrow$A} \\
        & $\mathcal{A} \uparrow$ & $\mathcal{F} \downarrow$ & $\mathcal{A} \uparrow$ & $\mathcal{F} \downarrow$ \\
        \midrule
        0.1 & 13.91 & 5.42 & \textbf{14.23} & 4.97 \\ 
        \textbf{0.4} & \textbf{14.16} & \textbf{4.38} & 14.07 & \textbf{4.65}  \\
        0.5 & 13.37 & 5.27 & 13.50 & 5.84 \\
        \bottomrule
    \end{tabular}
    \end{adjustbox}
    \label{tab:temperature_hyp}
    % \vspace{-0.1in}
\end{wraptable}

In our approach, we actively harness cross-attention maps from the AVM module computed in \Cref{eq:cross_att_compute}. During inference, we set the temperature hyperparameter $\beta$ to 0.4 for the \textit{Continual-VS} experiments. To examine the significance of $\beta$, we explore a range of the hyperparameter values, specifically 0.1, 0.4, and 0.5. The results, as summarized in~\Cref{tab:temperature_hyp}, indicate that the optimal temperature values typically reside within the range of approximately 0.1 to 0.4. This suggests the need for heightened emphasis on discriminative audio and video patches in order that those patches are more frequently selected in our selection framework in \Cref{eq:video_patch_select} and in \Cref{alg:audio_time_chunk_select_pytorch}.

\begin{figure*}[t]
    \centering
    \resizebox{1.0\linewidth}{!}{%
    \begin{tabular}{ccc}
        \includegraphics[width=0.33\linewidth]{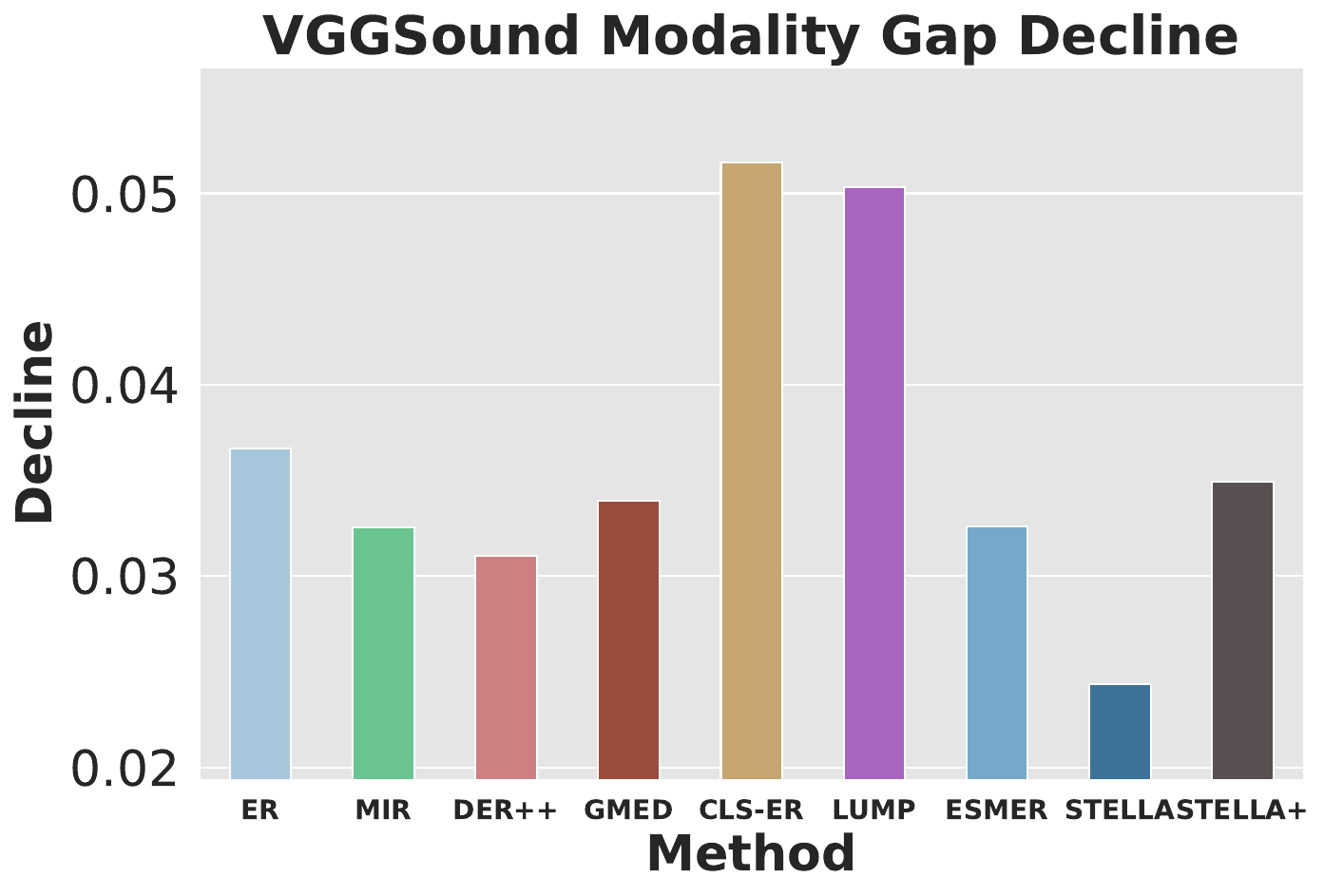}&
        \includegraphics[width=0.33\linewidth]{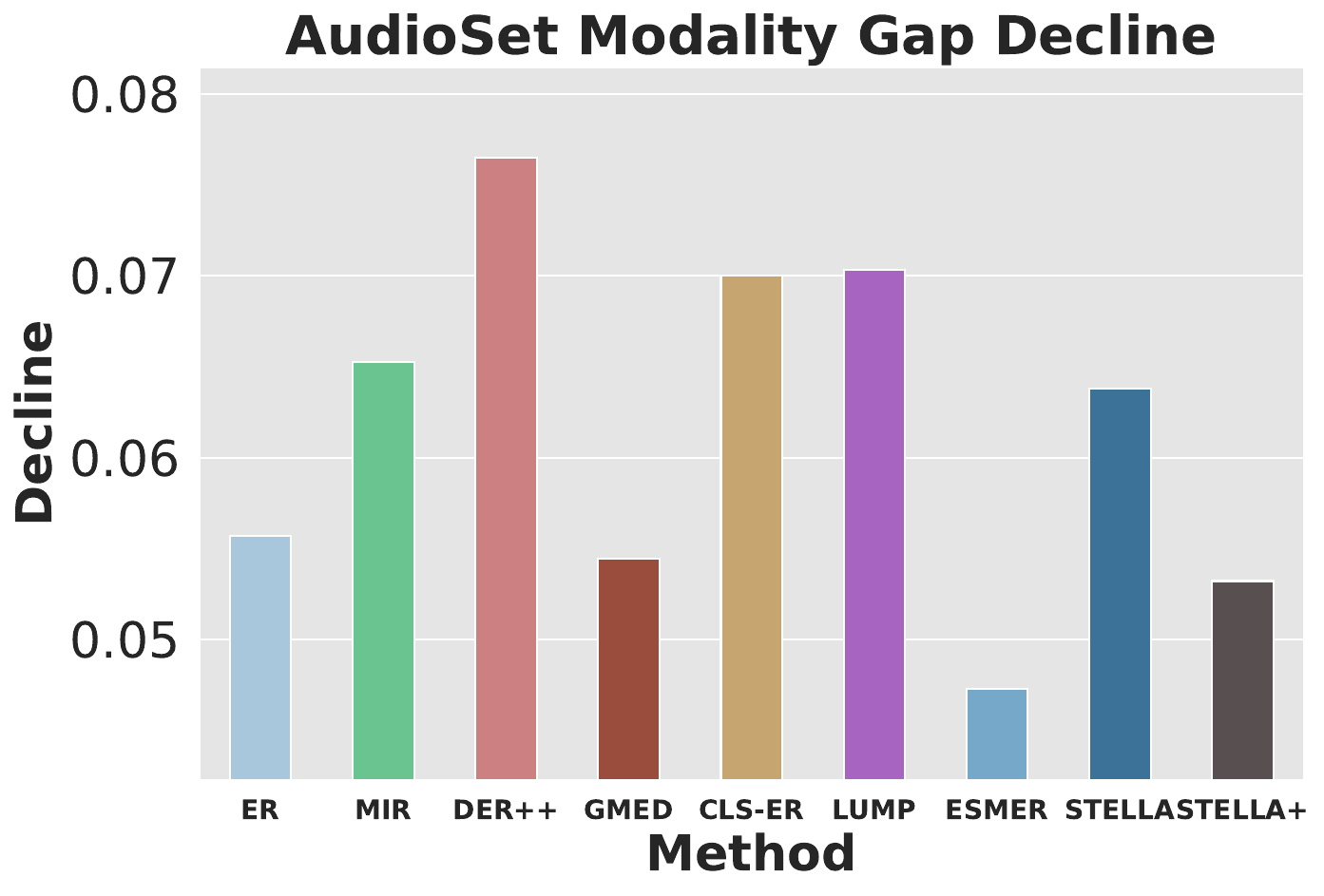}&
        \includegraphics[width=0.33\linewidth]{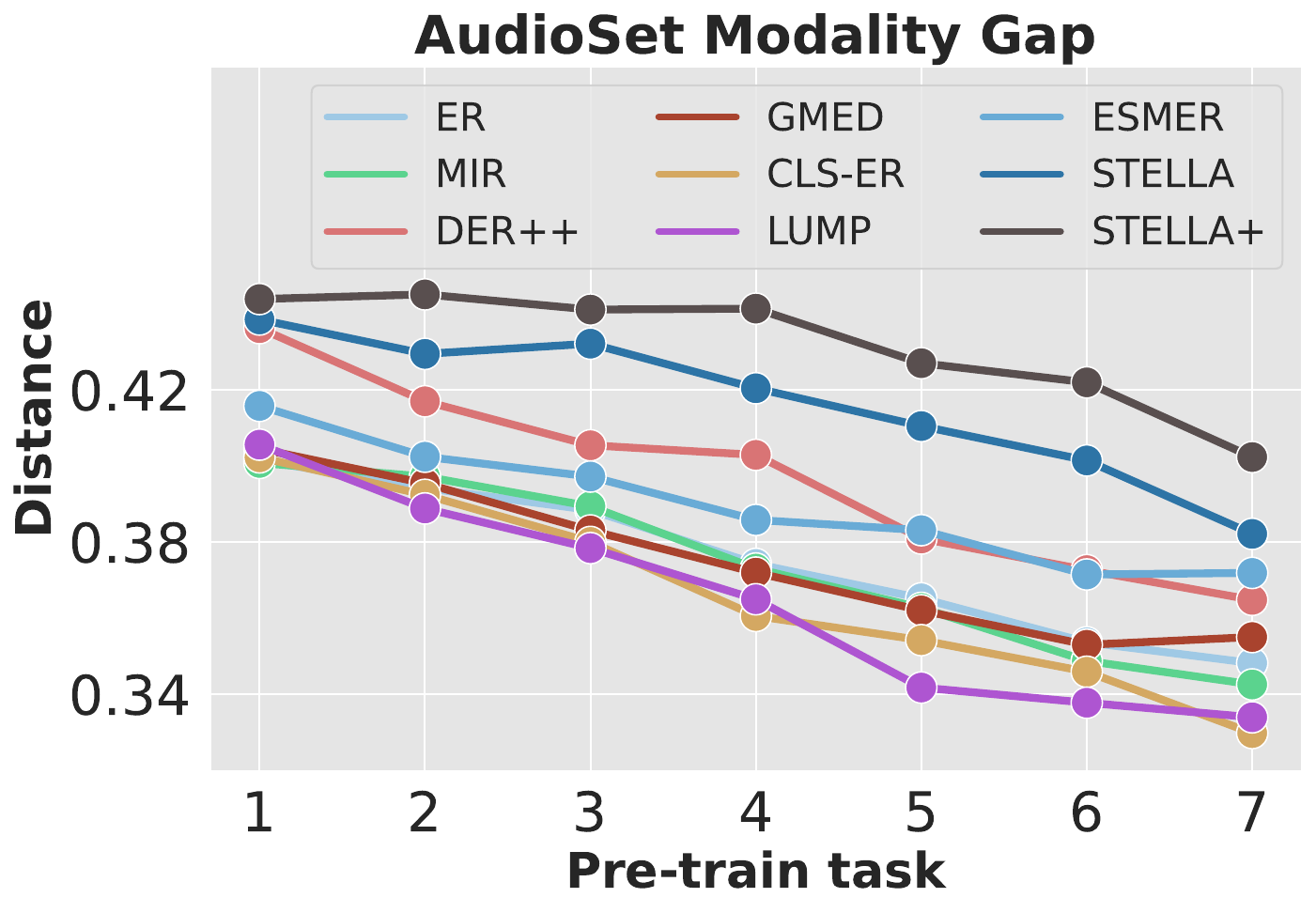} \\
        \multicolumn{2}{c}{\textbf{(a) Modality gap average decline}} &
        \textbf{(b) Modality gap after each task} \vspace{0.05in} \\
    \end{tabular}}
\caption{{\textbf{Modality gap estimation.} \textbf{(a)}:
Average modality gap decline between the modality gap estimated at the completion of the last task and the modality gap estimated at the completion of each task. \textbf{(b)}: Estimation of modality gap after the completion of each task (\textit{Continual-AS}).}}
% \vspace{-0.1in}
\label{fig:supple_modality_gap}
\end{figure*}

\section{Additional Analysis of Modality Gap \label{sec:supple:Modality Gap Analysis}}

\paragraph{Comprehensive analysis}
In the main paper, we examine the performance improvements of our approach in the context of continual audio-video pre-training with respect to the modality gap. In this section, we conduct a more detailed analysis; covering differences in the modality gap (\Cref{fig:supple_modality_gap} \highlight{(a)}), exploring the modality gap within the \textit{Continual-AS} (\Cref{fig:supple_modality_gap} \highlight{(b)}), and providing additional visualizations of the modality gap to support the effectiveness of our approach (\Cref{fig:supple_modality_gap} \highlight{(c)}).

In~\Cref{fig:supple_modality_gap} \highlight{(a)}, our approach stands out with the smallest average modality gap difference. However, our approach does not exhibit high resistance to modality gap fluctuations within the \textit{Continual-AS} experiment. An interesting observation emerges when comparing the average modality gap difference with the average forgetting in~\Cref{tab:retrieval_table}; a smaller average modality gap difference seems to correspond to lower average forgetting in the zero-shot retrieval tasks. This aligns with the relatively high average forgetting of our approach in the \textit{Continual-AS} experiment, suggesting that the modality gap difference holds potential as a metric for assessing the extent of forgetting in audio-video correlation. Meanwhile, our approach consistently maintains the highest modality gap in all pre-train tasks (\Cref{fig:supple_modality_gap} \highlight{(b)}), which explains the high average accuracy of our approach in the \textit{Continual-AS} retrieval tasks.

\begin{figure*}[t]
    \centering
    \resizebox{1.0\linewidth}{!}{%
    \begin{tabular}{ccc}
        \includegraphics[width=0.32\linewidth]{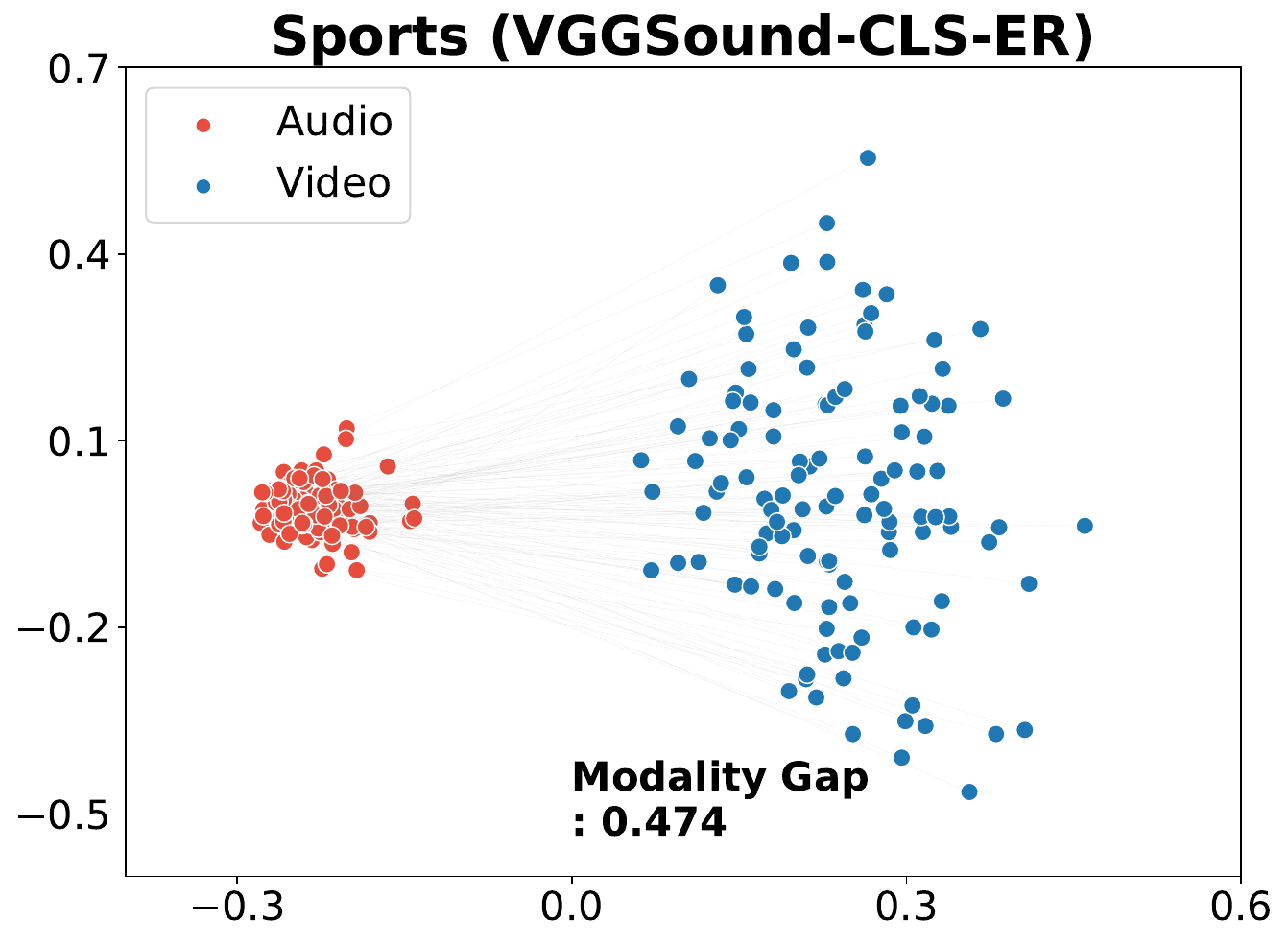}&
        \includegraphics[width=0.32\linewidth]{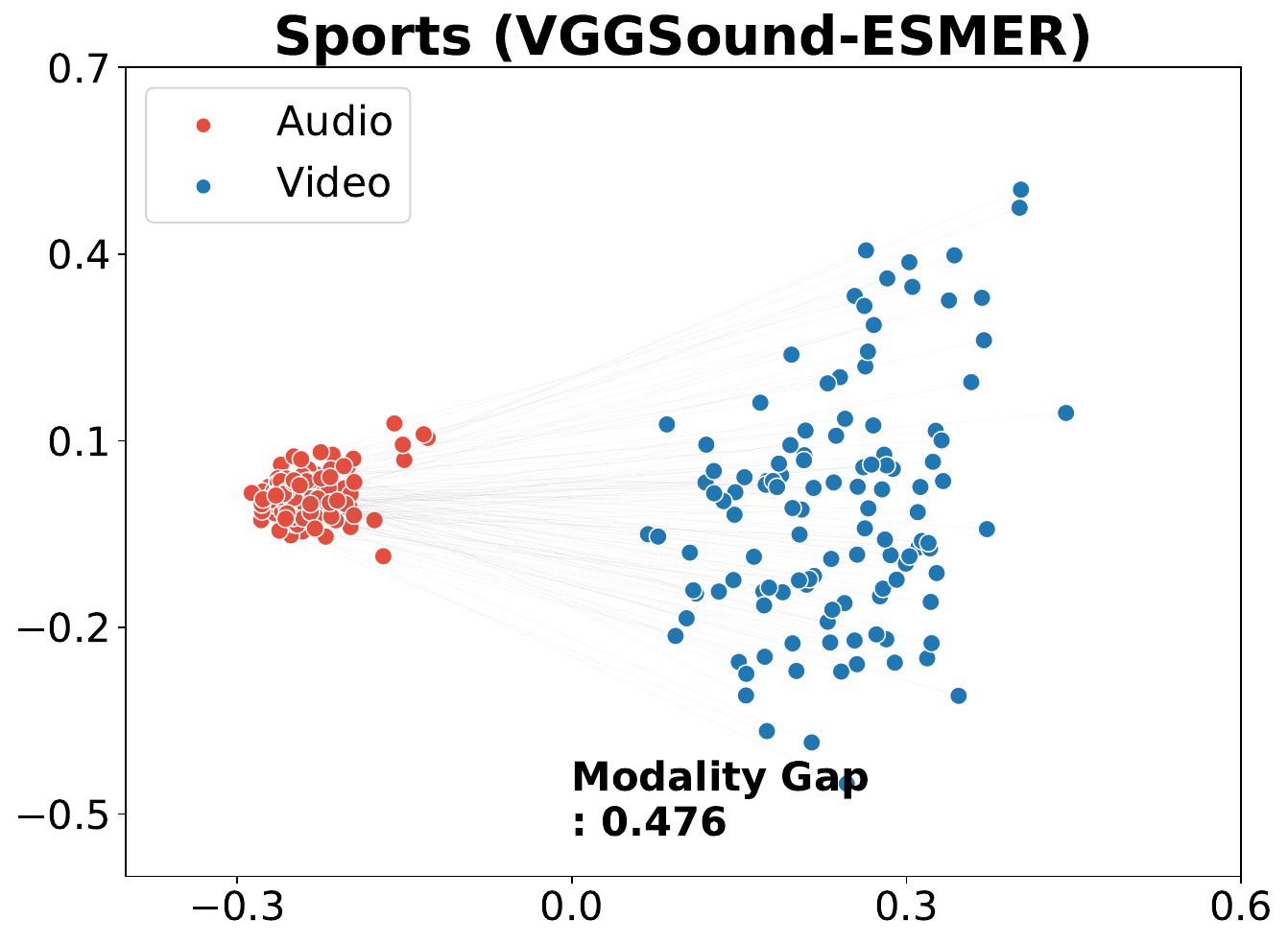}&
        \includegraphics[width=0.32\linewidth]{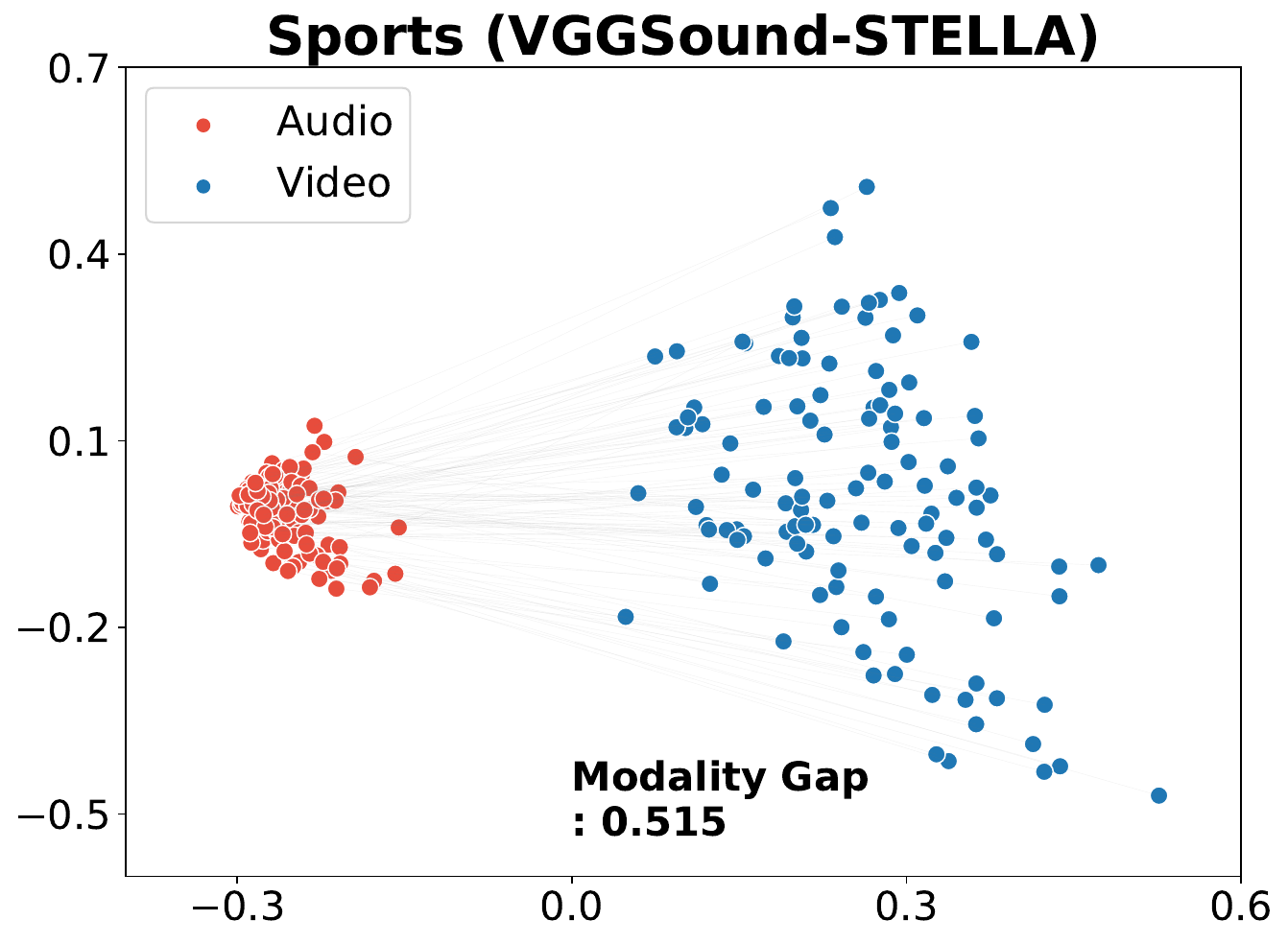} \\
        \multicolumn{3}{c}{\textbf{(a) Visualization in representation space (Continual-VS, sports)}} \vspace{0.12in} \\
        \includegraphics[width=0.32\linewidth]{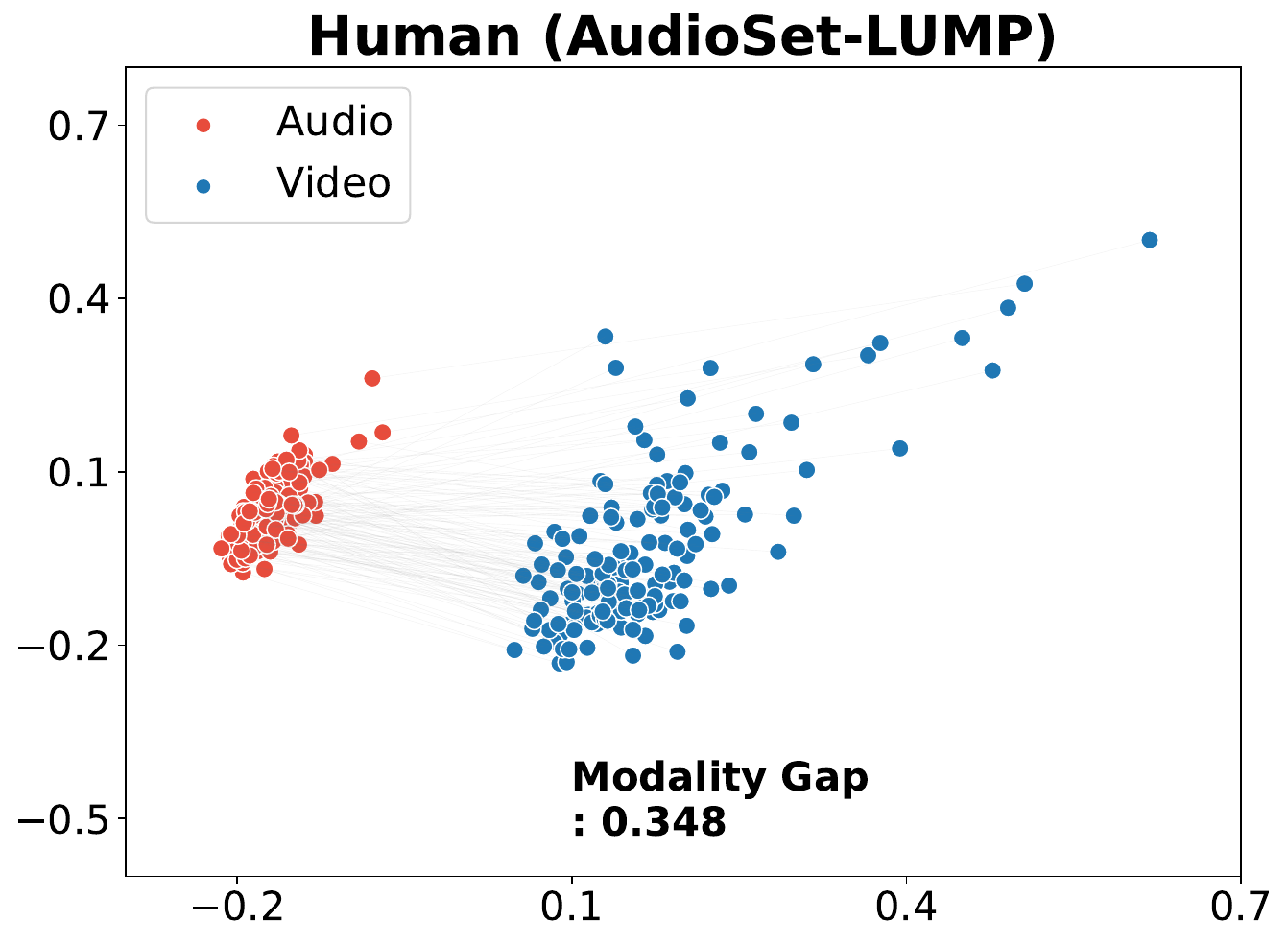}&
        \includegraphics[width=0.32\linewidth]{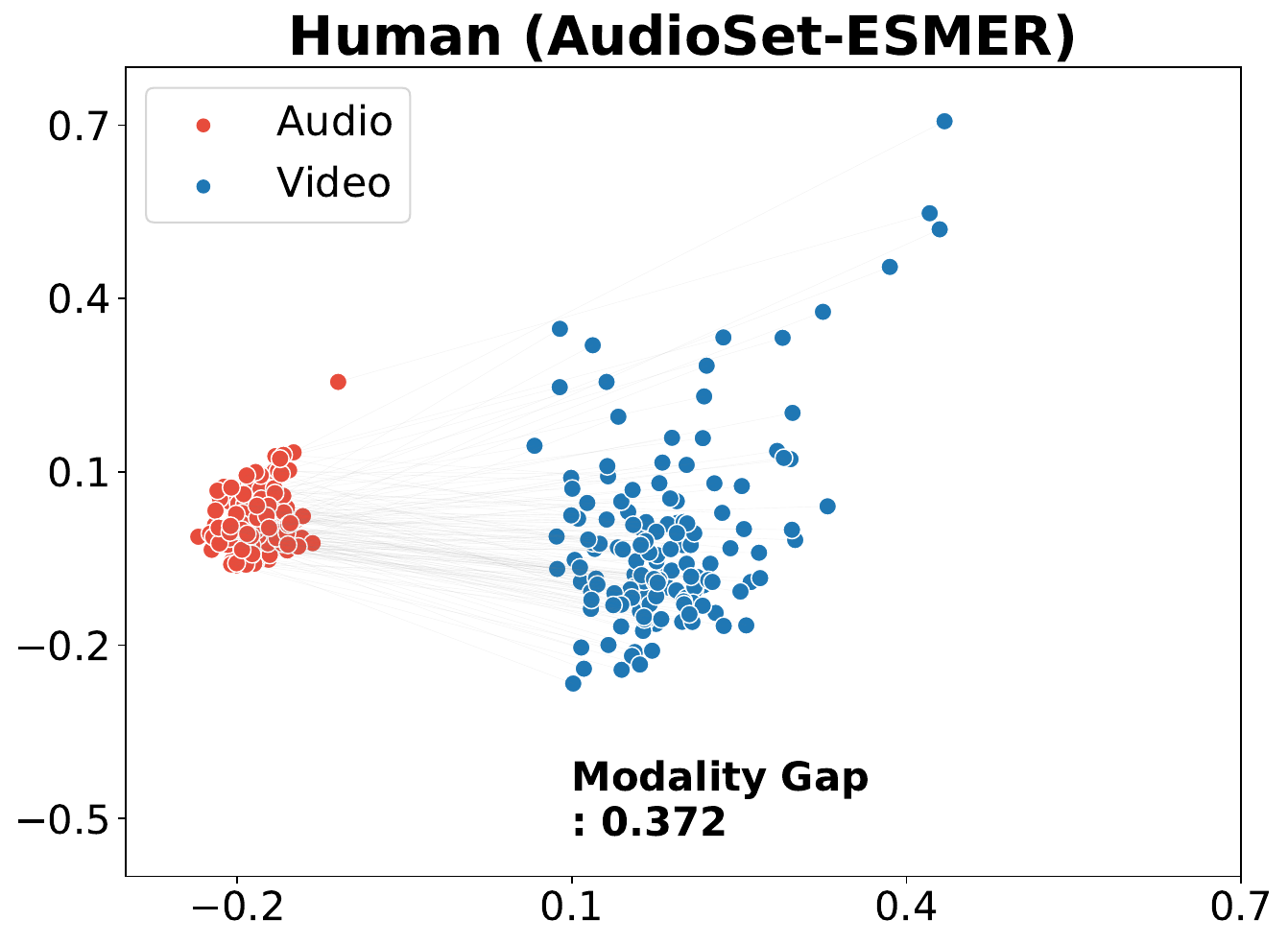}&
        \includegraphics[width=0.32\linewidth]{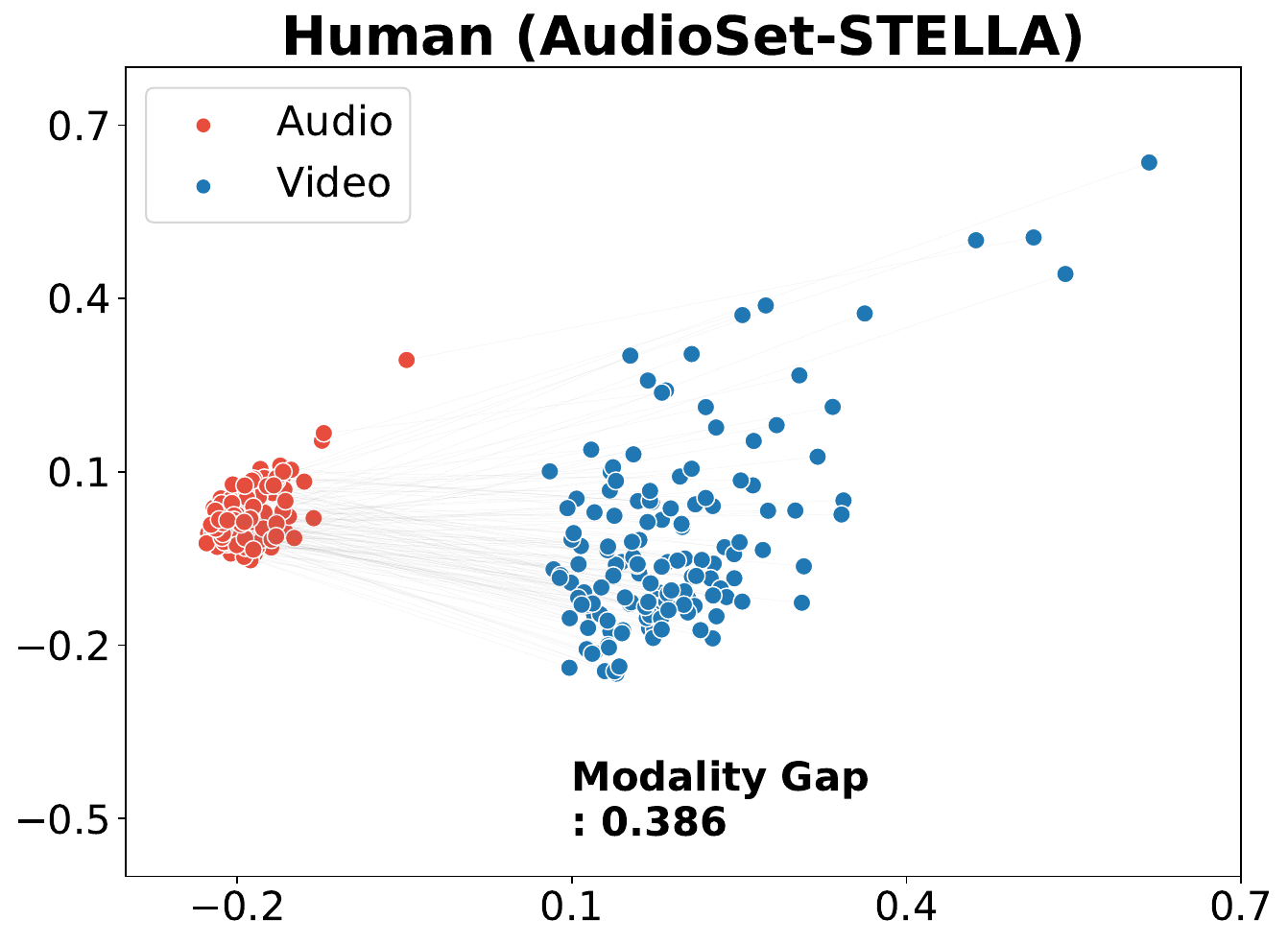} \\
        \multicolumn{3}{c}{\textbf{(b) Visualization in representation space (Continual-AS, human)}} \vspace{0.05in} \\
    \end{tabular}}
% \vspace{-0.1in}
\caption{{\textbf{Modality gap visualization.} \textbf{(a)}: Visualizations of the modality gap corresponding to the sports task with the model pre-trained up to the last task in the \textit{Continual-VS} experiment. \textbf{(b)}: Visualization of the modality gap corresponding to the human task with the model pre-trained up to the last task in the \textit{Continual-AS} experiment.}}
\label{fig:supple_modality_gap2}
\end{figure*}

We take our analysis a step further by visually representing the modality gap. In~\Cref{fig:supple_modality_gap2} \highlight{(a)}, we visualize evaluation audio-video data pairs from the sports task in the \textit{Continual-VS} experiments. Similarly, in~\Cref{fig:supple_modality_gap2} \highlight{(b)}, we visualize data from the human task in the \textit{Continual-AS} experiments. In both visualizations, we use the models that completed the continual pre-training phase. Remarkably, our approach consistently yields a larger gap in both cases. This suggests that the modality gap established from the initial task (sports, human) is effectively maintained, enabling the models to distinguish between different modalities, ultimately leading to enhanced performance.

\paragraph{Analysis on STELLA components}
\begin{figure*}[!ht]
    \centering
    \resizebox{1.0\linewidth}{!}{%
    \begin{tabular}{cc}
        \includegraphics[width=0.48\linewidth]{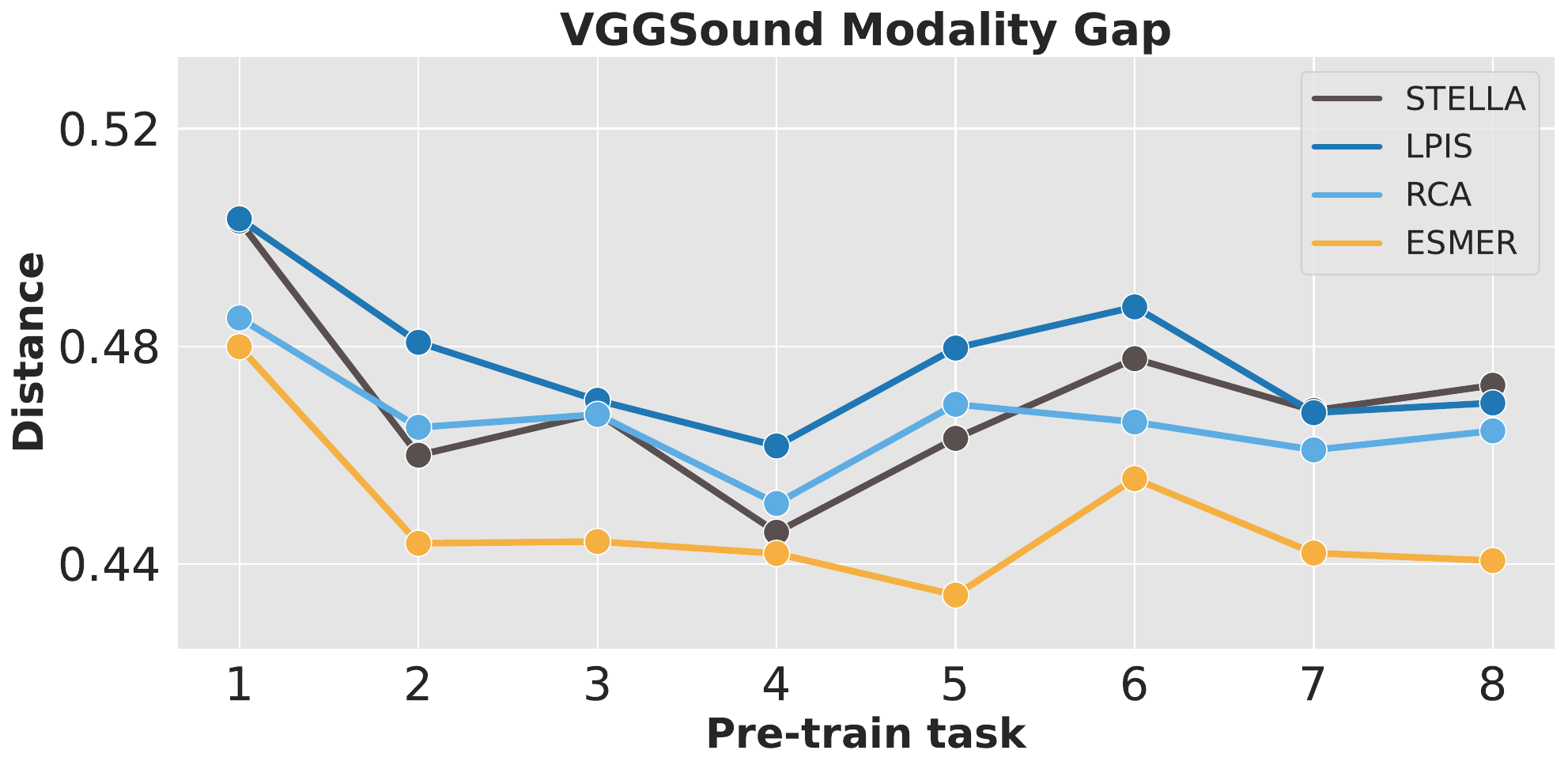}&
        \includegraphics[width=0.48\linewidth]{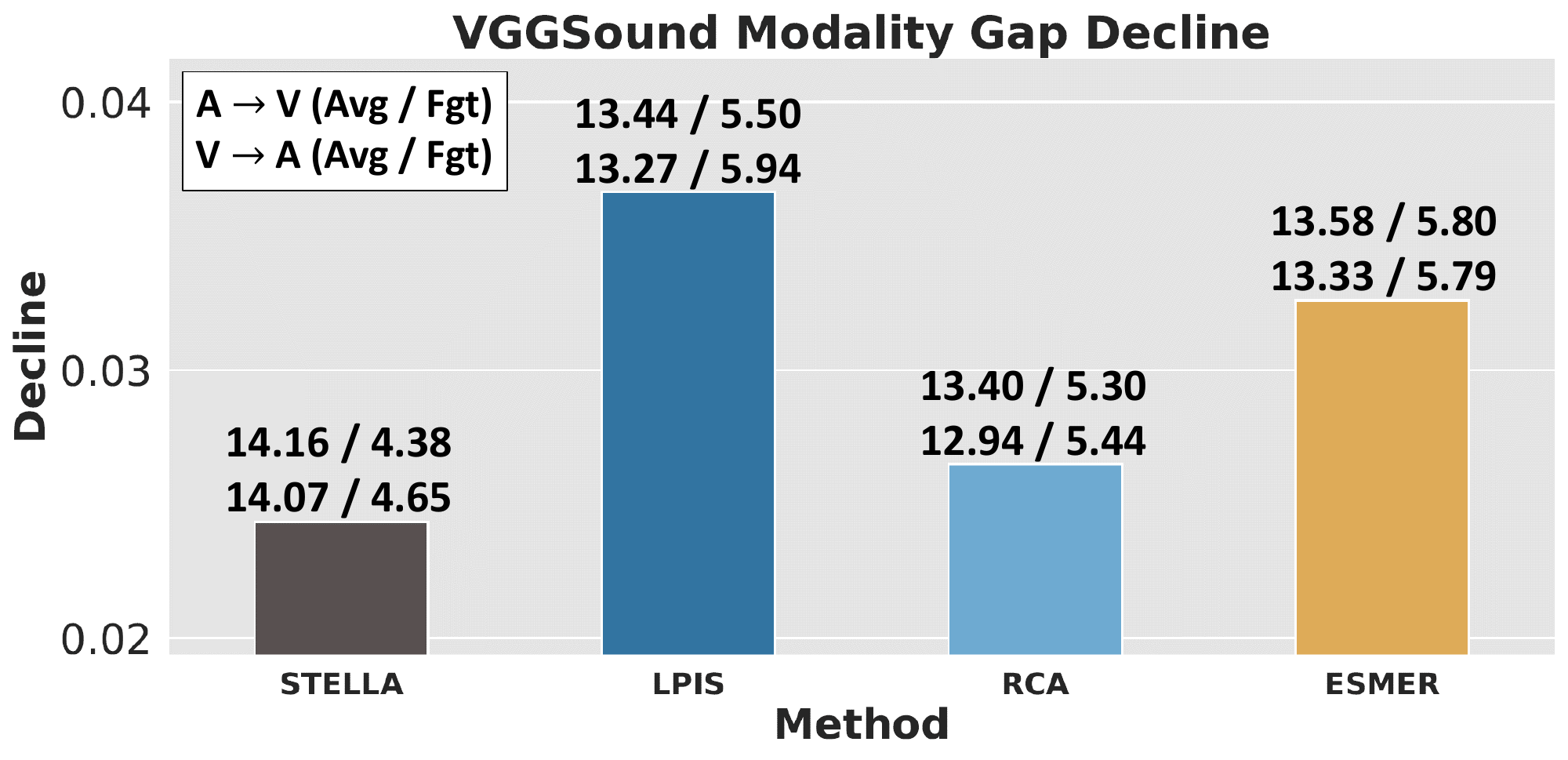} \\
        \textbf{(a) Modality gap after each task} &
        \textbf{(b) Modality gap average decline}
         \vspace{0.05in}
    \end{tabular}}
% \vspace{-0.1in}
\caption{\textbf{Modality gap estimation for each component of our proposed method.} {\textbf{(a)}: Estimation of modality gap after completing each task. \textbf{(b)}:
Average decline in modality gap between the completion of the last task and the completion of each task.}}
\label{fig:supple_component_modality_gap}
\end{figure*}

We estimate the modality gap of two key components within our proposed method: \textit{ELPP} (Efficient Localized Patch Pooling~\Cref{sec:subsec:positive region proposal}) and \textit{RCA} (Replay-guided Correlation Assessment~\Cref{sec:subsec:forget-robust selection}). The \textit{ELPP} consistently exhibits the highest modality gap across the tasks, as depicted in~\Cref{fig:supple_component_modality_gap} \highlight{(a)}. This underscores the effectiveness of the proposed method in~\Cref{sec:subsec:positive region proposal} in identifying patches that demonstrate high localized alignment with their modality pairs. Consequently, the \textit{ELPP} achieves better audio and video clustering within the multimodal representation space, resulting in enhanced average accuracy in~\Cref{tab:sub:sampling method}. This observation strongly supports our claim that the method outlined in~\Cref{sec:subsec:positive region proposal} adeptly selects informative multimodal patches from raw data.

The \textit{RCA} illustrates a relatively minor modality gap difference, as indicated in~\Cref{fig:supple_component_modality_gap} \highlight{(b)}. During the continual pre-training, the modality gap between the audio and video exhibits robustness to the effect of changing distribution. Hence, the model maintains learned audio-video alignment. This explains the small average forgetting exhibited by the \textit{RCA} in~\Cref{tab:sub:sampling method}. It affirms our claim that the method introduced in~\Cref{sec:subsec:forget-robust selection} proficiently selects forget-robust patches.

\section{Audio Patch Selection Pseudo Code \label{sec:supple:audio_constraint_pytorch}}
% \vspace{-0.3in}
\begin{minipage}[t]{1.0\linewidth}
\begin{algorithm}[H]
\caption{\small Audio time chunk selection in a PyTorch-like Style.}
\label{alg:audio_time_chunk_select_pytorch}
\small
\begin{algorithmic}[!]
\STATE \PyComment{I\_a: audio patch importance score}
\STATE \PyComment{P\_a: audio pruning probability matrix}
\STATE \PyComment{L\_c: audio time chunk size}
\STATE \PyComment{kappa\_a: target number of audio tokens}
\STATE \PyComment{num\_time: the number of tokens in time dimension}
\STATE \PyComment{num\_freq: the number of tokens in frequency dimension}
\STATE \PyCommand{def} \PyCode{audio\_time\_chunk\_selection(I\_a,P\_a):}
    \STATE \hspace{0.5cm} \PyCode{F\_a=bernoulli(P\_a)}
    \STATE \hspace{0.5cm} \PyCode{F\_a=F\_a.reshape(num\_time, num\_freq)} 
    \STATE \hspace{0.5cm} \PyCode{F\_a\_t=F\_a.sum(dim=1)} \PyComment{$\#$ of pruned patches}
    \STATE \hspace{0.5cm} \PyCode{I\_a\_t=I\_a.reshape(num\_time, num\_freq)}
    \STATE \hspace{0.5cm} \PyCode{I\_a\_t=I\_a\_time.sum(dim=1)} \PyComment{Time-wise importance}
    \STATE \hspace{0.5cm} \PyCode{I\_a\_c=avg\_pool(I\_a\_t, kernel\_size=L\_c)} \PyComment{Chunk-wise importance}
    \STATE \hspace{0.5cm} \PyCode{num\_chunk=len(I\_a\_c)}
    \STATE \hspace{0.5cm} \PyCode{t\_select=multinomial(I\_a\_c, num\_samples=num\_chunk)}
    \STATE \hspace{0.5cm} \PyCode{num\_tokens=0}
    \STATE \hspace{0.5cm} \PyCommand{for} \PyCode{j} \PyCommand{in} \PyCode{range(num\_chunk):}
        \STATE \hspace{1.0cm} \PyCode{t=t\_select[j]}
        \STATE \hspace{1.0cm} \PyCode{num\_prune=F\_a\_t[t*L\_c:(t+1)*L\_c].sum()} \PyComment{$\#$ of pruned patches}
        \STATE \hspace{1.0cm} \PyCode{num\_tokens+=(L\_c*num\_freq - num\_prune)} \PyComment{Count $\#$ of patches}
        \STATE \hspace{1.0cm} \PyCommand{if} \PyCode{num\_tokens > kappa\_a:}
            \STATE \hspace{1.5cm} \PyCode{F\_last=F\_a[t*L\_c:(t+1)*L\_c].view(-1)}
            \STATE \hspace{1.5cm} \PyCode{F\_last\_accum=cumsum(flip($\sim$F\_last))}
            \STATE \hspace{1.5cm} \PyCode{prune\_tail\_idx= F\_last\_accum == num\_tokens-kappa\_a}
            \STATE \hspace{1.5cm} \PyCode{F\_last[-(prune\_tail\_idx+1):]=True} \PyComment{Prune tail of last chunk}
        \STATE \hspace{1.5cm} \PyCode{F\_a[t*L\_c:(t+1)*L\_c]=F\_last.reshape(num\_time,num\_freq)}
        \STATE \hspace{1.5cm} \PyCommand{for} \PyCode{k} \PyCommand{in} \PyCode{range(j+1, num\_chunk):}
            \STATE \hspace{2.0cm} \PyCode{t\_prune=t\_select[k]}
            \STATE \hspace{2.0cm} \PyCode{F\_a[t\_prune*L\_c:(t\_prune+1)*L\_c]=True}
        \STATE \hspace{1.5cm} \PyCommand{break}
    \STATE \hspace{0.5cm} \PyCode{F\_a=F\_a.view(-1).float()}
    \STATE \hspace{0.5cm} \PyCode{S\_tilde\_a=argsort(F\_a)} \PyComment{Forget-robust audio sorted indices}
    \STATE \hspace{0.5cm} \PyCommand{return} \PyCode{S\_tilde\_a}
\end{algorithmic}
\end{algorithm}
\vspace{-0.2in}
\end{minipage}

% \newpage

\section{Algorithms of STELLA (Spatio-Temoporal Localized Alignment) and STELLA + \label{sec:supple:stella_p_algo}}
\begin{figure}[h!]
\vspace{-0.2in}
\input{materials/stella_algo}
\input{materials/stella_p_algo}
\end{figure}

\section{Visualization of Fading Audio-Visual Attention \label{sec:supple:fading audio-visual attention}}

\begin{figure*}[t]
    \centering
    % \small
    \begin{minipage}{\textwidth}
        \centering
        \includegraphics[width=0.41\linewidth]{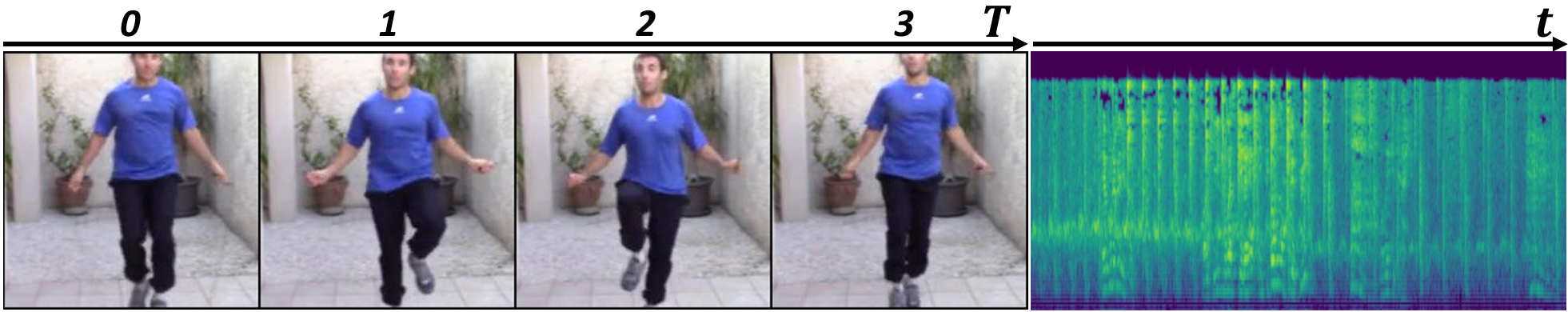}
        \includegraphics[width=0.41\linewidth]{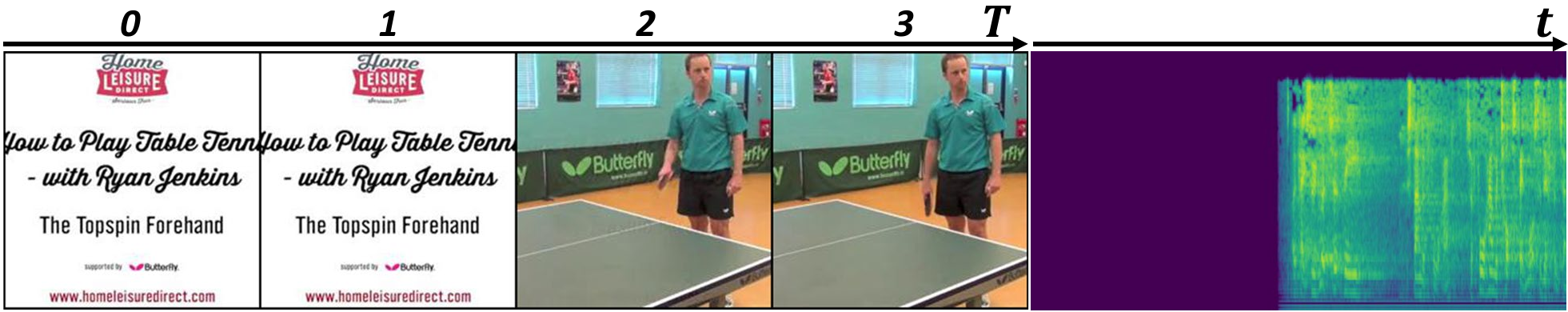}
    \end{minipage}
    \textbf{(a) Raw data}
    \vspace{0.05in}
    
    \begin{minipage}{\textwidth}
        \centering
        \includegraphics[width=0.41\linewidth]{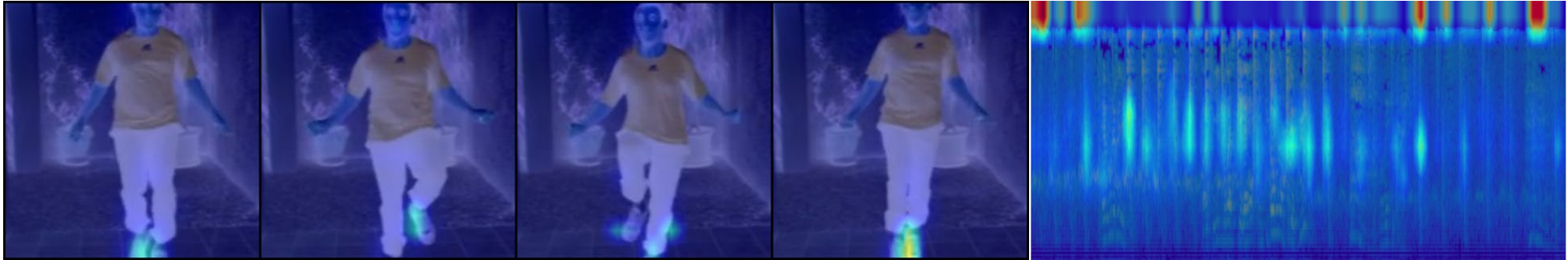}
        \includegraphics[width=0.41\linewidth]{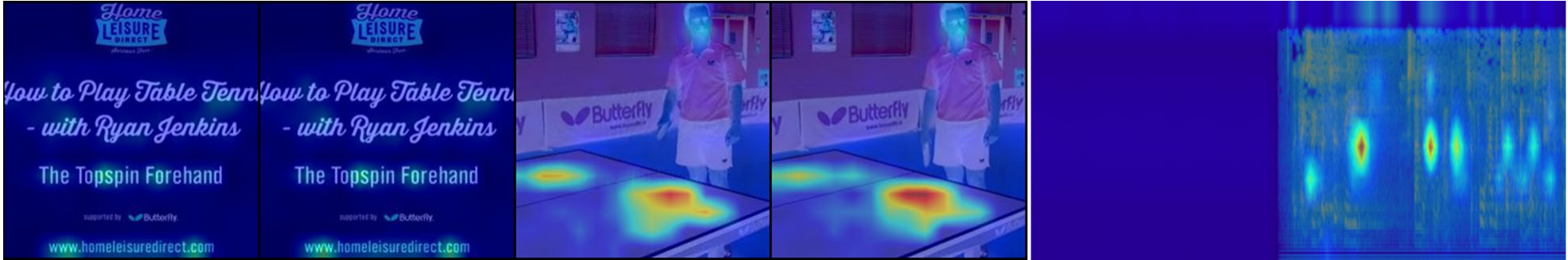}
    \end{minipage}
    \textbf{(b) Sparse audiovisual correlation}
    \vspace{0.05in}

    \begin{minipage}{\textwidth}
        \centering
        \includegraphics[width=0.41\linewidth]{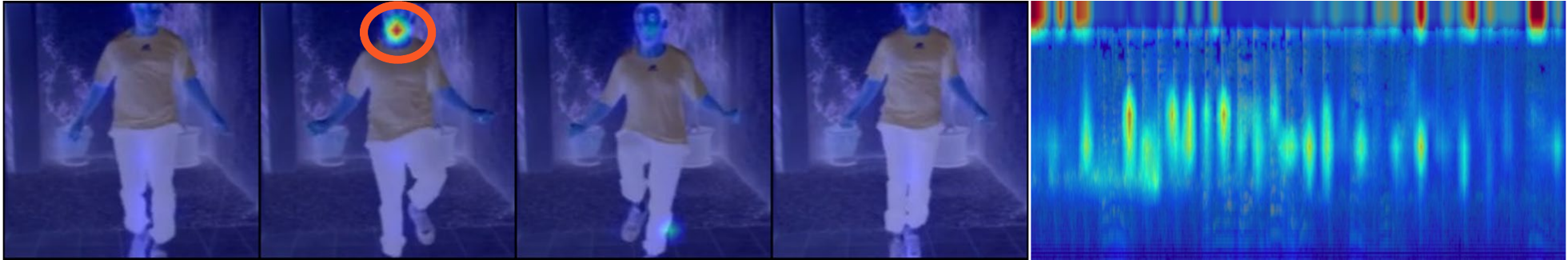}
        \includegraphics[width=0.41\linewidth]{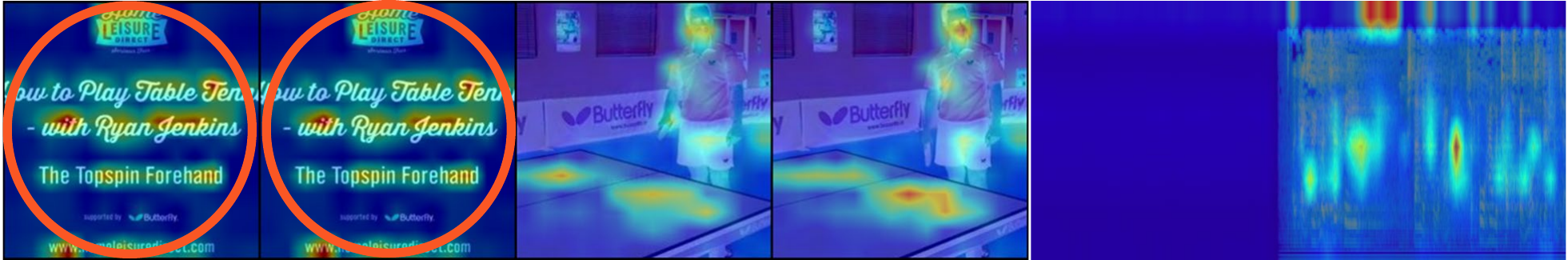}
    \end{minipage}
    \textbf{(c) Multimodal correlation forgetting (DER++)}
    \vspace{0.05in}

    \begin{minipage}{\textwidth}
        \centering
        \includegraphics[width=0.41\linewidth]{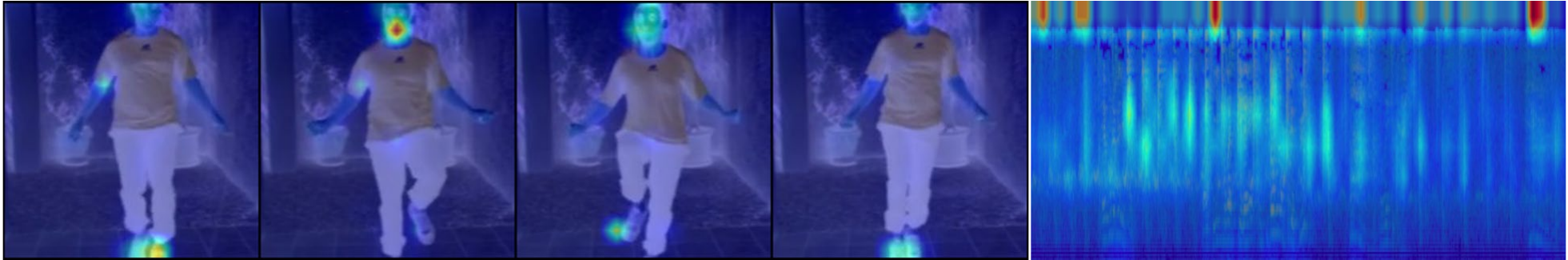}
        \includegraphics[width=0.41\linewidth]{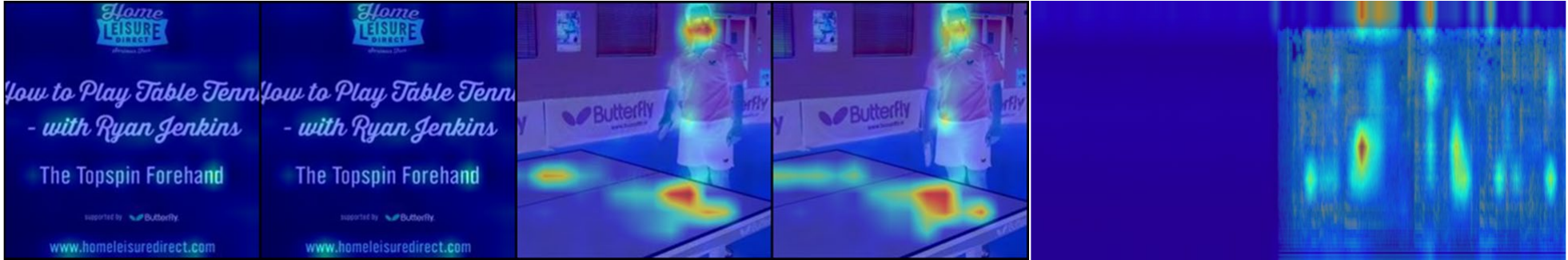}
    \end{minipage}
    \textbf{(d) Multimodal correlation forgetting (Ours)}
    \vspace{0.15in}

    \begin{minipage}{\textwidth}
        \centering
        \includegraphics[width=0.41\linewidth]{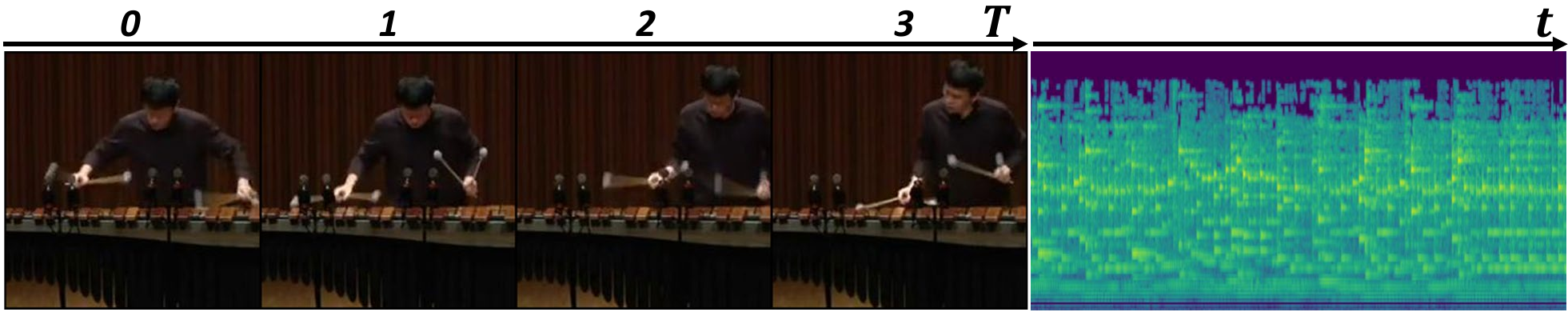}
        \includegraphics[width=0.41\linewidth]{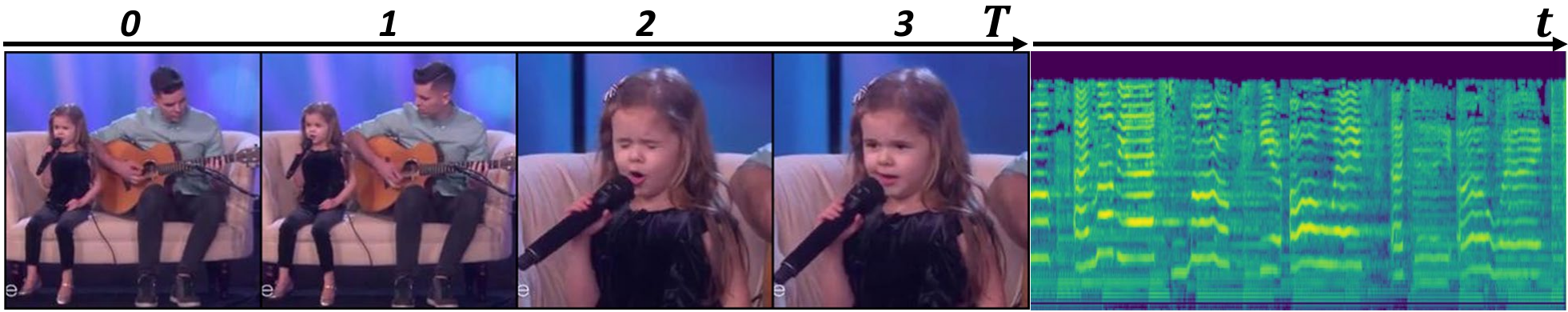}
    \end{minipage}
    \textbf{(a) Raw data}
    \vspace{0.05in}
    
    \begin{minipage}{\textwidth}
        \centering
        \includegraphics[width=0.41\linewidth]{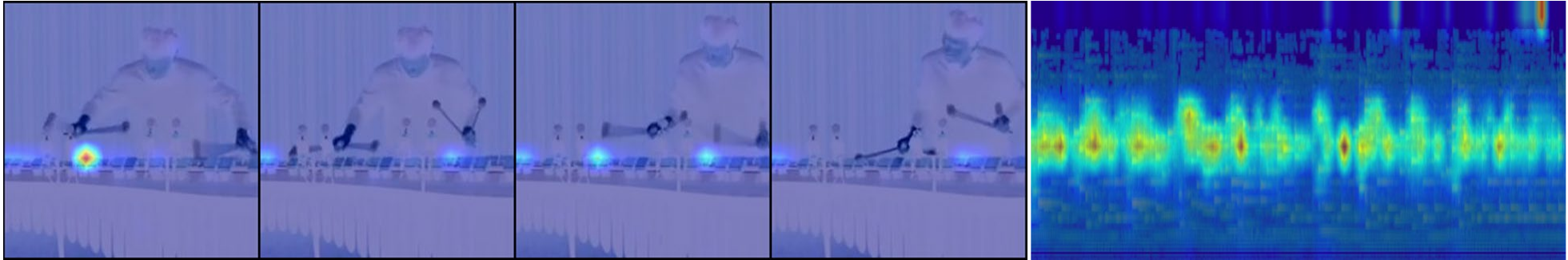}
        \includegraphics[width=0.41\linewidth]{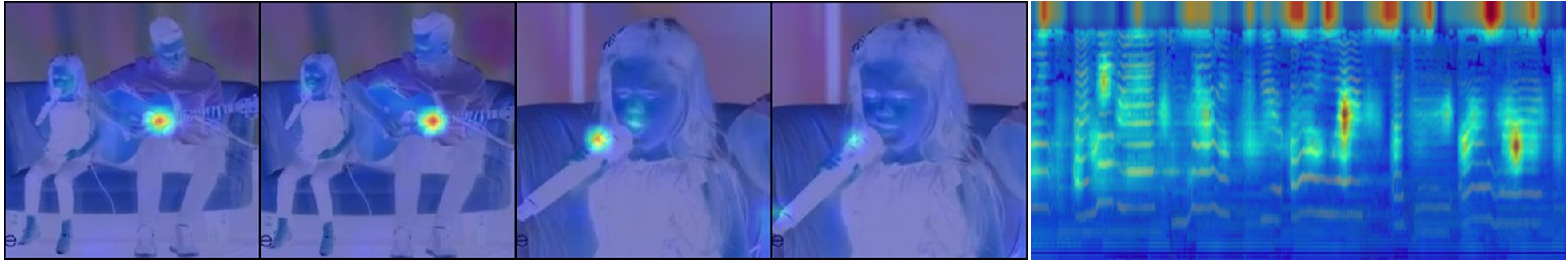}
    \end{minipage}
    \textbf{(b) Sparse audiovisual correlation}
    \vspace{0.05in}

    \begin{minipage}{\textwidth}
        \centering
        \includegraphics[width=0.41\linewidth]{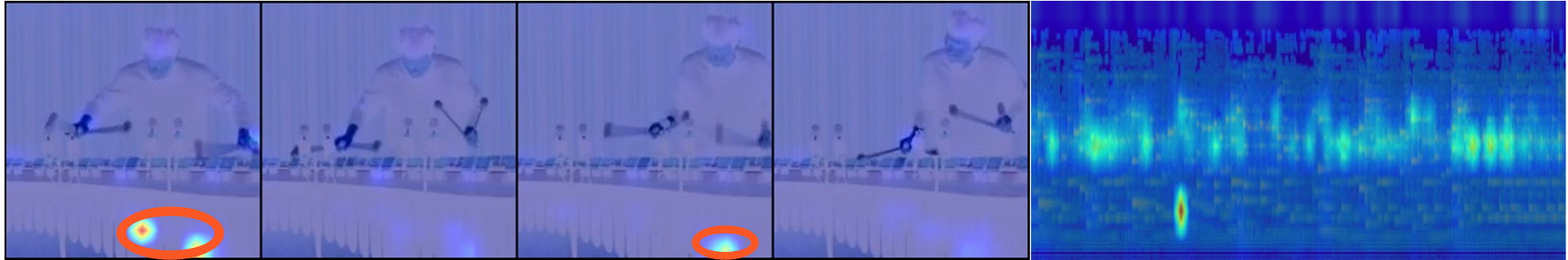}
        \includegraphics[width=0.41\linewidth]{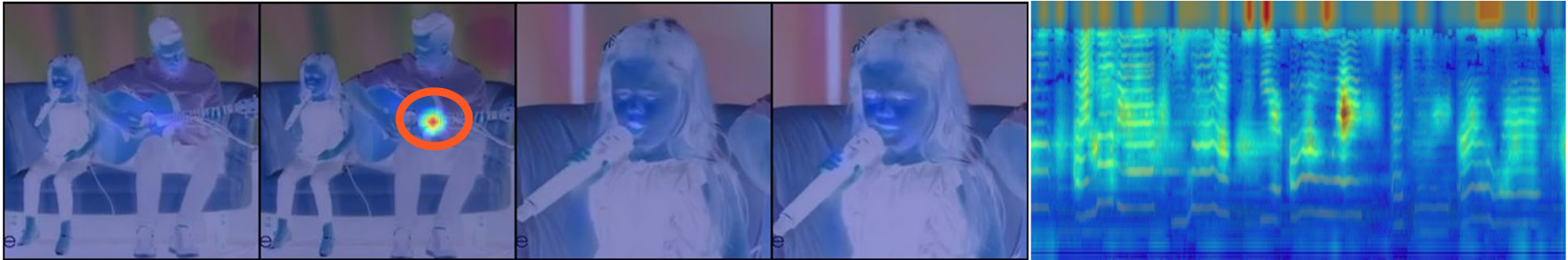}
    \end{minipage}
    \textbf{(c) Multimodal correlation forgetting (DER++)}
    \vspace{0.05in}

    \begin{minipage}{\textwidth}
        \centering
        \includegraphics[width=0.41\linewidth]{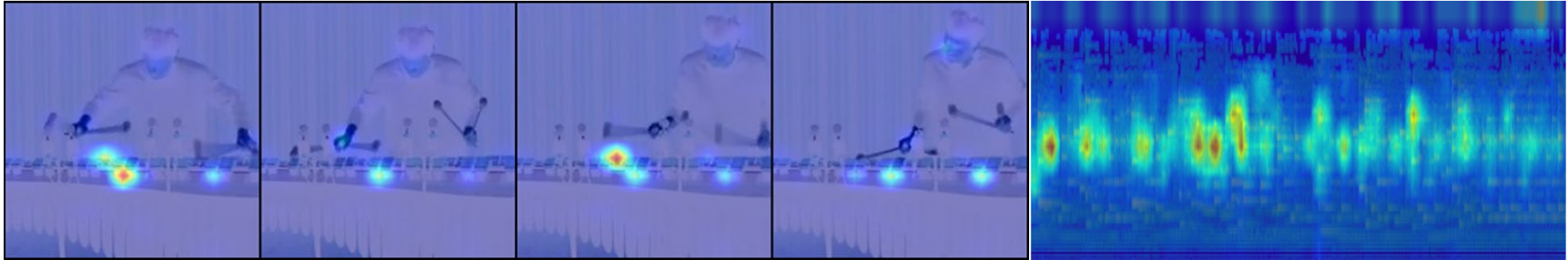}
        \includegraphics[width=0.41\linewidth]{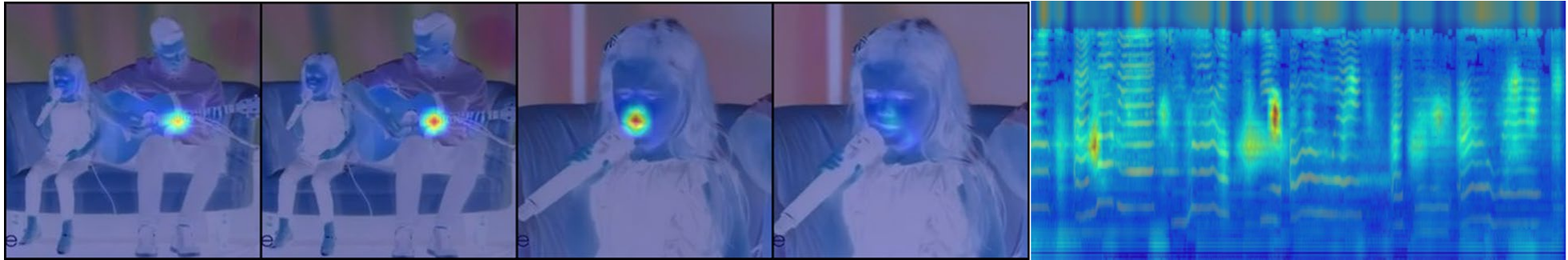}
    \end{minipage}
    \textbf{(d) Multimodal correlation forgetting (Ours)}
    \vspace{0.15in}

    \begin{minipage}{\textwidth}
        \centering
        \includegraphics[width=0.41\linewidth]{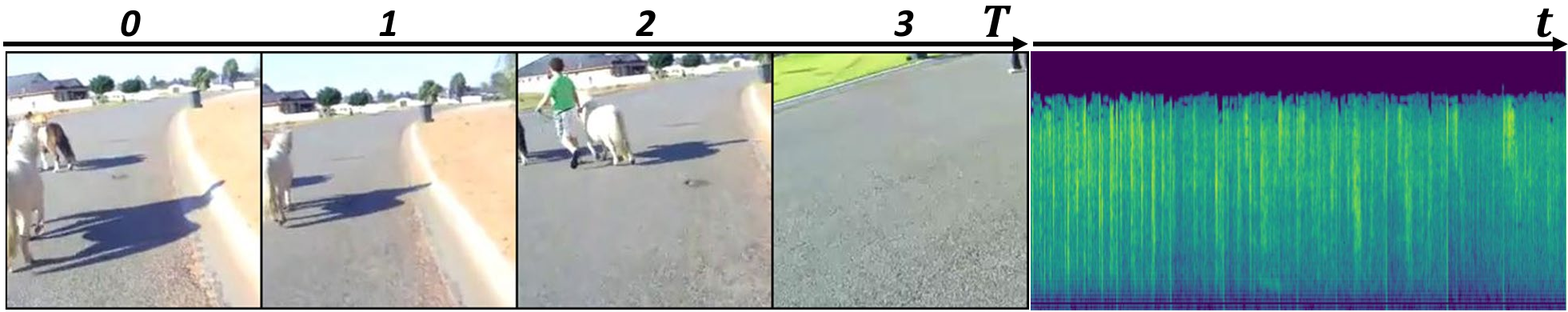}
        \includegraphics[width=0.41\linewidth]{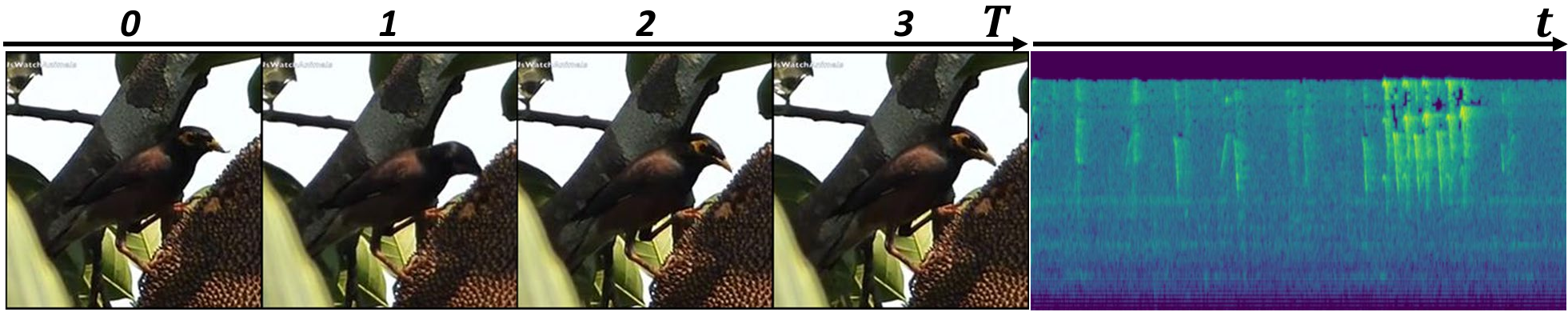}
    \end{minipage}
    \textbf{(a) Raw data}
    \vspace{0.025in}
    
    \begin{minipage}{\textwidth}
        \centering
        \includegraphics[width=0.41\linewidth]{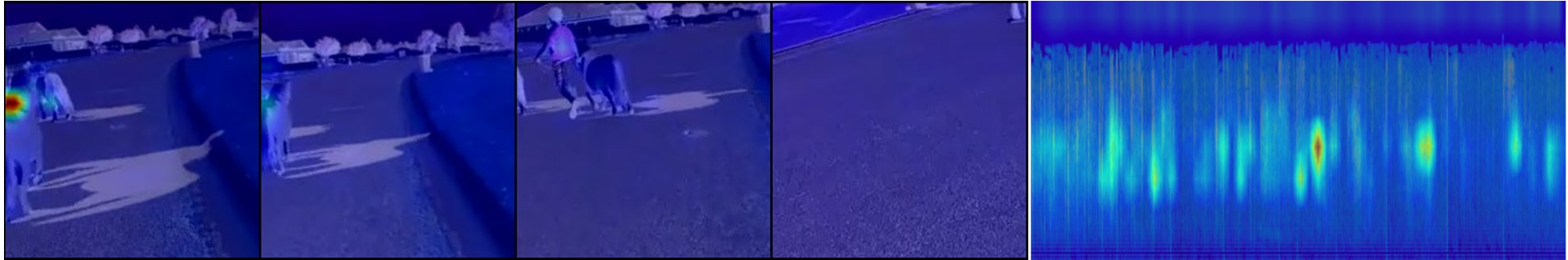}
        \includegraphics[width=0.41\linewidth]{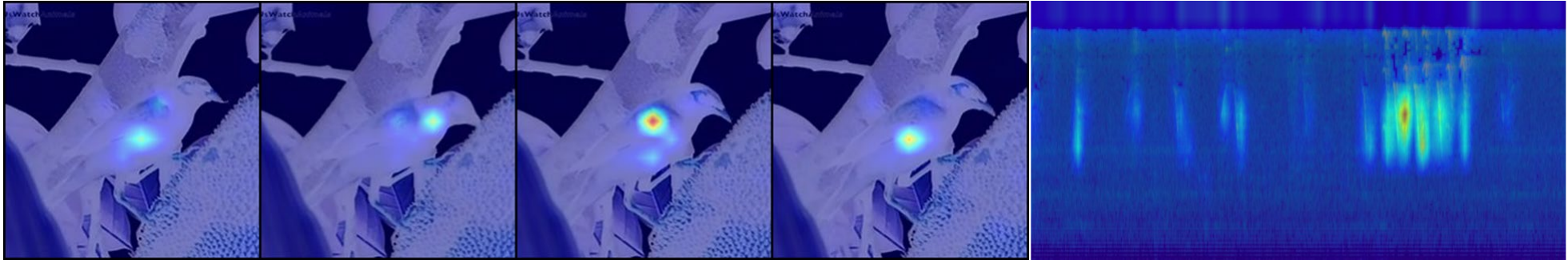}
    \end{minipage}
    \textbf{(b) Sparse audiovisual correlation}
    \vspace{0.025in}

    \begin{minipage}{\textwidth}
        \centering
        \includegraphics[width=0.41\linewidth]{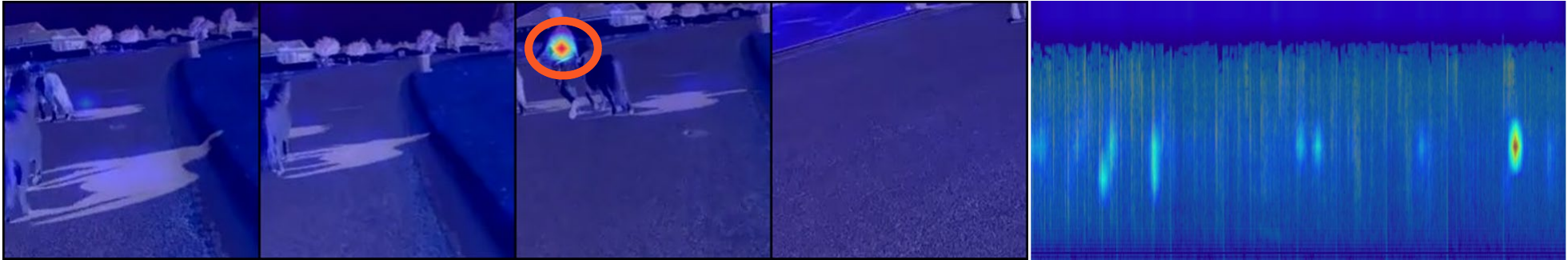}
        \includegraphics[width=0.41\linewidth]{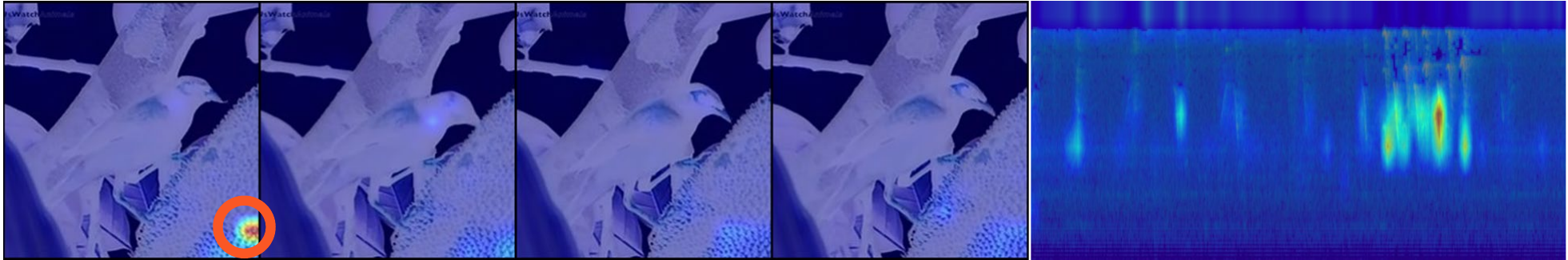}
    \end{minipage}
    \textbf{(c) Multimodal correlation forgetting (DER++)}
    \vspace{0.025in}

    \begin{minipage}{\textwidth}
        \centering
        \includegraphics[width=0.41\linewidth]{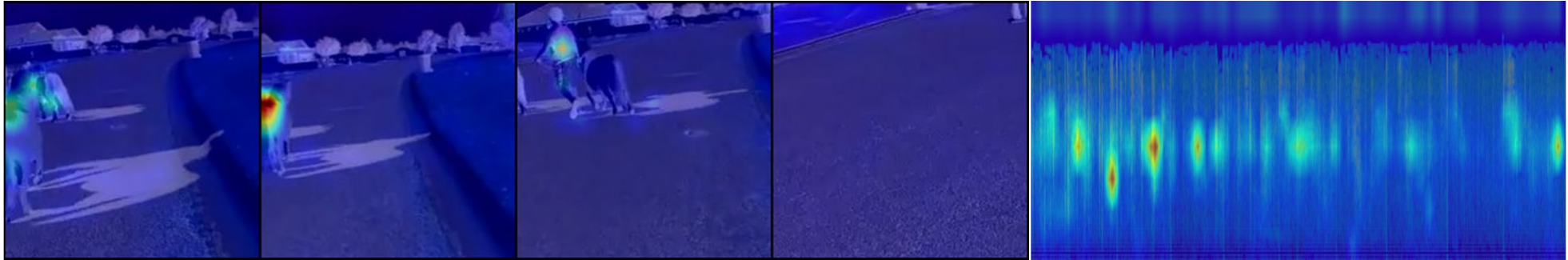}
        \includegraphics[width=0.41\linewidth]{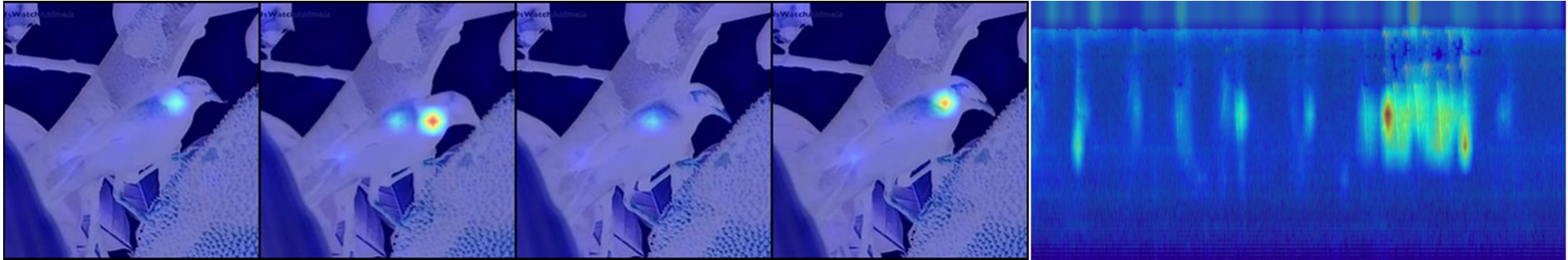}
    \end{minipage}
    \textbf{(d) Multimodal correlation forgetting (Ours)}
    % \vspace{-0.1in}
    
    \captionsetup{justification=justified} % Reset caption justification
    \caption{\small \textbf{Visualization of cross-attention maps.} \textbf{(a)} Examples of raw data pairs. We visualize cross-attention maps of the pairs in \textbf{(b)}. The closer the color is to red, the higher the attention score. While the baseline model using \textit{DER++} attends to entirely different parts as can be seen in \textbf{(c)}, our method attends to a similar part even after being trained on two additional tasks as presented in \textbf{(d)}. The wrong attention region is marked in an orange circle.}
    \label{fig:supple_fading_attention}
    % \vspace{-0.075in}
\end{figure*}

As shown in~\Cref{fig:fading_attention} of the main paper, we tackle the problem of forgetting the past audio-video correlation by visualizing the attention maps. In~\Cref{fig:supple_fading_attention}, we provide additional examples that vividly illustrate the challenge of forgetting past correlation as the model undergoes pre-training on sequential tasks.

In the top-left example of~\Cref{fig:supple_fading_attention}, we observe a video example where a person is engaged in rope skipping. The initial attention map concentrated on the feet (\highlight{(b)}). However, as the model adapts to new tasks, the attention map is shifted solely to the person's face (\highlight{(c)}), implying the gradual erosion of the correlation between the sound of rope skipping and the corresponding jumping motion. In the top-right example of~\Cref{fig:supple_fading_attention}, the attention map undergoes an intriguing shift towards an unrelated caption in the first two frames (\highlight{(c)}). Moving on to the middle-left example in~\Cref{fig:supple_fading_attention}, the model initially demonstrates a keen understanding of the xylophone's location where the sound originates (\highlight{(b)}). However, subsequent training on additional tasks weakens auditory attention, and the model fails to locate the sounding region (\highlight{(c)}). This challenge becomes more pronounced when multiple sounding objects are involved. In the middle-right example in~\Cref{fig:supple_fading_attention}, we explore a scenario where a child is singing alongside a man playing the guitar. The initial visual attention map correctly identifies both the guitar and the child's mouth. Nevertheless, as the model undergoes continuous training, the correlation between the singing voice and the child's visual presence diminishes, and the model connects the sound with the guitar only (\highlight{(c)}). Similarly, in the bottom-left example of~\Cref{fig:supple_fading_attention}, the visual attention map shifts from the horse to the human, accompanied by the weakening of auditory attention towards the horse's clip-clop sound (\highlight{(b)}). Lastly, in the bottom-right example of~\Cref{fig:supple_fading_attention}, despite the presence of only one prominent sounding object, the bird, the visual attention map is activated on the uncorrelated object. However, our approach successfully mitigates this forgetting problem, as demonstrated in \highlight{(d)} of the example, where the attention maps remain consistent with the initial attention maps.


\section{Limiations \label{sec:supple:limitations}}
Our approach involves an additional inference step for patch selection, leveraging the AVM module on top of the backbone model. While this significantly reduces GPU memory consumption, it does incur additional computational overhead, yielding a relatively small improvement in throughput. To address this challenge, one potential solution is to develop a student model that integrates the AVM module and utilizes knowledge distillation to transfer audio-video representation from the backbone model. Recognizing the importance of enhancing efficiency, we acknowledge the need for future research to explore effective strategies for utilizing the AVM module. This avenue of improvement is a key component of our future research.

\end{document}